\documentclass[]{clv3}

\usepackage{hyperref}
\usepackage{booktabs}
\usepackage{caption}
\usepackage{forest}
\usepackage{multirow}
\usepackage{subfigure}
\usepackage{graphicx}
\usepackage{subcaption}
\usepackage{indentfirst}
\usepackage{enumerate,enumitem}

\usepackage{xcolor}
\definecolor{darkblue}{rgb}{0, 0, 0.5}
\hypersetup{colorlinks=true,citecolor=darkblue, linkcolor=darkblue, urlcolor=darkblue}

\begin{document}


\runningtitle{Survey of Cultural Awareness in Language Models: Text and Beyond}

\runningauthor{Pawar \& Park et al.}


\title{Survey of Cultural Awareness in Language Models: Text and Beyond}


\author{Siddhesh Pawar\thanks{Equal contributions. E-mail: sipa@di.ku.dk, jjjunyeong9986@kaist.ac.kr}}
\affil{University of Copenhagen, Denmark}

\author{Junyeong Park$^{*}$}
\affil{KAIST, Republic of Korea}

\author{Jiho Jin}
\affil{KAIST, Republic of Korea}

\author{Arnav Arora}
\affil{University of Copenhagen, Denmark}

\author{Junho Myung}
\affil{KAIST, Republic of Korea}

\author{Srishti Yadav}
\affil{University of Copenhagen, Denmark}

\author{Faiz Ghifari Haznitrama}
\affil{KAIST, Republic of Korea}

\author{Inhwa Song}
\affil{KAIST, Republic of Korea}

\author{Alice Oh}
\affil{KAIST, Republic of Korea}

\author{Isabelle Augenstein}
\affil{University of Copenhagen, Denmark}

\maketitle

\begin{abstract}
Large-scale deployment of large language models (LLMs) in various applications, such as chatbots and virtual assistants, requires LLMs to be culturally sensitive to the user to ensure inclusivity.  Culture has been widely studied in psychology and anthropology, and there has been a recent surge in research on making LLMs more culturally inclusive in LLMs that goes beyond multilinguality and builds on findings from psychology and anthropology. In this paper, we survey efforts towards incorporating cultural awareness into text-based and multimodal LLMs. We start by defining cultural awareness in LLMs, taking the definitions of culture from anthropology and psychology as a point of departure. We then examine methodologies adopted for creating cross-cultural datasets, strategies for cultural inclusion in downstream tasks, and methodologies that have been used for benchmarking cultural awareness in LLMs. Further, we discuss the ethical implications of cultural alignment, the role of Human-Computer Interaction in driving cultural inclusion in LLMs, and the role of cultural alignment in driving social science research. We finally provide pointers to future research based on our findings about gaps in the literature.\footnote{We additionally organize the papers covered by this survey at \url{https://github.com/siddheshih/culture-awareness-llms.git}.}
\end{abstract}

\section{Introduction}

Language models are deployed in various user-facing applications, such as recommender systems \citep{bao2023bi}, customer service \citep{pandya2023automating}, and search applications \citep{xiong2024search}, which are increasingly used by people in all aspects of their life including education \citep{kasneci}, public health \citep{dehealth}, and professional writing \citep{jakesch2023co}.
These models reflect the Western perspective, predominantly trained on Western-centric data \citep{durmus2023towards}. This skewed perspective can lead to stereotyping and alienation of users, propagation of stereotypes due to a lack of cultural understanding (e.g., flattening of cultural identities), or responding in a culturally insensitive way \citep{cao-etal-2022-theory, cao-etal-2023-assessing}.
Therefore, cultural awareness is one of the critical factors that should be considered while creating NLP models.

In this work, we provide a comprehensive survey of the steps that the NLP community has taken to make language models more culturally inclusive. Furthermore, with advancements in multimodal foundation models and their adaption on NLP tasks \citep{fei2022towards}, we also examine efforts towards cultural inclusion in multimodal NLP systems (i.e., multimodal systems with language understanding as one of their components). As the notion of culture used by the NLP community (to define and ensure cultural inclusion in NLP systems) is adopted from social science research, we start by defining `cultural awareness in LLMs' based on definitions of culture in psychology and anthropology literature. We then consolidate the works that look into cultural inclusion in LLMs and multimodal models, including benchmark creation, training data creation, alignment methodologies, and evaluation methodologies. We also discuss the role of cultural alignment in accelerating social research. Human-computer interaction (HCI) also plays a role in ensuring cultural alignment in LLMs, as how studying different cultures reacts to certain levels of cultural (mis)alignment and matching varied expectations of people falls under the realm of HCI research ~\citep{weidinger2023sociotechnical}. Finally, we discuss the ethical and safety implications of current research directions and provide potential research avenues that the community could take to foster cultural inclusion in language models. While recent surveys \citep{liu2024culturally, adilazuarda2024towards} focus on the cultural alignment of LLMs in NLP and provide a taxonomy for grouping current cultural alignment works, we consolidate the literature from a broader scope. We survey and compare efforts towards incorporation and conceptualization of culture in NLP systems, and our survey spans several modalities, including images, videos, and audio, along with text. We position our survey at the intersection of NLP, multimodality, and social science. 

The key contributions and research goals of this survey are as follows:
\begin{enumerate}
\item We review 300+ papers to provide an overview of the current state of benchmarks and methods used for cultural inclusion in multimodal language models (we organize the papers in \S\ref{lang}, \S\ref{vis}, \S\ref{oth});
\item We provide an overview of common data sources used for creating cultural alignment datasets and how current benchmark creation and culturally relevant fine-tuning dataset creation methodologies leverage these common sources (\S\ref{sec:source}); we also discuss ethical implications and limitations of the dataset creation methodologies (\S\ref{ethical_implications});
\item We provide an overview of the coverage of current datasets for geographical regions and cultures (\S\ref{coverage}) and discuss measures that the community could take to foster equity in cultural inclusion (\S\ref{fut});
\item We also examine the societal impact and implications of deploying LLMs with or without cultural awareness and discuss the role of Human-Computer Interaction (HCI) research in cultural alignment (\S\ref{ethical_implications}).
\end{enumerate}

\textbf{Literature Collection Strategy.} As our paper focuses on multimodal and text-based NLP, we consider papers published in conferences including ACL and regional ACL chapters, EMNLP, ICLR, and ICML, computer vision conferences such as ICCV and CVPR as well as papers published in the ACL Anthology. The inclusion of cultural aspects in the NLP and CV community has been a recent one, with most (benchmark) papers published post-2016, so we consider cultural inclusion benchmarks post-2016. We also consider recent submissions to Arxiv to include recent NLP and social science papers, as the publication cycles for social science journals are typically 1–3 years. For alignment methodologies, we specifically focus on recent works published after 2022, following the release of ChatGPT~\cite{chatgpt}.

We define culture in \S\ref{def_cul} and organize our paper into three major parts. The first part discusses data sources and methodologies the community has used to create datasets and benchmarks for the cultural inclusion of LLMs (\S\ref{sec:source}). The second part discusses the methodologies and state of benchmarks that have been used or created for improving cultural awareness in LLMs across modalities (\S\ref{lang}, \S\ref{vis}, \S\ref{oth}). Finally, we discuss our observations: the state of cultural inclusion (\S\ref{coverage}), ethical issues related to cultural alignment, and the role of cultural alignment in accelerating social science research (\S\ref{ethical_implications}), and future research directions (\S\ref{fut}) in the last part.  In each of the subsections in \S\ref{lang}, \S\ref{vis}, and \S\ref{oth}, we identify specific research gaps and, based on the research gaps, provide concrete suggestions for future research in \S\ref{fut}.
\section{Definitions of Culture and Methodology}
\label{def_cul}

Culture is a complex construct and has been studied in psychology and anthropology with different considerations and assumptions. We adopt \citet{white19}'s view of culture, as they consolidate its definitions from the psychological and anthropological perspectives, distinguishing between human behavior and the study of culture. The psychological perspective considers the study of human behavior as the central part of the analysis and sees culture as an extension of human behavior. One of the main goals of cultural psychology is to study changes in human behavior with respect to culture. The theory and methods in cultural psychology begin with the assumption that psychological processes are socioculturally grounded. On the other hand, the anthropological perspective sees culture as an abstraction of human behavior. The abstraction is necessary to discard unimportant details and focus on actual human interactions, depending on the context. 

Both perspectives are important when considering cultural awareness in LLMs: the anthropological perspective looks at understanding the context and interpreting different elements of tasks based on the context, while the psychological perspective deals with how to process the current information (task and the context) to produce a response. The design of current LLMs is to mimic human behavior as closely as possible without consideration of the context (such as writing a correct summary and seeing how factually correct it is rather than seeing how the summary would be based on the context, considering context would involve considering the knowledge level of the user, and the purpose of the summary). 
The context consists of social factors, which also form an important part of the language, and culture is one of the main components of social factors; for a detailed discussion on modeling social factors of context into NLP systems, we refer the reader to \citet{hovy2021importance}. The two perspectives on culture guide the factors to consider while designing culturally aware LLMs.

From an anthropological perspective, culture is actions, things, and concepts viewed in the context of other actions and things. For instance, going for a vote is just an act in itself; it gains significance when considered in the context of democracy, autocracy, etc. Thus, the definition of culture also considers other humans' behaviors. The locus of culture (or understanding of the cultural context) consists of three dimensions: (1) ``Within humans'' (such as concepts, traditions, beliefs, social practices, etc.), (2) Between ``social interaction among human beings'',
(3) Outside of humans but ``within the patterns of social interaction'' (in materialized objects such as tools, arts, etc.). Point (1) deals with actual actions, things, and concepts (cultural knowledge and morals), while Points (2) and (3) consider the context of the actions and concepts. The human dimension (Point 1) forms the basis for understanding the cultural elements of the task; the other two dimensions are important for generating relevant answers and when LLMs are used as agents. \citet{white19} groups together concepts and actions that form an identity of a culture into a broad category called ``elements of culture''. These elements of culture have been studied in NLP under textual information tasks that are concerned with cultural commonsense knowledge, norms, values, morals, linguistic forms, etc., as well as visio-linguistic parts, such as concepts (and perceptions) associated with various (physical) objects and art forms. Research on cultural psychology has shown that various aspects of visual perception, such as perception of length, geometrical intuition, and depth, vary across people from different cultural backgrounds~\citep{Segall1967TheIO, Jahoda1974PictorialDP}. The variance of perception across cultures, in turn, affects how differences in cultural backgrounds affect the way that individuals attend to, understand, and talk about visual content \citep{Nisbett2003CultureAP}. The variance of perspective necessitates cultural adaptation of images and captions generated by the models. The variance of perspective across cultures has been studied in psychology across five major categories: architecture, clothing, dance and music, food and drink, and religion \citep{Halpern1955TheDE}. As the elements of culture and their perceptions in context vary vastly among cultures, it becomes necessary to study and model the variance in these elements across cultures to create culturally inclusive language technologies.

Concerning LLMs and NLP systems in general, cultural awareness can be thought of as the ability to understand the context in which they are asked to perform a particular task and how the context (and elements of culture in the context) varies with culture (cultural competence). Cultural awareness also includes the ability to understand the variance of cultural elements across different cultures.
Understating the cultural context consists of two things: (a) Recognizing the social context in which a task is performed and (b) based on the context, interpreting different elements of the task. Some tasks are context-sensitive, such as hate-speech detection (the definition of hate-speech varies from culture to culture as norms vary), while others are not (e.g., mathematical reasoning). Some recent works (e.g., \citet{alkhamissi-etal-2024-investigating}) have defined cultural alignment with respect to the model's views aligning with a group of people representing a culture; our definition of cultural awareness encompasses the definition of cultural alignment. So, when designing models for a task (understanding the input and producing an output), the LLM first needs to understand the context (e.g., where it is deployed, what is the end goal, etc.) and then decide if culture needs to be considered for that particular task. To understand the concepts, the LLMs should broadly consider the relationship (for example, how to converse when writing an application letter as a student), social context, and the ``containers'' of communications (such as the setup in which the LLM is deployed, the goal of the LLM) and demographics~\cite{liu2024culturally}. 
The context can also be a design choice while creating or fine-tuning LLM based on deployment goals (e.g., what data to collect and how the LLM will be used). For generalized LLMs, there is a need to have the capability to understand the context in LLMs.  Most efforts up until now have focused on creating context-specific datasets and benchmarks, and less effort has been focused on building and testing LLMs that automatically detect the context. 

Once the LLMs have recognized the context, the understanding of elements of the task and the response depends on cultural knowledge (e.g., the task: ``generate a story with causal conversations happening in Korea'' depends on elements such as LLMs' knowledge about the nature of causal conversations in Korea). We broadly use the term `elements' to include values, meaning of artifacts, and pragmatically motivated features relevant to the task. This raises the question: how do we enable LLMs to generate culturally appropriate responses and understand the cultural elements of a task? Explicitly modeling cultural knowledge in LLMs has been explored as one potential approach. Cultural knowledge consists of various aspects such as norms, morals, values, common sense knowledge, linguistic forms, artifacts, concepts, and meanings associated with artifacts, etc. 

In this survey, we discuss the multiple methods the community has explored to add and evaluate the cultural knowledge in LLMs. The major body of literature focuses on adding cultural knowledge to training/fine-tuning/alignment data and evaluation benchmarks. The data sources for cultural knowledge include direct sources (e.g., sociological surveys, culture bank \citep{shi2024culturebank}, etc.) or indirect sources, which include task-specific datasets in which cultural knowledge is implicitly but deliberately added. We discuss the creation of these sources in \S\ref{sec:source} and summarize task and usage-specific details in \S\ref{lang} for text-only datasets and in \S\ref{vis} and \S\ref{oth} for other modalities-based datasets. Most of the works in NLP look at cultures in isolation while modeling cross-cultural similarities and differences has received less attention \citep{hershcovich-etal-2022-challenges}. The study of cross-lingual similarities and differences has been central to cultural research in anthropology \citep{ember2009cross}. Modeling cross-cultural differences becomes an important aspect to consider while building multicultural datasets, as there is a risk of flattening identities and erasing cultural boundaries if detailed culture-specific data is unavailable (this generally happens for under-represented cultures). Given the progress in creating generalized models, careful consideration should be given to monocultures and sub-cultures within a culture. 
\section{Data Creation Methodology}
\label{sec:source}

\begin{figure*}
\centering
\resizebox{\textwidth}{!}{
\begin{forest}
for tree={
  grow=east,
  reversed=true,
  rectangle,
  draw,
  align=left,
  anchor=west,
  parent anchor=east,
  child anchor=west,
  font=\scriptsize,
}
[\textbf{Creation}\\\textbf{Methodology}
    [\textbf{Example Datasets \& Benchmarks}, tier=leaf, draw=none, no edge]
    [\textbf{Created Manually}, tier=1
        [\textbf{Experts}, tier=2
            [RoCulturaBench~\cite{masala2024vorbe}\\
            TurkishMMLU~\cite{yüksel2024turkishmmlumeasuringmassivemultitask}
            , tier=leaf
            ]
        ]
        [\textbf{Crowdsourced}, tier=2
            [BLEnD~\cite{myung2024blend}\\
            NORMBANK~\cite{ziems2023normbank}\\
            Crossmodal-3600~\cite{thapliyal-etal-2022-crossmodal}\\
            CVQA~\cite{romero2024cvqa}
            , tier=leaf
            ]
        ]
    ]
    [\textbf{Sourced From Web}, tier=1
        [CultureAtlas~\cite{fung2024massively}**\\
        CultureBank~\cite{shi2024culturebank}**\\
        CUNIT~\cite{li2024how}*\\
        Wordscape~\cite{weber2023wordscape}**
        , tier=leaf
        ]
    ]
    [\textbf{Adapted From}, tier=1
        [\textbf{NLP Datasets}, tier=2
            [OMGEval~\cite{liu2024omgeval}*\\
            CulturalRecipes~\cite{cao-etal-2024-cultural}**\\
            KoBBQ~\cite{jin-etal-2024-kobbq}*
            , tier=leaf
            ]
        ]
        [\textbf{CV Datasets}, tier=2
            [Food-500-cap~\cite{Ma2023Food500CA}*\\
            M5~\citep{schneider2024m}*\\
            , tier=leaf
            ]
        ]
        [\textbf{External Sources}, tier=2
            [KMMLU~\cite{son2024kmmlumeasuringmassivemultitask}\\
            WorldValuesBench~\cite{zhao-etal-2024-worldvaluesbench}
            , tier=leaf
            ]
        ]
    ]
    [\textbf{Generated with LLMs}, tier=1
        [CULTURALBENCH-V0.1~\cite{chiu2024culturalteaming}*\\
        CULTURE-GEN~\cite{li2024culture}**\\
        SeeGULL~\cite{jha-etal-2023-seegull}*
        , tier=leaf
        ]
    ]
]
\end{forest}
}
\caption{Overview of the data creation methodologies and example datasets and benchmarks. Datasets and benchmarks created using semi-automatic and fully automatic pipelines are marked with * and **, respectively.}
\label{fig:data-creation}
\end{figure*}
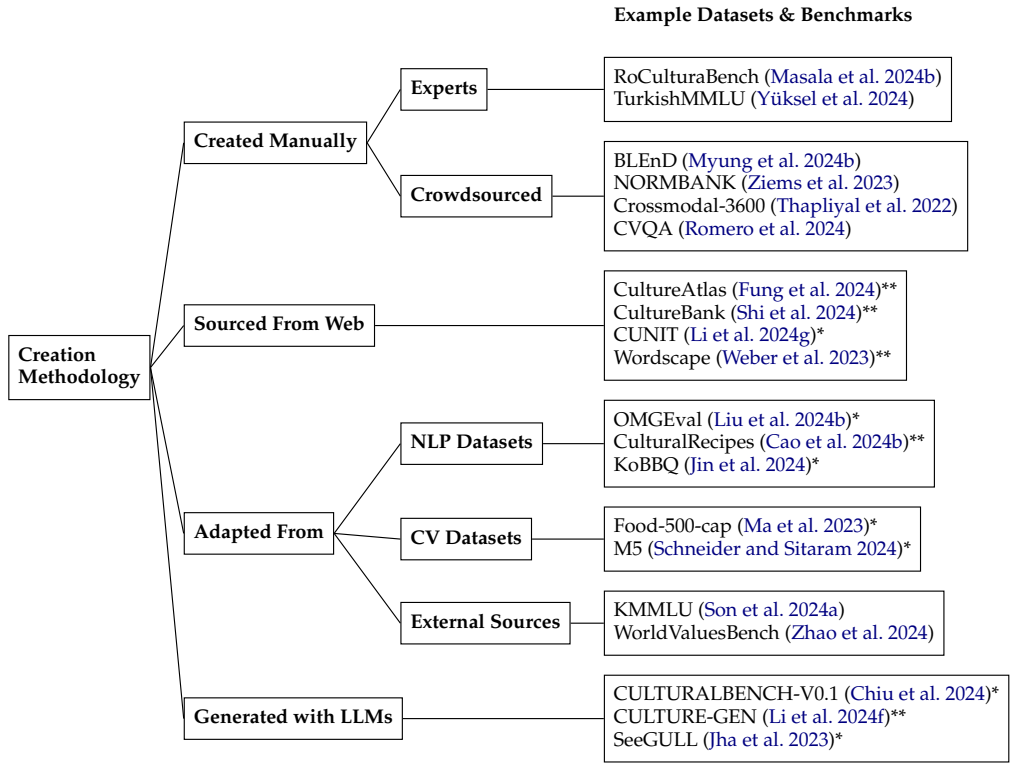

In this section, we examine the data source and creation methodology for culture-specific datasets and benchmarks. The dataset creation methodologies are organized into automatic pipelines (\S\ref{sec:source-auto}), semi-automatic pipelines (\S\ref{sec:source-semi-auto}) and manual creation (\S\ref{sec:source-manual}). Example benchmarks and datasets organized by data resource and dataset creation methodology are listed in Figure \ref{fig:data-creation}.

\subsection{Automatic Pipelines \& Model-in-the-Loop}
\label{sec:source-auto}
Most research focuses on automatic curation to gather cultural knowledge and create training data at scale, especially for pre-training. It primarily relies on publicly available multilingual large-scale corpora such as Wikipedia, CC100 \citep{conneau-etal-2020-unsupervised}, mC4 \citep{xue2021mt5}, and CulturaX \citep{nguyen-etal-2024-culturax}, which are processed raw web text corpora gathered from public web archives. These sources are then cleansed and filtered for specific cultures such as Korean \citep{yoo2024hyperclova}, Irish \citep{tran2024uccix}, Portuguese \citep{pires2023sabia, almeida2024sabi}, Arabic \citep{sengupta2023jais, huang-etal-2024-acegpt, aloui2024101}, Chinese \citep{du2024chinese}, Taiwanese \citep{lin2023taiwan}, Persian \citep{abbasi2023persianllama}, Thai \citep{pipatanakul2023typhoon}, Romanian \citep{masala2024vorbe}, Basque \citep{etxaniz2024bertaqa}, Ukrainian \citep{kiulian-etal-2024-bytes}, Ethiopian \citep{tonja2024ethiollm}, Indonesian \citep{owen2024komodo, cahyawijaya2024cendol}, or even multiple cultures \citep{imanigooghari-etal-2023-glot500, nguyen2023seallms, ustun2024aya}. The refined data is subsequently used to train culture-specific LLMs that are tailored to the knowledge of these cultures.

Current research has advanced by incorporating steps to improve data quality or increase data quantity using model-in-the-loop techniques. StereoKG \citep{deshpande-etal-2022-stereokg} gathers cultural knowledge by mining questions and statements from Reddit and Twitter based on templates, then generates the structured triplets using OpenIE \citep{mausam2016open}. CANDLE \citep{nguyen2023extracting} extracts high-quality cultural commonsense knowledge by building a pipeline to filter cultural text corpora and classify them using fine-tuned models. \citet{saulite-etal-2022-latvian} constructs LNCC from diverse Latvian language resources and automatically annotates them with a uniform morphosyntactic annotation scheme. MANGO \citep{nguyen2024multi} uses LLMs to generate cultural-specific knowledge using seed culture from CANDLE or concept from ConceptNet \citep{speer2017conceptnet}, while CultureAtlas \citep{fung2024massively} collects and processes data from Wikipedia and its sources to form cultural knowledge frames. 

Automatic methods are also used to create instruction data, with some works utilizing LLMs to generate culturally-aligned synthetic data. The \textsc{LLM\_Adapt} dataset from \citet{putri2024can} is created by asking LLM to culturally adapt the English CommonsenseQA dataset~\citep{talmor-etal-2019-commonsenseqa} to Indonesian and Sundanese culture. CultureBank \citep{shi2024culturebank} gathers data from social media platforms and uses models to extract and cluster cultural descriptions. CRAFT \citep{wang2024craft} collects cultural keywords, filters corpus chunks, and generates questions using LLMs. CultureLLM \citep{li2024culturellm} seeds data from the World Values Survey (WVS, \citet{wvs2022}) and augments them by generating semantically equivalent data and identifying replaceable components to be replaced by their synonyms. CulturePark \citep{li2024culturepark} employs a multi-agent framework for cross-cultural conversations based on specific questions initialized from the Pew Global Attitudes Survey (GAS, \citet{pew2022}) and WVS, followed by self-calibration and quality assurance processes. X-Instruction \citep{li-etal-2024-x} is a cross-lingual instruction tuning dataset created with a three-step pipeline that exploits the better generation performance of high-resource languages like English. Specifically, they use OpenAssistant Conversations corpus \citep{NEURIPS2023_949f0f8f} as seed data to generate instructions and use CulturaX \citep{nguyen-etal-2024-culturax} as a multilingual corpus to refine the instructions. In summary, these cases demonstrate that combining LLM-generated synthetic data with culturally relevant sources like surveys and social media is a common strategy for creating instruction datasets. 
Concerning computer vision models, automatic pipelines for data creation involve using geo-localized image datasets such as datasets from photo-sharing platforms such as Flikr, Google Photos, Pinterest \citep{kuznetsova2020open}, Wikimedia, Youtube, etc.; we categorize these sources as extracted using automatic pipelines because the geolocalized tags are already present when the dataset is being considered for cultural adaptation. The captions for the images are obtained from meta-data or using the associated text on Wikipedia \citep{srinivasan2021wit, weber2023wordscape}. The community has mostly used automatic pipelines to get the images, as captions obtained from metadata or from Wikipedia may not be reliable for creating benchmarks or datasets for cultural adaptation.

\subsection{Semi-Automatic: Human-in-the-Loop}
\label{sec:source-semi-auto}
Semi-automatic approaches combine human expert knowledge with machine processing to produce higher-quality datasets. Several recent works exemplify this methodology. For instance, the \textsc{LLM\_Gen} dataset introduced by \citet{putri2024can} begins with categories and concepts manually created by human annotators, which are then used to generate a commonsense question-answering dataset. Similarly, \citet{bai2024coig} developed COIG-CQIA, which relies on manually curated sources vetted by human experts to ensure the quality of Chinese instruction data before applying machine-based filtering and cleaning processes. \citet{alyafeai-etal-2024-cidar} developed the Arabic instruction-tuning dataset CIDAR by translating the AlpaGasus~\cite{chen2024alpagasus} dataset using ChatGPT~\cite{chatgpt}. They then go through manual cultural localization and review linguistic issues with native Arabic speakers. The STREAM framework \citep{wang2024stream} provides an even higher degree of human intervention. In this approach, human annotators first provide moral values to guide Large Language Models (LLMs) in generating scenarios consisting of situations and actions. The generated results are then manually screened by human annotators. Subsequently, human annotators and LLMs are tasked with evaluating randomly selected scenarios, which are used to measure the alignment of LLMs with human judgments. These examples illustrate that semi-automatic approaches are a moderate strategy, balancing quality and quantity by integrating human expertise with scalable machine processing.

For vision-language models, semi-automatic and human-in-the-loop have been the most common way of creating benchmarks and datasets for cultural adaptation. One of the methods includes using geolocalized images from photo-sharing platforms such as Flikr, Google Photos, Pinterest \citep{kuznetsova2020open}, Wikimedia, and Youtube, and then using local annotators to create a variety of datasets \citep{thapliyal-etal-2022-crossmodal, yin2021broaden}. Some works also use pre-existing computer vision benchmarks, such as ISIA Food-500 \citep{min2020isia}, Dollar Street \citep{NEURIPS2023_5fc47800}, etc., and refine the captions (or related information such as questions in VQA) to include cultural information \citep{Ma2023Food500CA, schneider2024m}.

\subsection{Manual: Handcrafted from Scratch}
\label{sec:source-manual}
Manually crafted datasets, created from scratch by human experts and annotators, remain the gold standard regarding quality and alignment with human values. These datasets typically involve human participants creating instruction-response pairs that reflect common usage patterns, providing feedback on responses to given prompts, or contributing text in specific languages to build corpora. The nature of this process ensures high-quality outputs that are well-aligned with human expectations and linguistic nuances. However, the manual approach has limitations, particularly in terms of scalability. The resources required for human-generated datasets, including time and financial investments, often result in smaller datasets than those produced by automatic or semi-automatic methods. Despite these constraints, the value of handcrafted datasets is evident in various applications within natural language processing.

For instance, \citet{putri2024can} developed \textsc{Human\_Gen} dataset entirely from scratch, encompassing everything from category and concept ideation to constructing commonsense question-answering data. The quality of this dataset was maintained through rigorous evaluation by a group of annotators. Manual creation is also important when cultural knowledge is not explicitly documented. \citet{myung2024blend} aims to capture everyday mundane knowledge often not documented online. Thus, the dataset was manually created by recruiting native annotators through a crowdsourcing platform. In cases where the pool of annotators is limited, such as with North Korean annotators, they directly recruited participants without relying on crowdsourcing platforms. Similarly, in the case of low-resource languages, \citet{le2023parallel} manually created a parallel corpus for Central and Northern Vietnamese dialects with native dialect speakers. Furthermore, manual annotation is crucial for subjective tasks such as cultural-specific hate speech detection~\cite{jeong-etal-2022-kold} or inspiring content detection~\cite{ignat2024crossculturalinspirationdetectionanalysis}.

Several language models have leveraged handcrafted datasets to enhance their performance and cultural relevance. HyperClovaX \citep{yoo2024hyperclova} utilized high-quality human-annotated Korean datasets for instruction tuning following pretraining. TaiwanLLM \citep{lin2023taiwan} incorporated human instructions, multi-turn dialogues, and human feedback based on real user interactions that encompass Chinese cultural knowledge. In the Arabic language space, both AceGPT \citep{huang-etal-2024-acegpt} and Jais \citep{sengupta2023jais} augmented their training data with native Arabic instructions as part of their supervised fine-tuning process. The value of handcrafted datasets is particularly pronounced in low-resource language settings. Models like Komodo \citep{owen2024komodo} and Aya \citep{ustun2024aya} exemplify this approach, covering many languages, including some with extremely limited language resources. Komodo collaborated with local language experts to collect data for various local languages, while Aya extended this approach to 101 languages. These examples highlight that while leveraging handcrafted datasets has become a popular strategy to enhance cultural relevance and performance, there remains a significant research gap in systematically developing high-quality datasets for underrepresented languages, pointing to a future direction of broader dataset creation efforts across diverse linguistic and cultural contexts.

For vision-language data, manual data collection methods include starting with an initial list of questions and concepts and asking the annotators to search relevant images on the internet~\citep{Baek2024EvaluatingVA, Wang2024CVLUEAN} followed by processing of images to create various types of benchmarks (e.g., Captioning, VQA, Image retrieval, etc.). Native annotators can also drive the cultural topics and objects to prioritize objects and concepts with significant cross-cultural differences \citep{Wang2024CVLUEAN}. Most of the video and speech datasets are also created manually, where annotators annotate emotions on curated videos or web series ~\cite{amiriparian2024exhubert, zhao-etal-2022-m3ed}. 
There have been a few examples in the literature where annotators are asked to click the relevant photos based on initial concepts \citep{romero2024cvqa}. Searching for culture-related photos can be limited since only aesthetically pleasing images are often uploaded to the internet, leading to a lack of photos with everyday objects and common sense knowledge. Also, because not all cultures have a significant online presence, it could possibly be discriminative towards certain cultures \citep{liu2022not}.
In the upcoming sections, we discuss how data-creation methodologies discussed in this section have been used for creating task-specific data and cultural alignment of language and vision models.

\section{Language Models and Culture}
\label{lang}

There has been a growing recognition of the cultural biases, stereotypes, and lack of diverse cultural knowledge present in LLMs~\cite{hershcovich2022challenges, 10.1145/3597307}. Those issues directly lead to problems, particularly in applications like dialogue systems, where LLMs may overlook users’ cultural backgrounds, potentially leading to inaccurate information or the reinforcement of cultural stereotypes. To address these limitations and make LLMs more culturally inclusive, two key approaches have emerged: a) pre-training and fine-tuning models with culturally relevant data, and b) employing prompt-based methods that do not require retraining. Section \ref{sec:lang-align} provides a detailed explanation of these methods, focusing on how they aim to enhance LLMs' cultural adaptability. Furthermore, there is increasing attention toward developing benchmarks and evaluation frameworks that measure how well LLMs align with diverse cultural contexts. Section \ref{eval} elaborates on these benchmarks and evaluation frameworks. However, both alignment methodologies and evaluation techniques remain fragmented, with no universally established standards.





\subsection{Cultural Alignment: Methodologies and Goals}
\label{sec:lang-align}
Cultural alignment refers to the process of aligning an AI system with the set of shared beliefs, values, and norms of the group of users that interact with the system, as defined by \citet{masoud2023cultural} based on the foundational works of \citet{Hofstede2010-ah} and \citet{bennett1994cultural}. The importance of cultural alignment was demonstrated by \citet{masoud2023cultural} through their Cultural Alignment Test (Hofstede's CAT), which revealed that current LLMs struggle to fully comprehend cultural values. Their research suggested that this limitation could be addressed through fine-tuning models with culture-specific language. Complementary studies by \citet{li2024culture} and \citet{tao2024cultural} reached similar conclusions, though their findings emphasized the effectiveness of prompting techniques for achieving cultural alignment. Given the importance of these findings, we examine the current state of cultural alignment in AI systems through two distinct perspectives. First, we analyze the methodologies for achieving cultural alignment, focusing on two primary approaches: model training and prompting techniques. Second, we explore how different cultural objectives influence and shape alignment efforts.

\begin{figure*}
\centering
\resizebox{\textwidth}{!}{
\begin{forest}
for tree={
  grow=east,
  reversed=true,
  rectangle,
  draw,
  align=left,
  anchor=west,
  parent anchor=east,
  child anchor=west,
  font=\scriptsize,
}
[\textbf{Cultural}\\\textbf{Alignment}\\\textbf{Methdologies}
        [\textbf{Training-based}, tier=1
            [\textbf{Pre-training}, tier=2
                [\textbf{Pre-training from Scratch}\\
                HyperClovaX~\cite{yoo2024hyperclova}*\\
                PersianLLaMA~\cite{abbasi2023persianllama}*\\
                JASMINE~\cite{billah-nagoudi-etal-2023-jasmine}*, tier=leaf]
                [\textbf{Continue Pre-training}\\
                UCCIX~\cite{tran2024uccix}*\\
                SeaLLM~\cite{nguyen2023seallms}*\\
                TaiwanLLM~\cite{lin2023taiwan}*\\
                Komodo~\cite{owen2024komodo}*\\
                Typhoon~\cite{pipatanakul2023typhoon}*\\
                Sabiá~\cite{pires2023sabia}*\\
                AceGPT~\cite{huang-etal-2024-acegpt}*\\
                Jais~\cite{sengupta2023jais}*\\
                EthioLLM~\cite{tonja2024ethiollm}*\\
                RoLLM~\citet{masala2024vorbecstiromanecsterecipetrain}*\\
                \citet{etxaniz2024bertaqa}, tier=leaf]
            ]
            [\textbf{Supervised}\\\textbf{Fine-tuning}, tier=2
                [\textbf{Instruction-tuning}\\
                Cendol~\cite{cahyawijaya2024cendol}*\\
                \citet{bai2024coig}\\
                \citet{wang2024craft}\\
                \citet{li2024culturepark}\\
                \citet{li2024culturellm}\\
                \citet{shi2024culturebank}\\
                \citet{zhang2024methodology}\\
                \citet{bhatia-shwartz-2023-gd}, tier=leaf]
                [\textbf{Task-specific}\\
                Offensive Language Detection:~\citet{zhou2023cultural}\\
                Emotion Analysis:~\citet{kim-etal-2024-moral}\\
                Ethical Judgment:~\citet{shen-etal-2022-social}\\
                Hate Speech Detection:~\citet{dehghan-yanikoglu-2024-multi};\\
                \citet{singh-thakur-2024-generalizable}, tier=leaf]
            ]
            [\textbf{Others}, tier=2
                [Modular Pluralism:~\citet{feng2024modular}\\
                Unsupervised Learning:~\citet{li-etal-2024-unsupervised}\\
                Preference-tuning:~\citet{jinnai-2024-cross}, tier=leaf]
            ]
        ]
        [\textbf{Training-free}, tier=1
            [\textbf{Anthropological}\\\textbf{Prompting}, tier=2
                [\citet{alkhamissi-etal-2024-investigating}, tier=leaf]
            ]
            [\textbf{Cultural}\\\textbf{Prompting}, tier=2
                [\citet{tao2024cultural}, tier=leaf]
            ]
            [\textbf{Sociodemographic}\\\textbf{Prompting}, tier=2
                [\citet{li2024culture}\\
                \citet{deshpande-etal-2023-toxicity}\\
                \citet{pmlr-v202-santurkar23a}\\
                \citet{hwang-etal-2023-aligning}\\
                \citet{cheng-etal-2023-marked}\\
                \citet{zhou2024does}\\
                \citet{shen-etal-2024-understanding}, tier=leaf]
            ]
        ]
    [\textbf{Goal-specific}, tier=1
        [
        Content Moderation:~\citet{chan2023harmonizing}\\
        Dataset Construction:~\citet{hasan2024nativqa}\\
        Response Diversity:~\citet{lahoti2023improving}\\
        \citet{hayati2023far}\\
        Bias Mitigation:~\citet{lee2023kosbi};\\
        \citet{zhao-etal-2023-chbias};\\
        \citet{narayan2024bias};\\
        \citet{khandelwal2024indian}, tier=leaf]
    ]
]
\end{forest}
}
\caption{Cultural alignment methodologies for language models based on methodologies and goals. Model names are marked with *.}
\label{fig:cultural-alignment}
\end{figure*}
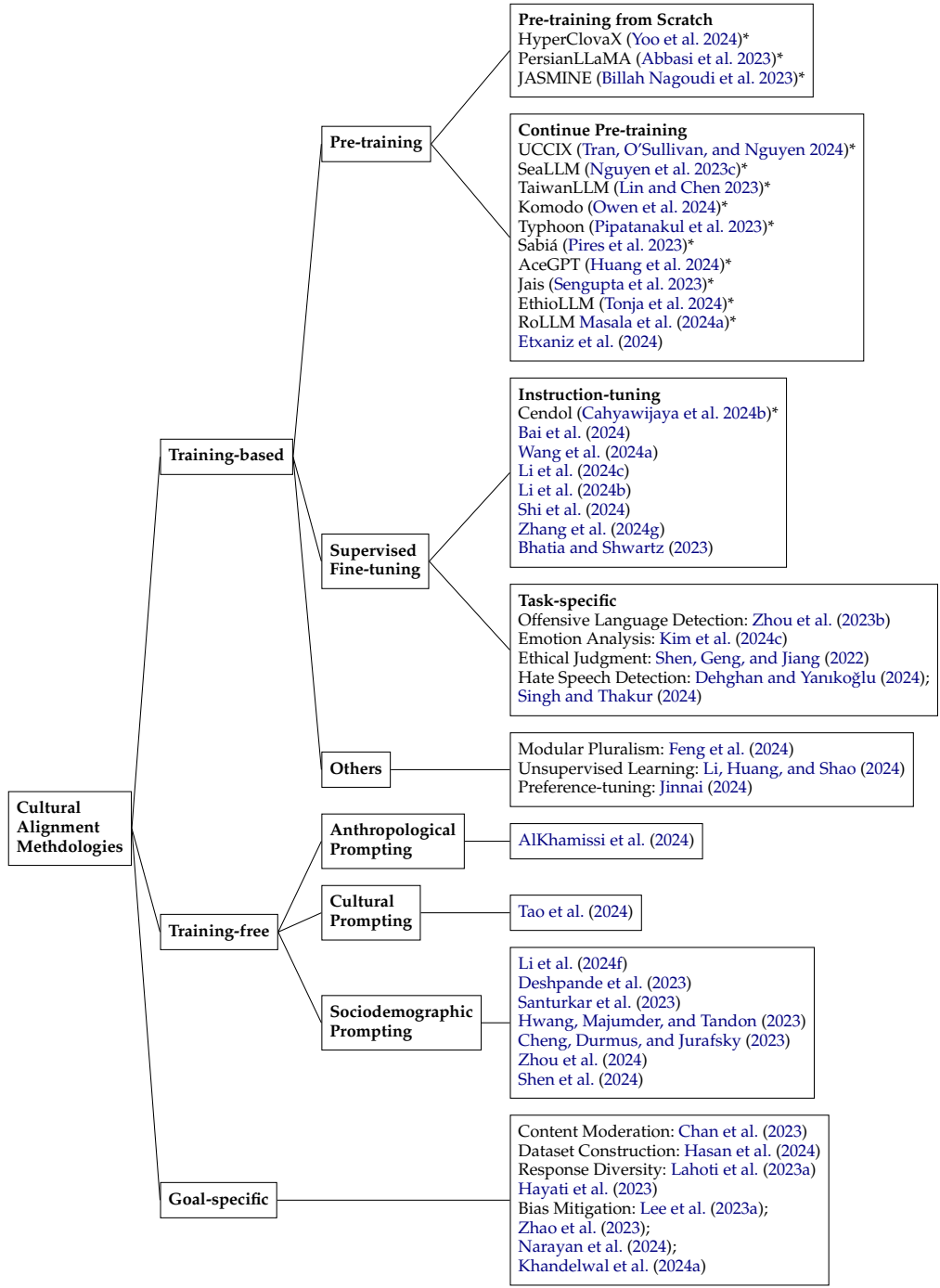


In this section, we discuss the methodologies used for cultural alignment. In general, cultural alignment can be done through two approaches: training-based (\S\ref{sec:lang-align-train}) and training-free (\S\ref{sec:lang-align-prompt}). In addition, there are goal-based alignment methods for specific goals, such as content moderation (\S\ref{sec:lang-align-goal}). The papers are organized in Figure \ref{fig:cultural-alignment}.

\subsubsection{Training-Based Methods}
\label{sec:lang-align-train}
Training a language model is one way to achieve cultural alignment. The key differentiating factor in this approach lies in the training data, which must contain culturally relevant knowledge, norms, and values specific to the target culture, as previously discussed in section \ref{sec:source}. 

The training of language models for cultural alignment can be broadly categorized into two main approaches: pre-training and fine-tuning. Pre-training is the step where we train the model through a large corpus to learn the general features of the data, which, for the purpose of cultural alignment, includes culture-specific knowledge, norms, and values obtained inside. Pre-training can be further categorized into two strategies: initiating pre-training from scratch using culturally relevant data, or continuing from an existing pre-trained LLM. Pre-training the model from scratch is expensive, as a result of training the whole model parameters and the large size of the data. Therefore, not a lot of cultural alignment is done with this method, as only HyperClovaX \citep{yoo2024hyperclova}, PersianLLaMA \citep{abbasi2023persianllama}, and JASMINE \citep{billah-nagoudi-etal-2023-jasmine} have done the pre-training from scratch. 

Continued pre-training is another way of pre-training, which involves taking an existing pre-trained model and training it further on culturally relevant data. This method has two key advantages: it avoids the computational expense of pre-training from scratch, and it requires only raw text data rather than the labeled datasets needed for supervised fine-tuning. There are a lot of works that incorporate continued pre-training as part of their cultural alignment effort as shown by a lot of culture-specific LLM \citep{tran2024uccix, nguyen2023seallms, lin2023taiwan, owen2024komodo, pipatanakul2023typhoon, pires2023sabia, huang-etal-2024-acegpt, sengupta2023jais, tonja2024ethiollm, masala2024vorbecstiromanecsterecipetrain, etxaniz2024bertaqa} using continued pre-training.

Fine-tuning for cultural alignment involves further training a pre-trained model using culturally relevant labeled datasets. Unlike continued pre-training, which uses raw text data, fine-tuning utilizes data specifically labeled for the intended task. This approach can be applied to task-specific objectives such as hate-speech detection and emotion classification or to general-purpose applications like instruction-following and conversational abilities. Instruction-tuning, a specific form of fine-tuning that uses instruction-response pairs, has been widely adopted by researchers developing culture-specific LLMs \citep{yoo2024hyperclova, lin2023taiwan, owen2024komodo, huang-etal-2024-acegpt, cahyawijaya2024cendol, sengupta2023jais, masala2024vorbecstiromanecsterecipetrain, nguyen2023seallms, bai2024coig}. Another work that leverages instruction-tuning is from \citet{zhang2024methodology}, which proposes a rapid adaptation method for large models in specific cultural contexts based on specific cultural knowledge and safety values data. Recent research \citep{li2024culturepark, li2024culturellm, wang2024craft, shi2024culturebank} demonstrates that instruction-tuning enables models to effectively reason across multiple cultures in conversations. Additionally, \citet{bhatia-shwartz-2023-gd} showed that fine-tuning on CANDLE data \citep{10.1145/3543507.3583535} allows models to both capture and generate culturally nuanced commonsense knowledge. 

Many works also show a positive impact on various cultural alignment applications by doing task-specific fine-tuning. For offensive language detection, \citet{zhou2023cultural} achieved effective results by fine-tuning models on cultural value survey data. In hate speech detection, which is deeply intertwined with cultural context, researchers have employed diverse approaches. \citet{dehghan-yanikoglu-2024-multi} apply dual-contrastive learning when fine-tuning and incorporating paralinguistic features such as emoji, while \citet{singh-thakur-2024-generalizable} use a federated approach that utilizes continuous adaptation and fine-tuning to detect hate speech that is highly affected by cultural nuances. In emotion analysis, \citet{kim-etal-2024-moral} achieved promising results in moral emotions classification, where the model utilizes information on moral emotions embedded in the data and can perceive different emotions for different cultures. For ethical judgment, \citet{shen-etal-2022-social} involved grounding complex narrative situations with social norms using a pre-trained encoder-decoder and integrating these norms with a classification model.

Beyond pre-training and fine-tuning, some innovative approaches offer unique perspectives on training-based alignment. One such approach is Modular Pluralism \citep{feng2024modular}, which employs smaller language models alongside larger ones to guide them in incorporating cultural knowledge and values into their responses according to the given cultural context. From an unsupervised perspective, \citet{li-etal-2024-unsupervised} utilizes an adaptive context-aware unsupervised learning framework to convert between traditional and simplified Chinese characters, which is an important aspect of understanding Chinese culture. There is also a study by \citet{jinnai-2024-cross} that uses preference-tuning through Direct Preference Optimization (DPO) \citep{rafailov2024direct} instead of fine-tuning to investigate how cross-cultural alignment affects an LLM's commonsense morality. 

Several works have investigated the actual impact of training for cultural alignment. \citet{mukherjee-etal-2024-global} found that while there are improvements in terms of cultural competence, they still fall short, particularly in non-western contexts. They highlight the need to incorporate more than the target language during the training process. \citet{ladhak-etal-2023-pre} find that while pre-training can make the model aligned with the specific culture, the resulting model can possess bias contained in the pre-training data. They also find that fine-tuning with smaller parameters, such as adapter-fine-tuning techniques like LoRA, provides better generalization and debiasing rather than training the entire model. \citet{choenni-etal-2024-echoes} investigate how cultural value shifts during fine-tuning, and find that language has a minor role in cultural shifts and positively affects alignment with human values, but it varies considerably across languages.

\subsubsection{Training-Free Methods}
\label{sec:lang-align-prompt}
Cultural alignment in language models can be achieved without additional training, primarily through prompting techniques. Research by \citet{alkhamissi-etal-2024-investigating, zhou2024does, arora-etal-2023-probing} demonstrates that cultural alignment is influenced by the training data and the prompts used during inference. \citet{alkhamissi-etal-2024-investigating} observed that models exhibit stronger cultural alignment when prompted in a culture-specific language. Building on this insight, they introduced anthropological prompting, incorporating anthropological reasoning aspects into the prompt to enhance cultural alignment. Another promising approach is the Collective, Critique, and Self-Voting (CCSV) method, which is proposed by \citet{lahoti-etal-2023-improving}. Their findings suggest that language models can comprehend the concept of diversity and are capable of reasoning about and critiquing their responses to improve cultural diversity in their outputs. \citet{tao2024cultural} also propose a prompt methodology called cultural prompting, which instructs the language model to answer like a person from another society. They investigated the method by comparing the model responses to nationally representative survey data and found cultural prompting works quite well to increase the alignment of the model with the nationally representative survey data.

Sociodemographic prompting has gained widespread attention among researchers for cultural alignment \citep{deshpande-etal-2023-toxicity, pmlr-v202-santurkar23a, hwang-etal-2023-aligning, cheng-etal-2023-marked, zhou2024does, li2024culture, shen-etal-2024-understanding}. This method involves enriching prompts with sociodemographic information or cultural context, expecting that the model's output will align with the information given in the prompt. Sociodemographic prompting has shown potential in applications such as data augmentation \citep{hartvigsen-etal-2022-toxigen} and social computing simulation \citep{10.1145/3526113.3545616}. However, concerns have been raised regarding the robustness of sociodemographic prompting. \citet{mukherjee2024culturalconditioningplaceboeffectiveness} found that most models exhibit similar variations in response to culturally conditioned cues as they do to non-cultural ones, particularly in terms of eliciting cultural bias. Similarly, \citet{beck-etal-2024-sensitivity} observed that model outcomes vary significantly across different model types, sizes, and datasets. These findings suggest that sociodemographic prompting should be employed cautiously, especially in sensitive applications.

\subsubsection{Goal-Specific Alignment Strategies}
\label{sec:lang-align-goal}
In this section, we discuss works that have been done specifically to test and improve cultural alignment in language models for particular goals.



\citet{chan2023harmonizing} train large language models on extensive datasets of media news and articles to create culturally attuned models for content moderation; the goal is to capture the nuances of communication and offensive content across cultures. \citet{lahoti2023improving} propose metrics to measure diversity in LLM-generated responses along people and culture axes and propose a new prompting technique to self-improve people diversity of LLMs. On similar lines, \citet{hayati2023far} propose a step-by-step recall prompting-based method to increase the diversity of responses (with cultural diversity increase being one of the outcomes). \citet{lee2023kosbi} provide a (fine-tuning) dataset specific to Korean culture for mitigating social bias in generated content. \citet{zhou2023cultural} studies the importance of cultural features in determining the success of transfer learning in the case of offensive language detection. \citet{khandelwal2024indian} provide a dataset of Indian stereotypes and anti-stereotypes and propose interventions to reduce both stereotypical and anti-stereotypical biases in language models, thereby aligning them with Indian Culture. \citet{narayan2024bias} proposes a framework to quantify and mitigate biases within LLMs by creating a new metric that detects, measures, and mitigates racial and cultural biases in LLMs without reliance on demographic annotations. \citet{hasan2024nativqa} propose a language-independent framework to construct culturally and regionally aligned QA datasets in native languages for LLM evaluation and demonstrate the efficacy of the framework by designing a multilingual natural QA dataset, MultiNativQA, consisting of around 64k manually annotated QA pairs in seven languages, ranging from high to extremely low resources, based on queries from native speakers from 9 regions covering 18 topics. \citet{zhao-etal-2023-chbias} introduce CHBias, a dataset for bias evaluation and mitigation of Chinese conversational language models with culture-specific biases.

\textbf{\textsl{Takeaways from \S\ref{sec:lang-align}.}} Various alignment methods, such as continued pretraining, fine-tuning, instruction tuning, and prompt tuning, are used to create more culturally-aware LLMs. However, most efforts are concentrated on aligning LLMs with individual local cultures, with limited research dedicated to developing cross-cultural LLMs that encompass comprehensive knowledge of multiple cultures. More work is needed in this area to enhance cross-cultural understanding in LLMs.

\subsection{Benchmarks and Evaluation}

\label{eval}

\begin{figure*}
\centering
\resizebox{\textwidth}{!}{
\begin{forest}
for tree={
  font=\scriptsize,
  rectangle,
  draw,
  align=center,
  edge path={\noexpand\path[\forestoption{edge}] (\forestOve{\forestove{@parent}}{name}.parent anchor) -- +(0,-12pt)-| (\forestove{name}.child anchor)\forestoption{edge label};}
}
[NLP Evaluation
    [
    Academic\\Knowledge\\(\S\ref{sec:academic_knowledge})]
    [
    Commonsense\\Knowledge\\(\S\ref{sec:commonsense_knowledge})]
    [
    Social\\Values\\(\S\ref{sec:value})]
    [
    Social Norms\\and Morals\\(\S\ref{sec:norms_morals})]
    [
    Social Bias\\and Stereotype\\(\S\ref{sec:bias})]
    [
    Toxicity\\and Safety\\(\S\ref{sec:toxic})]
    [
    Emotional and\\Subjective Topics\\(\S\ref{sec:subjective})]
    [
    Linguistics\\(\S\ref{sec:linguistics})]
]
\end{forest}
}
\caption{An overview of domains of text-based culturally-aware benchmarks}
\label{fig:nlp_eval}
\end{figure*}
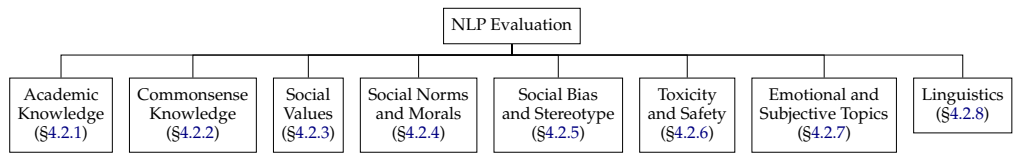

In this section, we provide an overview of various benchmarks designed to assess cultural elements through text-based tasks. The cultural elements are categorized into eight domains as specified in Figure \ref{fig:nlp_eval}.

The Academic Knowledge section (\S\ref{sec:academic_knowledge}) focuses on evaluating knowledge sourced from human educational materials. The Commonsense Knowledge section (\S\ref{sec:commonsense_knowledge}) covers diverse datasets and benchmarks that assess general cultural knowledge, such as food, family, holidays, sports, and entertainment. In the Social Values section (\S\ref{sec:value}), social science studies are used to evaluate LLMs’ alignment with human social values. The Social Norms and Morals section (\S\ref{sec:norms_morals}) examines specific cultural norms and morals, exploring how these values shift depending on the social context. In the Social Bias and Stereotypes section (\S\ref{sec:bias}), the focus is on adapting bias benchmarks to local languages and cultures, expanding to cross-cultural perspectives. The Toxicity and Safety section (\S\ref{sec:toxic}) addresses offensive and hate speech detection in local languages and cultures. The Emotional and Subjective Topics section (\S\ref{sec:subjective}) explores psychological cultural difference including emotion prediction, sentiment analysis and subjective topic classification. Lastly, the Linguistics section (\S\ref{sec:linguistics}) delves into how culture is reflected in language, the ways language varieties and literary forms embody cultural elements, and how translation and dialogue systems can become more culturally aware.

Each cultural element is evaluated through element-specific approaches. For instance, commonsense knowledge is typically assessed using multiple-choice questions (MCQ) or short-answer questions that require cultural knowledge. Meanwhile, social values are often examined using sociological surveys like the World Values Survey (WVS) to test cross-cultural differences in LLMs' understanding of social values.

\subsubsection{Academic Knowledge}
\label{sec:academic_knowledge}

\begin{figure*}
\centering
\resizebox{\textwidth}{!}{
\begin{forest}
for tree={
  grow=east,
  reversed=true,
  rectangle,
  draw,
  align=left,
  anchor=west,
  parent anchor=east,
  child anchor=west,
  font=\scriptsize,
}
[\textbf{Academic}\\\textbf{Knowledge}
    [\textbf{MMLU}\\\textbf{Series}, tier=1
        [ArabicMMLU~\cite{koto-etal-2024-arabicmmlu}; CMMLU~\cite{li-etal-2024-cmmlu}; IndoMMLU~\cite{koto-etal-2023-large}; \\
        JMMLU~\cite{yin2024should}; KMMLU~\cite{son2024kmmlumeasuringmassivemultitask}; TurkishMMLU~\cite{yüksel2024turkishmmlumeasuringmassivemultitask}; \\
        PersianMMLU~\cite{ghahroodi2024khayyam}
        , tier=leaf
        ]
    ]
    [\textbf{Medical}\\\textbf{Knowledge}, tier=1
        [K-NLEKMD~\cite{jang2023gpt}; CMB~\cite{wang2024cmb}
        , tier=leaf
        ]
    ]
    [\textbf{Others}, tier=1
        [KorNAT~\cite{lee2024kornat}; INVALSI~\cite{mercorio2024disce}; FoundaBench~\cite{li2024foundabench}; \\
        M3Exam~\cite{zhang2023m3exam}
        , tier=leaf
        ]
    ]
]
\end{forest}
}
\caption{Academic Knowledge Evaluation Benchmarks}
\label{fig:academic}
\end{figure*}
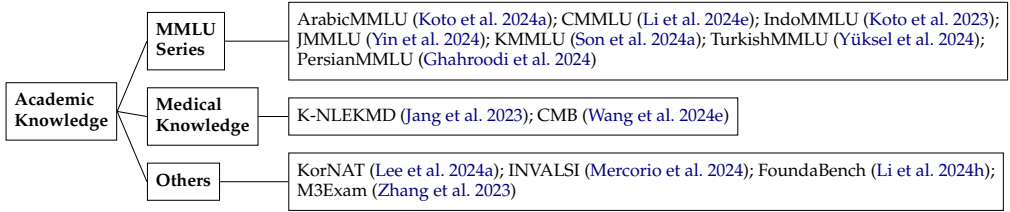

Human educational resources such as exam questions or textbooks are being utilized to assess language understanding and general knowledge capability of LLMs. For instance, the MMLU dataset~\cite{hendrycks2021measuring} is sourced from practice exam questions, such as Graduate Record Examination (GRE). This dataset is commonly used to evaluate LLMs’ language understanding and problem-solving abilities across various domains, including STEM, humanities, social science. Among these academic domains, fields such as history, law and literature in particular, often require knowledge specific to certain region. Thus, benchmarks have been developed from local educational materials to evaluate regional knowledge. The overall hierarchy of the papers in this section is specified in Figure~\ref{fig:academic}.

\begin{table}[]
\resizebox{\columnwidth}{!}{%
\begin{tabular}{@{}llllllrr@{}}
\toprule
 &
  \textbf{Languages} &
  \textbf{\begin{tabular}[c]{@{}l@{}}Evaluation\\method\end{tabular}} &
  \textbf{Creation method} &
  \textbf{Educational stages} &
  \textbf{Domains} &
  \textbf{\begin{tabular}[c]{@{}r@{}}Size(k)\end{tabular}} &
  \textbf{\begin{tabular}[c]{@{}r@{}}Cultural\\question\\ratio(\%)\end{tabular}} \\ \midrule
  \begin{tabular}[c]{@{}l@{}}\textbf{MMLU}\\\cite{hendrycks2021measuring}\end{tabular} &
  English &
  MCQ &
  Manually created &
  \begin{tabular}[c]{@{}l@{}}Elementary\\ High school\\ College\\ Professional\end{tabular} &
  \begin{tabular}[c]{@{}l@{}}Humanities\\ Social Science\\ STEM\end{tabular} &
  15.9 &
  - \\ \midrule
\begin{tabular}[c]{@{}l@{}}\textbf{ArabicMMLU}\\\cite{koto-etal-2024-arabicmmlu}\end{tabular} &
  \begin{tabular}[c]{@{}l@{}}Modern\\ Standard\\ Arabic\end{tabular} &
  MCQ &
  Manually created &
  \begin{tabular}[c]{@{}l@{}}Primary school\\ Middle school\\ High school\\ University\\ Professional\end{tabular} &
  \begin{tabular}[c]{@{}l@{}}Humanities\\ Social Science\\ STEM\\ Language\end{tabular} &
  14.5 &
  57.7 \\ \midrule
\begin{tabular}[c]{@{}l@{}}\textbf{CMMLU}\\\cite{li-etal-2024-cmmlu}\end{tabular} &
  \begin{tabular}[c]{@{}l@{}}Mandarin\\ Chinese\end{tabular} &
  MCQ &
  Manually created &
  \begin{tabular}[c]{@{}l@{}}Primary school\\ Middle/high school\\ College\\ Professional\end{tabular} &
  \begin{tabular}[c]{@{}l@{}}Humanities\\ Social Science\\ STEM\end{tabular} &
  11.5 &
  $\sim$25.3 \\ \midrule
\begin{tabular}[c]{@{}l@{}}\textbf{IndoMMLU}\\\cite{koto-etal-2023-large}\end{tabular} &
  \begin{tabular}[c]{@{}l@{}}Indonesian and\\ local languages\end{tabular} &
  MCQ &
  \begin{tabular}[c]{@{}l@{}}Manually created\\ by experts\end{tabular} &
  \begin{tabular}[c]{@{}l@{}}Primary school\\ Junior high school\\ Senior high school\\ University\end{tabular} &
  \begin{tabular}[c]{@{}l@{}}Humanities\\ Social Science\\ STEM\\ Indonesian Language\\ Local Languages\\Local Cultures\end{tabular} &
  14.9 &
  46 \\ \midrule
\begin{tabular}[c]{@{}l@{}}\textbf{JMMLU}\\\cite{yin2024should}\end{tabular} &
  Japanese &
  MCQ &
  \begin{tabular}[c]{@{}l@{}}Adapted from MMLU\\and manually created\\ by experts\end{tabular} &
  \begin{tabular}[c]{@{}l@{}}Elementary\\ High school\\ College\\ Professional\end{tabular} &
  \begin{tabular}[c]{@{}l@{}}Humanities\\ Social Science\\ STEM\end{tabular} &
  7.5 &
  - \\ \midrule
\begin{tabular}[c]{@{}l@{}}\textbf{KMMLU}\\\cite{son2024kmmlumeasuringmassivemultitask}\end{tabular} &
  Korean &
  MCQ &
  \begin{tabular}[c]{@{}l@{}}Automatically\\ extracted\end{tabular} &
  Expert &
  \begin{tabular}[c]{@{}l@{}}Humanities\\ Social Science\\ STEM\\ Applied Science\end{tabular} &
  35 &
  20.4 \\ \midrule
\begin{tabular}[c]{@{}l@{}}\textbf{TurkishMMLU}\\\cite{yüksel2024turkishmmlumeasuringmassivemultitask}\end{tabular} &
  Turkish &
  MCQ &
  \begin{tabular}[c]{@{}l@{}}Manually created\\ by experts\end{tabular} &
  High school &
  \begin{tabular}[c]{@{}l@{}}Humanities\\ Social Science\\ Math\\ Natural Sciences\\ Language\end{tabular} &
  10 &
  - \\ \midrule
\begin{tabular}[c]{@{}l@{}}\textbf{PersianMMLU}\\\cite{ghahroodi2024khayyam}\end{tabular} &
  Persian &
  MCQ &
  \begin{tabular}[c]{@{}l@{}}Automatically\\ extracted\end{tabular} &
  \begin{tabular}[c]{@{}l@{}}Lower primary school,\\ Upper primary school,\\ Lower secondary school,\\ Upper secondary school\end{tabular} &
  \begin{tabular}[c]{@{}l@{}}Humanities\\ Social Science\\ Natural Science\\ Mathematics\end{tabular} &
  20 &
  - \\ \bottomrule
\end{tabular}%
}
\caption{Details of MMLU-series benchmarks.}
\label{tab:mmlu-series}
\end{table}

One of the shortcomings of the MMLU dataset is that it primarily focuses on knowledge related to the United States. Addressing this, the dataset has been adapted into several linguistically and culturally specific benchmarks, including ArabicMMLU~\cite{koto-etal-2024-arabicmmlu}, CMMLU~\cite{li-etal-2024-cmmlu}, IndoMMLU~\cite{koto-etal-2023-large}, JapaneseMMLU (JMMLU~\cite{yin2024should}, KMMLU~\cite{son2024kmmlumeasuringmassivemultitask}, TurkishMMLU~\cite{yüksel2024turkishmmlumeasuringmassivemultitask}, and PersianMMLU~\cite{ghahroodi2024khayyam}. In particular, IndoMMLU also includes nine local cultures and eight local languages in Indonesia and ArabicMMLU is sourced from eight different countries in North Africa, the Levant, and the Gulf. The details about MMLU-series benchmarks are specified in Table~\ref{tab:mmlu-series}.

All benchmarks are built in multiple-choice questions (MCQ) format, although the number of candidate answers vary. Most benchmarks are built based on local exam questions and educational materials, with the exception of JMMLU~\cite{yin2024should}, which is is partially composed of translated questions from MMLU dataset~\cite{hendrycks2021measuring}. Although the benchmarks encompass diverse range of knowledge from K-12 education to professional and even industrial knowledge, each benchmark is split into different educational stages because each country has different educational curricula. KMMLU~\cite{son2024kmmlumeasuringmassivemultitask} and TurkishMMLU, are specialized for expert-level and high school-level questions respectively.

Beyond the MMLU-series benchmarks, KorNAT~\cite{lee2024kornat}, INVALSI~\cite{mercorio2024disce}, and FoundaBench~\cite{li2024foundabench} are created to test educational knowledge in South Korea, Italy, and China respectively. 
KorNAT~\cite{lee2024kornat} includes social value and common knowledge datasets. Specifically, the common knowledge dataset is developed based on the national compulsory education curriculum, covering seven subjects from the Korean GED syllabus. All questions are manually created by rephrasing the reference materials to MCQ format questions.
Similarly, the INVALSI benchmark ~\cite{mercorio2024disce} is structured based on the INVALSI test, a popular educational assessment criteria across Italy. The INVALSI test includes various domains, including mathematics, but it especially focuses on assessing a student's linguistic proficiency through various tasks. It consists of both MCQ and multiple complex choice questions (MCCQ) format.
Half the questions in FoundaBench~\cite{li2024foundabench} evaluate Chinese K-12 subject knowledge. K-12 education in China refers to compulsory primary and secondary education in China. Other than collecting questions from Chinese academic exams, they also automatically generate questions with GPT-4~\cite{openai2024gpt4technicalreport}. For automatic generation, they first extract key contents from collected documents, then manually formulate and refine optimal prompts through several iterations.
The M3Exam~\cite{zhang2023m3exam} specifically focuses on evaluating LLMs in a multilingual context. It includes nine languages from high-resource languages like English and Chinese, to extremely low-resource languages such as Javanese, each reflecting a distinct cultural background. They recruit native speakers from each region to manually collect official graduation exams in primary, middle, and high school.

In the medical domain, CMB~\cite{wang2024cmb} and K-NLEKMD~\cite{jang2023gpt} evaluate region-specific medical knowledge in Korea and China. Medical knowledge is often shaped by regional factors such as climate, diet, and ethnicity, leading to unique medical systems in each country.
CMB~\cite{wang2024cmb} is a comprehensive medical benchmark in Chinese that covers six categories of medical knowledge, including Physician, Nurse and Pharmacist domains. The questions are sourced from publicly available exam questions with solutions provided by medical experts.
K-NLEKMD~\cite{jang2023gpt} assesses language models' decision-making skills in Traditional Korean Medicine (TKM) using the Korean national licensing examination for Korean medicine doctors.

\textbf{\textsl{Takeaways from \S\ref{sec:academic_knowledge}.}}
Local educational resources, such as exams, are being leveraged to develop cultural knowledge evaluation benchmarks. However, subjects like mathematics typically cover more general knowledge, leading to the need of identifying culturally-specific questions within these benchmark. While some studies have identified tasks or questions that require culture-specific information~\cite{li-etal-2024-cmmlu, son2024kmmlumeasuringmassivemultitask}, there is still a lack of clarity regarding what specific cultural information is needed. Providing this cultural context could aid in the creation of more robust cross-cultural knowledge benchmarks.

\subsubsection{Commonsense Knowledge}
\label{sec:commonsense_knowledge}

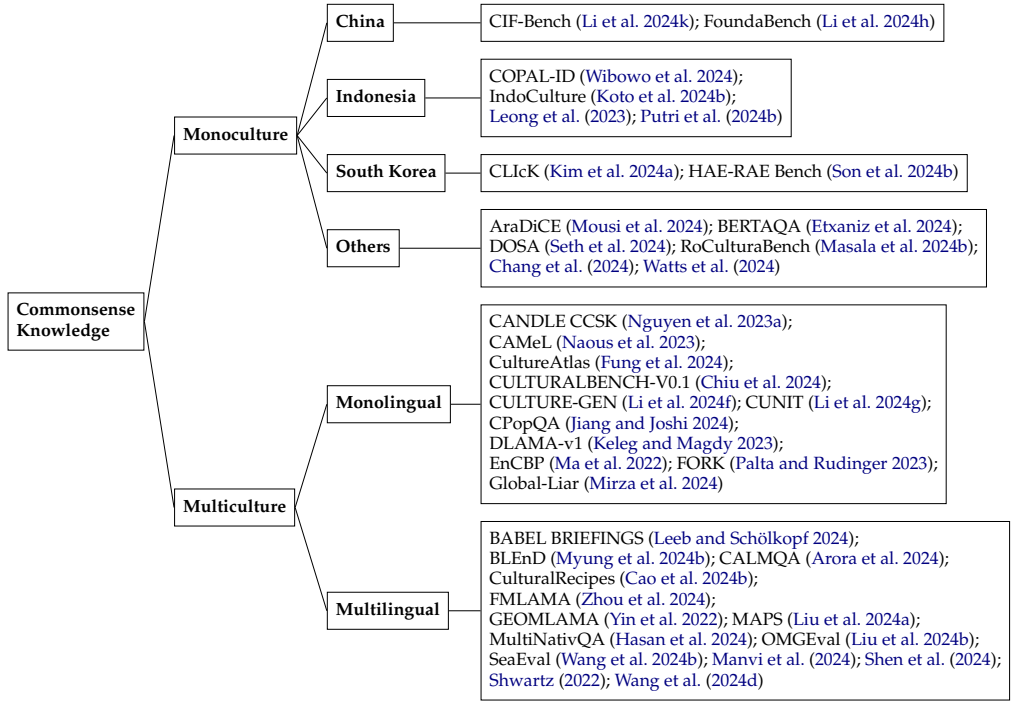
\begin{figure*}
\centering
\resizebox{\textwidth}{!}{
\begin{forest}
for tree={
  grow=east,
  reversed=true,
  rectangle,
  draw,
  align=left,
  anchor=west,
  parent anchor=east,
  child anchor=west,
  font=\scriptsize,
}
[\textbf{Commonsense}\\\textbf{Knowledge}
    [\textbf{Monoculture}, tier = 1
        [\textbf{China}, tier = 2
            [CIF-Bench~\cite{li2024cif};
            FoundaBench~\cite{li2024foundabench}
            , tier = leaf
            ]
        ]
        [\textbf{Indonesia}, tier = 2
            [COPAL-ID~\cite{wibowo2023copal};\\
            IndoCulture~\cite{koto2024indoculture};\\
            \citet{leong2023bhasa};
            \citet{putri2024llmgenerateculturallyrelevant}
            , tier = leaf
            ]
        ]
        [\textbf{South Korea}, tier = 2
            [CLIcK~\cite{kim2024click};
            HAE-RAE Bench~\cite{son2024hae}
            , tier = leaf
            ]
        ]
        [\textbf{Others}, tier = 2
            [AraDiCE~\cite{mousi2024aradice};
            BERTAQA~\cite{etxaniz2024bertaqa};\\
            DOSA~\cite{seth2024dosa};
            RoCulturaBench~\cite{masala2024vorbe};\\
            \citet{chang2024benchmarking};
            \citet{watts2024pariksha}
            , tier = leaf
            ]
        ]
    ]
    [\textbf{Multiculture}, tier = 1
        [\textbf{Monolingual}, tier = 2
            [CANDLE CCSK~\cite{nguyen2023extracting};\\
            CAMeL~\cite{naous2024havingbeerprayermeasuring};\\
            CultureAtlas~\cite{fung2024massively};\\
            CULTURALBENCH-V0.1~\cite{chiu2024culturalteaming};\\
            CULTURE-GEN~\cite{li2024culture};
            CUNIT~\cite{li2024how};\\
            CPopQA~\cite{jiang-joshi-2024-cpopqa};\\
            DLAMA-v1~\cite{keleg-magdy-2023-dlama};\\
            EnCBP~\cite{ma-etal-2022-encbp};
            FORK~\cite{palta-rudinger-2023-fork};\\
            Global-Liar~\cite{mirza2024global}
            , tier = leaf
            ]
        ]
        [\textbf{Multilingual}, tier = 2
            [BABEL BRIEFINGS~\cite{leeb-scholkopf-2024-diverse};\\
            BLEnD~\cite{myung2024blend};
            CALMQA~\cite{arora2024calmqa};\\
            CulturalRecipes~\cite{cao-etal-2024-cultural};\\
            FMLAMA~\cite{zhou2024does};\\
            GEOMLAMA~\cite{yin-etal-2022-geomlama};
            MAPS~\cite{liu-etal-2024-multilingual};\\
            MultiNativQA~\cite{hasan2024nativqa};
            OMGEval~\cite{liu2024omgeval};\\
            SeaEval~\cite{wang-etal-2024-seaeval};
            \citet{pmlr-v235-manvi24a};
            \citet{shen-etal-2024-understanding}; \\
            \citet{shwartz-2022-good};
            \citet{wang-etal-2024-countries}
            , tier = leaf
            ]
        ]
    ]
]
\end{forest}
}
\caption{Cultural commonsense knowledge evaluation benchmarks}
\label{fig:commonsense}

\end{figure*}

\begin{figure}[h!]
  \centering
  \begin{minipage}{\columnwidth}
    \centering
    \resizebox{\columnwidth}{!}{%
\begin{tabular}{@{}llll@{}}
\toprule
\begin{tabular}[c]{@{}l@{}} \textbf{Evaluation}\\\textbf{Method} \end{tabular} &
  \textbf{Question} &
  \textbf{Answer} &
  \textbf{Example Dataset/Benchmarks} \\ \midrule
\begin{tabular}[c]{@{}l@{}}\textbf{Binary QA} \end{tabular}&
  \begin{tabular}[c]{@{}l@{}}
      During \textbf{Chinese New Year}, red envelopes\\are given by the married to the unmarried.
  \end{tabular} &
  TRUE &
  \begin{tabular}[c]{@{}l@{}}
      CultureAtlas~\cite{fung2024massively}\\
      Global-Liar~\cite{mirza2024global}
  \end{tabular} \\ \midrule
\begin{tabular}[c]{@{}l@{}}\textbf{Multiple-}\\\textbf{Choice QA} \end{tabular}&
  \begin{tabular}[c]{@{}l@{}}What is the most common spice/herb used\\in dishes from \textbf{Greece}?\\A. BlackPepper\\ B. Cumin\\ C. Epazote\\ D. Oregano\end{tabular} &
  \begin{tabular}[c]{@{}l@{}}
  D. Oregano\\
  \end{tabular} &
  \begin{tabular}[c]{@{}l@{}}
      FoundaBench~\cite{li2024foundabench}\\
      CULTURALBENCH-V0.1~\cite{chiu2024culturalteaming}\\
      BLEnD~\cite{myung2024blend}
  \end{tabular} \\ \midrule
\textbf{Mask Filling} &
  \begin{tabular}[c]{@{}l@{}}
      In traditional \textbf{American} weddings, the\\color of wedding dress is usually \textbf{[MASK]}.
  \end{tabular} &
  \begin{tabular}[c]{@{}l@{}}
      White
  \end{tabular} &
  \begin{tabular}[c]{@{}l@{}}
      CAMeL~\cite{naous2024havingbeerprayermeasuring}\\
      DLAMA-v1~\cite{keleg-magdy-2023-dlama}\\
      GEOMLAMA~\cite{yin-etal-2022-geomlama}\\
  \end{tabular} \\ \midrule
\begin{tabular}[c]{@{}l@{}}\textbf{Short Answer}\\\textbf{Generation}\end{tabular} &
  \textbf{Which} is the biggest lake in \textbf{Nepal}? &
  \begin{tabular}[c]{@{}l@{}}
      The largest lake in Nepal is\\\textbf{Rara Lake} in Karnali Province.
  \end{tabular} &
  \begin{tabular}[c]{@{}l@{}}
      BLEnD~\cite{myung2024blend}\\
      MultiNativQA~\cite{hasan2024nativqa}
  \end{tabular} \\ \midrule
\begin{tabular}[c]{@{}l@{}}\textbf{Long Form}\\\textbf{Generation} \end{tabular} &
  \begin{tabular}[c]{@{}l@{}}
      When a person walks home late at night,\\\textbf{why} is it said that they should throw a\\stone as far as they can before entering\\their house?
  \end{tabular} &
  \begin{tabular}[c]{@{}l@{}}
      The idea of throwing a stone\\before entering your house late\\at night is rooted in \textbf{folklore,} \\\textbf{superstition}, and...
  \end{tabular} &
  \begin{tabular}[c]{@{}l@{}}
      CULTURE-GEN~\cite{li2024culture}\\
      CAMeL~\cite{naous2024havingbeerprayermeasuring}\\
      OMGEval~\cite{liu2024omgeval}\\
  \end{tabular} \\ \bottomrule
\end{tabular}%
}
\captionof{table}{Evaluation methods in academic and commonsense knowledge benchmarks}
\label{tab:eval-type}

  \end{minipage}
  \vspace{0.7cm}
  \begin{minipage}{\columnwidth}
    \centering
    \includegraphics[width=0.75\columnwidth]{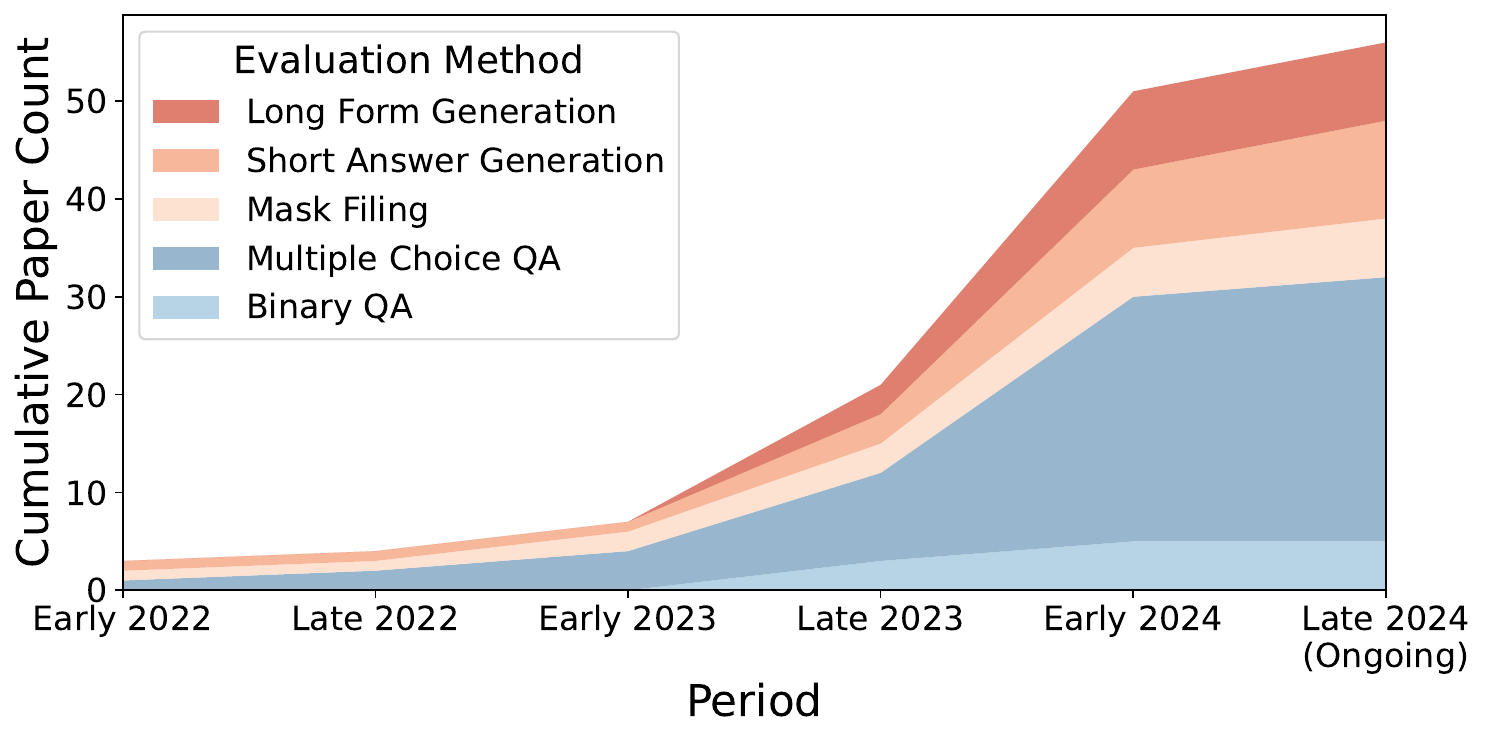}
    \captionof{figure}{Total Number of Academic and Commonsense Knowledge Benchmark Papers by Evaluation Method (2022 -- 2024)}
    \label{fig:eval-type}
  \end{minipage}  
\end{figure}

Evaluating commonsense knowledge has been widely recognized as a fundamental task in natural language understanding systems~\cite{davis2015commonsense}. To address this, commonsense knowledge bases like ConceptNet~\cite{speer2017conceptnet} and various commonsense reasoning and knowledge datasets, including COPA~\cite{roemmele2011choice}, SWAG~\cite{zellers2018swag}, and CommonsenseQA~\cite{talmor2019commonsenseqa}, have been developed. However, existing datasets often focus on Western commonsense knowledge, overlooking regional differences. In this section, we will discuss commonsense knowledge benchmarks that are developed to address this gap. We organize the papers according to their focus on culture and languages. Some benchmarks focus on a single culture, while others address cross-cultural differences by constructing multicultural benchmarks. The overall hierarchy of the papers in this section is specified in Figure \ref{fig:commonsense}. The academic and commonsense knowledge are evaluated with various methods specified in Table \ref{tab:eval-type}.

\textbf{Culture-specific Benchmarks.}
Cultural-specific commonsense knowledge benchmarks have been developed for various geographical regions and countries, including Indonesia~\cite{koto2024indoculture, leong2023bhasa, putri2024llmgenerateculturallyrelevant, talmor2019commonsenseqa, wibowo2023copal}, China~\cite{li2024foundabench, li2024cif}, Korea~\cite{kim2024click, son2024hae}, Taiwan~\cite{chang2024benchmarking}, India~\cite{seth2024dosa}, Romania~\cite{masala2024vorbe}, Basque Country~\cite{etxaniz2024bertaqa}, and Arabic regions~\cite{mousi2024aradice}. Each benchmark aims to capture the unique cultural knowledge of the target region.

Indonesia's diverse local cultures and languages have led to creation of various benchmarks. IndoCulture~\cite{koto2024indoculture} is designed to assess cultural knowledge across eleven Indonesian provinces using a sentence completion task. Each sample is in a multiple-choice format, providing a one-sentence premise with three plausible options and one correct answer. The dataset is manually created with the help of native speakers and covers twelve predefined topics spanning local customs and knowledge.
COPAL-ID~\cite{wibowo2023copal} follows COPA’s~\cite{roemmele2011choice} commonsense causal reasoning format. It also has manually created the dataset in which the local residents are involved to capture local cultural nuances, including local customs, terminology, and language nuances; this dataset is presented in standard Indonesian and Jakartan Indonesian.
\citet{leong2023bhasa} similarly develops a cultural diagnostics dataset with native speakers to evaluate basic cultural knowledge in Indonesian and Tamil languages. They categorize cultural knowledge into language, literature, history, and customs. To evaluate LLMs, they use a free-form generation prompts and analyze each response qualitatively.
In contrast, \citet{putri2024llmgenerateculturallyrelevant} build MCQ dataset in Indonesian and Sundanese languages by applying three different dataset generation methods. They first automatically adapt English CommonsenseQA~\cite{talmor2019commonsenseqa} dataset into target languages with LLMs. Also, they manually construct the questions with native speakers, and use LLMs to generate additional data based on the manually defined list of categories and concepts.

In China, FoundaBench~\cite{li2024foundabench} and CIF-Bench~\cite{li2024cif} are developed.
While half of the questions in FoundaBench evaluate K-12 academic knowledge, the other half are related to commonsense knowledge. Similar to the academic knowledge section of the dataset, they collect questions from Internet resources and automatically generate questions using GPT-4~\cite{openai2024gpt4technicalreport}. However, for commonsense knowledge questions, they additionally gather questions from online users about traditional Chinese culture and their life experiences.
CIF-Bench evaluates the zero-shot generalizability of LLMs in Chinese across 150 tasks. Of these, 113 tasks are adapted from existing datasets such as SNI~\cite{wang-etal-2022-super}. In addition, 37 tasks such as those related to traditional Chinese are manually created. The benchmark defines four output categories with corresponding evaluation metrics, such as using accuracy for multi-class classification tasks. They categorize the task output into the four categories and suggest evaluation metrics for each type. For multi-class classification and multi-label classification, they use accuracy and F1 score respectively. For creative generation tasks that have no absolute golden answer, they use model-based evaluators to evaluate creativity, fluency, the level of instruction-following and the confidence of the evaluator. For the remaining tasks, they use semantic similarity between the golden answer and the model output.

For South Korean culture, HAE-RAE Bench~\cite{son2024hae} is developed to capture culture-specific nuances in the Korean language. It consists of six downstream tasks, including general knowledge and history. The general knowledge questions are crowdsourced and includes sub-topics such as tradition, law, and Korean drama. For the history section, the authors manually craft questions from Namuwiki pages related with Korean history.
CLIcK~\cite{kim2024click} is a dataset focusing on Korean cultural and linguistic knowledge. The cultural commonsense knowledge part covers topics such as society, tradition, pop culture, and history. The questions are selected from standardized Korean exams, and additional questions are generated using GPT-4~\cite{openai2024gpt4technicalreport} based on textbook contents.

Various Asian countries other than Indonesia, China and Korea have developed following datasets and benchmarks.
\citet{chang2024benchmarking} build Taiwanese Hakka culture dataset. It draws primarily from Hakka Culture Encyclopedia and Taiwan Ministry of Education’s Hakka Knowledge Base. The questions are designed to include culturally relevant topics such as Hakka language, customs, history and architecture. Moreover, they specifically create questions in regards to Bloom's Taxonomy~\cite{bloom1956handbook, furst1981bloom} to assess LLMs' ability to apply, analyze, evaluate, and creatively utilize cultural knowledge.
\citet{seth2024dosa} construct DOSA to study India's local cultural identities based on India’s geographic states. This dataset is community-generated, and includes 615 social artifacts and represents 19 different Indian geographic subcultures. They initially use a survey to collect important subculture social artifacts. Then suggest a pipeline to get further annotation on each artifacts from state local residents.
\citet{watts2024pariksha} evaluate 30 models across 10 Indic languages with 20 manually created long-form generation prompts. The prompts include topics such as health, finance, and culturally nuanced questions. They compare model's generation abilities by performing pairwise comparisons with both LLM evaluators and human evaluators.
In addition, the AraDiCE~\cite{mousi2024aradice} benchmark includes a fine-grained dataset called AraDiCE-Culture. This dataset is specifically designed to assess regional Arabic cultural awareness across the Gulf, Egypt, and Levant regions. The questions are related to culturally significant topics such as public holidays, food, geography, and history.

Related to the European region, \citet{masala2024vorbe} develop RoCulturaBench to evaluate how well LLMs are grounded in the historical, cultural, and social realities of Romania. A team of Romanian humanities scholars manually constructed questions covering two subtopics. First is factual information about Romania, including its geography, history, and demography. The second part includes aspects of how Romanians perceive themselves and the world, with topics such as traditions, customs, beliefs, and stereotypes.
BERTAQA~\cite{etxaniz2024bertaqa} is a multiple-choice trivia dataset divided into two subsets, one focusing on local knowledge about the Basque Country, and the other covering global knowledge. The questions span 8 diverse categories, including society and tradition, sports and leisure, and science and technology. They initially create the dataset in Basque by crawling public sources, then create English version using a professional translation service.

Several Asian countries, including Indonesia, China, and Korea, are actively developing culture-specific commonsense knowledge evaluation benchmarks. Benchmarks for Indonesia in particular, aims to capture the country's diverse local cultures and languages, making an effort to represent local differences in the dataset~\cite{koto2024indoculture, putri2024llmgenerateculturallyrelevant}. Similarly, when developing culture-specific benchmarks, it is crucial to include local cultures rather than treating the entire country as a single, homogeneous culture. This would especially be important in ethnographically diverse countries where careful attention is needed to accurately reflect cultural diversity.

\textbf{Multicultural and Monolingual Benchmarks.}
In the following, we describe cross-cultural commonsense knowledge benchmarks that encompass a wide range of cultures. Most of them are built in English with one exception of Arabic~\cite{naous2024havingbeerprayermeasuring}. This enable the NLP community to conduct cross-cultural comparison on LLMs' cultural knowledge and reasoning ability with unified tasks. 

FORK is a food-related dataset that is manually created, containing 184 CommonsenseQA-style~\cite{talmor2019commonsenseqa} questions. These questions are categorized into three types based on how explicitly the reference country is mentioned.
In contrast, CULTURALBENCH-V0.1~\cite{chiu2024culturalteaming} is created semi-automatically through a combination of human expertise and AI assistance. They use a red-teaming approach~\cite{perez-etal-2022-red, ganguli2022red} to develop an AI-assisted system called CulturalTeaming, which integrates the creativity and cultural knowledge of human annotators with the scalability and standardization capabilities of LLMs. With this system, 45 human annotators create 252 MCQ dataset covering 34 different cultures.
CULTURE-GEN~\cite{li2024culture} is fully automatically generated. They leverage LLMs to generate response on eight culture-related topics across 110 countries and regions, using a country list sourced from the World Value Survey~\cite{haerpfer2012world}. From these LLM outputs, cultural symbols are automatically extracted and matched to their respective cultures. Using the linguistic concept of ``markedness''~\cite{WAUGH+1982+299+318}, they found that culture-specific generations are characterized by distinct cultural symbols.

CUNIT~\cite{li2024how}, CAMeL~\cite{naous2024havingbeerprayermeasuring}, and EnCBP~\cite{ma-etal-2022-encbp} are semi-automatically constructed benchmarks that source data from web resources such as Wikipedia and social media platforms like Twitter. After automatically gathering data from these online sources, they undergo additional human annotation or validation to enhance their quality and relevance.
CUNIT~\cite{li2024how} evaluates LLMs' ability to identify culturally similar concept pairs. It focuses on traditional culture-specific concepts related to clothing and food across 10 countries. The dataset is created by first collecting cultural concepts and descriptions from Wikipedia, followed by detailed manual annotation of culturally significant features.
CAMeL~\cite{naous2024havingbeerprayermeasuring} is an Arabic benchmark that comprises entities extracted from
Wikidata and CommonCrawl corpus. Each entities have human-annotated culture labels. The masked prompts that is used to evaluate LM's cultural adaptation ability are retrieved from X (formerly Twitter) for natural context. The cultural bias and stereotypes are evaluated by analyzing adjectives and doing sentiment analysis on story generation with Arab and Western entities. Furthermore, they define a Cultural Bias Score (CBS) to measure the preference of cultural entities in masked token prediction.
EnCBP~\cite{ma-etal-2022-encbp} is designed for cultural background prediction using English-language news articles. The dataset consists of articles collected from major news outlets in five English-speaking countries and four U.S. states. Through manual validation via MTurk,\footnote{\url{https://www.mturk.com}} a crowdsourcing platform and cultural domain compatibility assessments, the study demonstrates that cultural background heavily influences writing style, even within the same language.

Several benchmarks are created automatically by utilizing various web resources.
DLAMA-v1~\cite{keleg-magdy-2023-dlama} evaluates models' factual knowledge across cultures by automatically generating factual triples using SPARQL queries from Wikidata. This method produces factual knowledge triples with 20 relation predicates covering 21 Western, 22 Arab, 13 Asian, and 12 South American countries.
CultureAtlas~\cite{fung2024massively} introduces a multicultural knowledge extraction approach by systematically navigating Wikipedia documents on cultural topics through a network of linked pages. The dataset not only includes positive cultural knowledge samples but also creates negative samples to assess LLMs' understanding of multicultural knowledge. It spans over a hundred countries and covers cultural topics such as etiquette, holidays, and traditional clothing.
Similary, CANDLE-CCSK~\cite{nguyen2023extracting} conduct a large web crawl. It introduces CANDLE, an end-to-end methodology for automatically extracting cultural commonsense knowledge (CCSK) at scale. CANDLE extracts 1.1 million CCSK assertions, organizing them into clusters across three domains and five cultural facets. Among the three domains, the geography domain includes 196 countries. The cultural facets include food, clothing, and traditions.
\citet{jiang-joshi-2024-cpopqa} introduce CPopQA, a ranking-based statistical QA task that compares the popularity of cultural concepts across 58 countries. The dataset is automatically constructed using Wikipedia’s list of public holidays and Google Books Ngram Viewer (GBNV) corpus\footnote{\url{https://books.google.com/ngrams/}}. GBNV is used to estimate the popularity of each holiday within a country by leveraging the statistical frequency of the holiday’s name and the publication country of each book.
Global-Liar~\cite{mirza2024global} source true-false statements from online websites to evaluate the fact-checking performance of LLMs. The dataset covers six global regions, Africa, Asia-Pacific, Europe, Latin America, North America, and the Middle East, with 100 true-false statements per region. The true statements are sourced from reputable news outlets in their respective regions, while false statements were obtained from AFP FactCheck.\footnote{\url{https://factcheck.afp.com}}

\textbf{Multicultural and Multilingual Benchmarks.}
The following introduces multicultural and multilingual commonsense knowledge benchmarks. Each benchmark contains cultural knowledge tied to specific regions, often represented by each regions' native language. Some benchmarks use language as a proxy for culture, aligning a single language with a particular culture or country. Others recognize that cultural regions may be linguistically diverse, and some languages are spoken across multiple cultural regions. Thus, language-culture pairs do not always have a one-to-one correlation in each benchmark.

Benchmarks such as BLEnD~\cite{myung2024blend}, GEOMLAMA~\cite{yin-etal-2022-geomlama} are created manually either by recruiting local annotators directly, or through crowdsoucing platforms.
BLEnD~\cite{myung2024blend} is specifically designed to capture everyday knowledge that is often not explicitly documented in online data source. It spans six categories including food, sports, and family. It is manually created by recruiting native annotators both directly and through a crowdsourcing platform Prolific.\footnote{\url{https://www.prolific.com/}} It covers 13 languages spoken across 16 different countries and regions, including underrepresented areas such as West Java and North Korea. The final dataset contains 52.6k QA pairs, comprising 15k short-answer and 37.6k multiple-choice questions.
GEOMLAMA~\cite{yin-etal-2022-geomlama} covers geo-diverse knowledge about the United States, China, India, Iran, and Kenya, with prompts constructed with native language for each country, English, Chinese, Hindi, Persian, and Swahili. It is also manually created by recruiting native annotators from each country. The dataset contains 3K masked prompts related to geo-diverse concept including culture and customs, and provides different gold answer for each country.
CALMQA~\cite{arora2024calmqa} is a multilingual long-form question-answering dataset focused on culturally specific questions. It contains 1.5K questions across 23 high to low resource languages for broad range of topics such as governance and society, religion and customs, and history. The dataset is built by collecting naturally occurring questions from community web forums and by hiring native speakers to create questions in under-resourced, rarely-studied languages like Fijian and Kirundi.
\citet{shwartz-2022-good} propose the task of mapping time expressions across different cultures. They collect gold standard annotations through a crowdsourcing platform for the start and end times of five time expressions, morning, noon, afternoon, evening, and night. The annotations span English, Hindi, Italian, and Portuguese, representing the cultural contexts of the US, India, Italy, and Brazil.

Web resources in multiple languages can be leveraged to automatically construct cross-cultural knowledge benchmarks.
FMLAMA~\cite{zhou2024does} is a cross-cultural culinary knowledge dataset. The dataset is created by systematically querying Wikidata to extract a broad range of food-related information. It focuses on topologically diverse set of languages, including English, Chinese, Arabic, Korean, Russian, and Hebrew. 
\citet{hasan2024nativqa} introduces the NativQA framework, designed to create culturally and regionally specific natural question-answering datasets. The resulting MultiNativQA benchmark comprises over 72K QA pairs across seven languages and seven cities, spanning languages such as English, Bangla, Hindi, Nepali and Assamese. It also captures linguistic diversity by incorporating various dialects, including multiple Arabic dialects and two distinct variations of Bangla.
\citet{liu-etal-2024-multilingual} present MAPS, a dataset of proverbs across six geographically diverse languages. They collect proverbs and sayings from Wikiquote and Wiktionary. By testing MAPS with a wide range of open source LLMs, they show that LLMs possess knowledge of proverbs and sayings to varying degrees, although significantly biased toward English and Chinese.
\citet{leeb-scholkopf-2024-diverse} introduce BABEL BRIEFINGS dataset with 4.7m news headlines from August 2020 to November 2021, across 30 languages and 54 locations worldwide. They automatically collect news headlines using the News API.\footnote{\url{https://newsapi.org/}} This dataset can be utilized to compare the coverage of events across different countries and languages, or identifying cultural biases in reporting.

Some benchmarks utilize existing NLP benchmarks built in different cultural contexts to create cross-cultural knowledge benchmarks~\cite{liu2024omgeval, cao-etal-2024-cultural}. Others, like SeaEval~\cite{wang-etal-2024-seaeval}, aim for comprehensive cross-cultural evaluations by merging various NLP datasets.
OMGEval~\cite{liu2024omgeval} is an open-source multilingual generative test set designed to evaluate LLMs' general knowledge and capabilities. It provides 804 open-ended questions across five languages, building on AlpacaEval~\cite{NEURIPS2023_5fc47800}, with 805 entries as foundational data. The dataset underwent multilingual translation, manual localization, and thorough manual verification to ensure global relevance.
CulturalRecipes~\cite{cao-etal-2024-cultural} is a bidirectional Chinese-English dataset focused on cross-cultural recipe adaptation. It draws from two existing monolingual corpora, RecipeNLG~\cite{bien-etal-2020-recipenlg} and XiaChuFang~\cite{liu-etal-2022-counterfactual}. The authors also manually create a small golden dataset for cultural recipe adaptation. They use both reference-based automatic metrics and human evaluation to assess cross-cultural recipe adaptation in text generation.
SeaEval~\cite{wang-etal-2024-seaeval} is a benchmark for evaluating multilingual foundation models on language capabilities, complex reasoning, and cultural understanding. Covering eight languages, it incorporates 29 datasets with 13,263 samples. SeaEval draws on existing benchmarks for fundamental language skills and reasoning, while manually constructing four datasets—US-Eval, SG-Eval, CN-Eval, and PH-Eval—focused on distinct cultural regions. Also to evaluate cross-lingual consistency, SeaEval introduces two new datasets, Cross-MMLU and Cross-LogiQA, based on the MMLU~\cite{hendrycks2021measuring} and LogiQA2.0~\cite{10174688} datasets.
\citet{shen-etal-2024-understanding} offer a comprehensive evaluation of LLMs' performance in cultural commonsense tasks. The study examines culture-specific commonsense knowledge using datasets like GeoMLAMA~\cite{yin-etal-2022-geomlama} and CANDLE~\cite{nguyen2023extracting}, and explores the influence of cultural context in general commonsense reasoning using GenericsKB~\cite{bhakthavatsalam2020genericskb}. Their findings reveal significant performance disparities across cultures, showing that LLMs often associate general commonsense with dominant cultures. They also highlight that the language used to query LLMs has a substantial impact on performance in culture-related tasks.

We can also directly analyze and evaluate LLM-generated outputs for cross-cultural commonsense.
\citet{wang-etal-2024-countries} evaluate cultural dominance by building a multilingual dataset that includes both concrete and abstract cultural objects. LLMs are prompted to list 10 concrete cultural objects in 11 languages, and the authors introduce an ``In-Culture'' score to measure cultural dominance by assessing how many responses align with the culture of the corresponding language, based on Wikipedia.
\citet{pmlr-v235-manvi24a} evaluate the geographic bias of LLMs using prompts that elicit zero-shot predictions based on specific geographic locations. While the models' predictions show strong correlations with ground truths on objective topics like annual precipitation, population density, and infant mortality rate, they often consistently overestimate or underestimate the ranks of certain regions.

\textbf{\textsl{Takeaways from \S\ref{sec:commonsense_knowledge}}.} Commonsense knowledge benchmarks are being developed in various cultures. Most of these benchmarks use language or country as proxies to define cultural boundaries. However, inconsistencies in these definitions makes cross-cultural evaluation across existing benchmarks challenging. Furthermore, some studies do not adequately consider sociolinguistic factors when defining these boundaries. To advance future cross-cultural research, it is essential to establish well-defined and consistent cultural boundaries. Also, as shown in Figure \ref{fig:eval-type}, while most evaluation methods are based on multiple-choice QAs, recent studies have begun to explore the evaluation of LLMs' text generation capabilities with short answer and long-form generation tasks. However, most current long-form evaluation approaches depend on human evaluation or LLM-as-a-judge methods~\cite{NEURIPS2023_91f18a12}, which are limited in scalability and lack culturally-specific evaluation. Therefore, further research is needed to develop robust automatic evaluation methods, especially for long-form generation tasks.



\subsubsection{Social Values}
\label{sec:value}

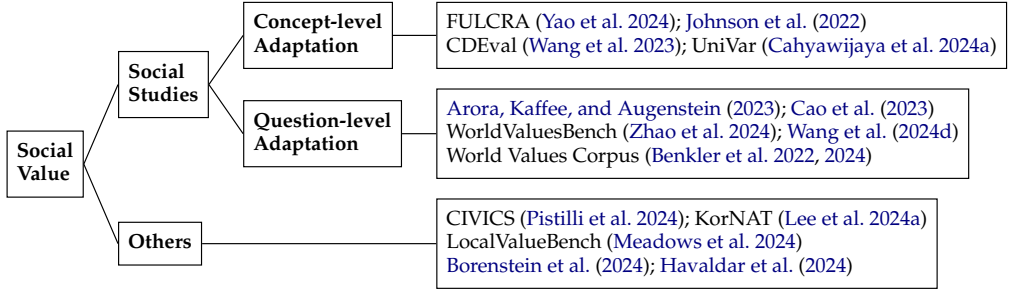
\begin{figure*}
\centering
\resizebox{\textwidth}{!}{
\begin{forest}
for tree={
  grow=east,
  reversed=true,
  rectangle,
  draw,
  align=left,
  anchor=west,
  parent anchor=east,
  child anchor=west,
  font=\scriptsize,
}
[\textbf{Social}\\\textbf{Value}
    [\textbf{Social}\\\textbf{Studies}, tier=1
        [\textbf{Concept-level}\\\textbf{Adaptation}, tier=2
            [FULCRA~\cite{yao-etal-2024-value};
            \citet{johnson2022ghost}\\
            CDEval~\cite{wang2023cdeval};
            UniVar~\cite{cahyawijaya2024high}
            ,tier=leaf
            ]
        ]
        [\textbf{Question-level}\\\textbf{Adaptation},tier=2
            [\citet{arora-etal-2023-probing};
            \citet{cao-etal-2023-assessing}\\
            WorldValuesBench~\cite{zhao-etal-2024-worldvaluesbench};
            \citet{wang-etal-2024-countries}\\
            World Values Corpus~\cite{benkler2022cultural, benkler-etal-2024-recognizing}
            ,tier=leaf
            ]
        ]
    ]
    [\textbf{Others}, tier=1
        [CIVICS~\cite{pistilli2024civics};
        KorNAT~\cite{lee2024kornat}\\
        LocalValueBench~\cite{meadows2024localvaluebench}\\
        \citet{borenstein2024investigatinghumanvaluesonline,havaldar-etal-2024-building}\\
        ,tier=leaf
        ]
    ]
]
\end{forest}
}
\caption{Culturally-aware evaluation of social values}
\label{fig:value}
\end{figure*}

Social values refer to the common beliefs in a society about what is good, desirable, and important. They reflect what a society or individual considers important and prioritize certain outcomes or behaviors (e.g., equality, freedom, solidarity). Social values are not necessarily prescriptive rules, but they are common goals that influence how people behave and make decisions. The overall hierarchy of the papers in this section is specified in Figure~\ref{fig:value}.

Most studies in cultural NLP that focus on social values use studies from social sciences for evaluation, such as Hofstede’s Cultural Dimensions Theory~\cite{hofstede-2005} and the World Values Survey (WVS,~\citet{wvs2022}). 
WorldValuesBench~\cite{zhao-etal-2024-worldvaluesbench} leverages questions from WVS to create a large-scale benchmark for multi-cultural value prediction, where models are required to predict the social values conditioned on various demographic contexts.
The World Values Corpus~\cite{benkler2022cultural, benkler-etal-2024-recognizing} introduces a new task called Recognizing Value Resonance (RVR) and constructs a dataset based on questions from the World Values Survey (WVS). This dataset is designed to assess the models' perspectives on implicit cultural values and beliefs through the analysis of text pairs.
\citet{wang-etal-2024-countries} leverage survey questions from WVS and the Political Coordinates Test (PCT) to assess the cultural dominance of LLMs on abstract concepts such as values and opinions.
CDEval~\cite{wang2023cdeval} introduces a questionnaire-based benchmark designed to measure the cultural dimensions of LLMs, focusing on the six cultural dimensions defined by Hofstede’s Cultural Dimensions Theory.
\citet{cao-etal-2023-assessing} employ survey questions based on Hofstede’s Cultural Dimensions Theory to assess the cultural alignment between LLMs and human societies in 5 different languages and cultures.
\citet{arora-etal-2023-probing} utilize survey questions from Hofstede’s Cultural Dimensions Theory and WVS to show that multilingual pre-trained language models learn cross-cultural value differences, but they weakly correlate with the surveys.
\citet{johnson2022ghost} use WVS as a comparative framework to assess how GPT-3 tends to align the values in its generated outputs with the social values prevalent in the U.S., often leading to conflicts with input texts that originate from other cultural contexts.
FULCRA~\cite{yao-etal-2024-value} applies Schwartz’s Theory of Basic Values~\cite{schwartz2012overview} to assess the underlying values guiding LLMs' behaviors, facilitating the identification of current safety risks and the prediction of future risks.
UniVar~\cite{cahyawijaya2024high} identifies 87 reference human values by synthesizing insights from existing studies, such as the World Values Survey (WVS) and Hofstede’s Cultural Dimensions Theory. These values are then used to construct value-eliciting QA pairs in 25 languages, which serve as a basis for evaluating how current LLMs reflect human values across different languages.
These studies commonly highlight the challenges LLMs face in aligning their values with diverse cultural contexts, and emphasize that these models tend to reflect values more aligned with WEIRD (Western, Educated, Industrialized, Rich, and Democratic) societies.

Other studies have focused on evaluating LLMs' alignment with specific regional or cultural values. CIVICS~\cite{pistilli2024civics} collects text excerpts from authoritative sources in Singapore, France, Canada, the United Kingdom, and Australia to create prompts for evaluating LLMs' responses to value-sensitive topics, including immigration, LGBTQI rights, and social welfare.
KorNAT~\cite{lee2024kornat} develops a social value dataset designed to assess LLMs' alignment with the social values of Korean citizens, based on a large-scale survey featuring questions generated using social conflict keywords and timely keywords specific in Korea.
LocalValueBench~\cite{meadows2024localvaluebench} presents a benchmark to evaluate LLMs' alignment with Australian values, addressing topics such as social norms, legal principles, and cultural practices.
\citet{borenstein2024investigatinghumanvaluesonline} conduct a large-scale study of differences in Schwartz values between online communities on Reddit.
\citet{havaldar-etal-2024-building} introduce a knowledge-guided lexicon to model cultural variation within a country, highlighting the significance of measuring cultural differences across its regions and applying this framework to NLP models.

\textbf{\textsl{Takeaways from \S\ref{sec:value}.}} Most papers examining social values rely on existing global surveys from the social sciences, resulting in high regional coverage. However, it is important to note that while social values can vary significantly at sub-country levels \cite{havaldar-etal-2024-building}, most studies concentrate solely on country-level analyses. This highlights the need for more granular research that captures local nuances in social values beyond national boundaries.

\subsubsection{Social Norms and Morals}
\label{sec:norms_morals}

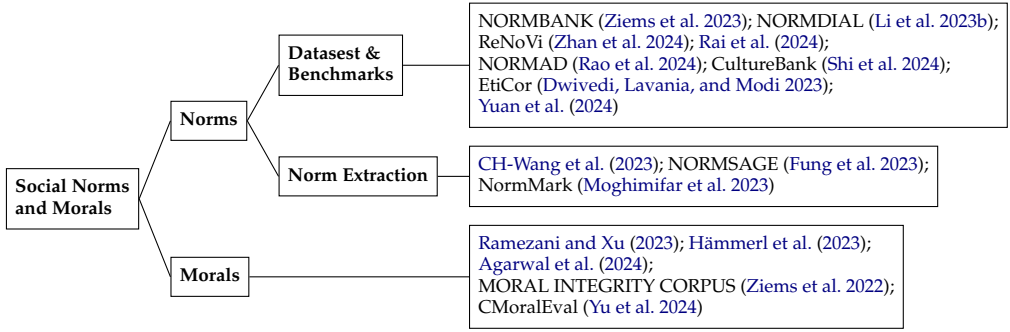
\begin{figure*}
\centering
\resizebox{\textwidth}{!}{
\begin{forest}
for tree={
  grow=east,
  reversed=true,
  rectangle,
  draw,
  align=left,
  anchor=west,
  parent anchor=east,
  child anchor=west,
  font=\scriptsize,
}
[\textbf{Social Norms}\\\textbf{and Morals}
    [\textbf{Norms}, tier=1
        [\textbf{Datasest \&}\\\textbf{Benchmarks}, tier=2
            [NORMBANK~\cite{ziems2023normbank};
            NORMDIAL~\cite{li2023normdial};\\
            ReNoVi~\cite{zhan-etal-2024-renovi};
            \citet{rai2024cross};\\
            NORMAD~\cite{rao2024normad};
            CultureBank~\cite{shi2024culturebank};\\
            EtiCor~\cite{dwivedi-etal-2023-eticor};\\
            \citet{yuan-etal-2024-measuring}
            ,tier=leaf
            ]
        ]
        [\textbf{Norm Extraction}, tier=2
            [\citet{ch-wang-etal-2023-sociocultural};
            NORMSAGE~\cite{fung-etal-2023-normsage};\\
            NormMark~\cite{moghimifar-etal-2023-normmark}
            ,tier=leaf
            ]
        ]
    ]
    [\textbf{Morals}
        [\citet{ramezani2023knowledge}; \citet{haemmerl-etal-2023-speaking}; \\\citet{agarwal-etal-2024-ethical};\\MORAL INTEGRITY CORPUS \cite{ziems-etal-2022-moral};\\CMoralEval \cite{yu-etal-2024-cmoraleval}
        ,tier=leaf
        ]
    ]
]
\end{forest}
}
\caption{Culturally-aware evaluation of social norms and morals}
\label{fig:norms_morals}
\end{figure*}

Social norms and morals refer to more specific rules or principles that dictate how individuals should behave in everyday situations. They often reflect shared expectations within a community about what is acceptable or unacceptable behavior. It differs from social values in that values represent broader, abstract ideals or goals that guide what people strive for, while norms and morals provide concrete guidelines for behavior and decision-making within specific contexts~\cite{matsumoto2007culture}. The overall hierarchy of the papers in this section is specified in Figure~\ref{fig:norms_morals}.

Recent research has emphasized the need to equip language models with a nuanced understanding of these norms to effectively navigate diverse social settings. 
\citet{ziems2023normbank} introduced NORMBANK, a hierarchical knowledge bank of social norms designed to support non-monotonic reasoning over cultural norms, which are viewed as flexible standards that change with context.
Expanding on the need for cultural adaptability, \citet{li2023normdial} presented NORMDIAL, a bilingual dataset capturing social norm adherence and violations within dialogues for Chinese and American contexts. By modeling norm observance at a turn-by-turn level, this dataset demonstrates how conversational nuances and expectations vary between languages, providing insights into how language models can handle violations across languages. 
\citet{zhan-etal-2024-renovi} introduce ReNoVi, a large-scale corpus of 9,258 multi-turn dialogues annotated with social norms, designed to help AI systems understand and remediate norm violations.
\citet{rai2024cross} presented the first cross-cultural dataset of self-conscious emotions drawn from Bollywood and Hollywood films, alongside over 10K identified social norms, underscoring cultural differences such as Bollywood's focus on social roles and family honor. 
\citet{yuan-etal-2024-measuring} present a new social norms benchmark based on the U.S. K-12 curriculum, designed to evaluate LLMs' understanding of social norms. They also introduce a multi-agent framework that improves LLMs' social norm comprehension, bringing it closer to human-level understanding.

In a broader exploration, \citet{rao2024normad} introduced NORMAD, a dataset encompassing social and cultural norms from 75 countries, revealing that LLMs tend to demonstrate stronger adaptability to English-centric cultures.
\citet{shi2024culturebank} introduce CultureBank, a large-scale cultural knowledge base built from cultural descriptors sourced from TikTok and Reddit, used to evaluate LLMs' cultural knowledge across 2K cultural groups and 36 cultural topics, including social norms.
\citet{dwivedi-etal-2023-eticor} present EtiCor, an Etiquettes Corpus containing texts about social norms from five global regions, designed to evaluate LLMs' understanding of region-specific etiquettes.

Other studies focus on developing frameworks to extract culture-specific norms from existing text, which can be further used to evaluate language models. \citet{ch-wang-etal-2023-sociocultural} propose a novel approach to discover and reach descriptive social norms across Chinese and American cultures using a human-AI cooperation framework, and introduce the task of explainable social norm entailment to test the models' reasoning across cultures. \citet{fung-etal-2023-normsage} present NORMSAGE, a framework that automatically extracts culture-specific norms from multilingual conversations using GPT-3, offering explainable self-verification to ensure the norms' correctness in a conversation on the fly. \citet{moghimifar-etal-2023-normmark} propose NormMark, a probabilistic generative Markov model that captures latent features throughout a dialogue to improve norm recognition, outperforming existing methods, including GPT-3, on weakly annotated data by leveraging variational techniques and conversation history.

In terms of morals, \citet{ramezani2023knowledge} examined whether language models can capture moral norm variations across different countries using global datasets. They emphasized the limitations of monolingual English models in generalizing across cultures, particularly regarding sensitive topics such as homosexuality and divorce. \citet{haemmerl-etal-2023-speaking} examine the moral biases embedded in pre-trained multilingual language models (PMLMs) and their implications for cross-lingual transfer in German, Czech, Arabic, Chinese, and English. Their findings reveal that PMLMs encode varying moral biases that often misalign with cultural differences and human judgments, which can lead to harmful consequences in cross-lingual applications.
\citet{agarwal-etal-2024-ethical} investigate how LLMs perform ethical reasoning across multiple languages---English, Spanish, Russian, Chinese, Hindi, and Swahili---examining whether the language of the prompt influences the models' moral judgments.
\citet{ziems-etal-2022-moral} introduce the MORAL INTEGRITY CORPUS (MIC), a resource comprising 38,000 prompt-reply pairs and 99,000 Rules of Thumb (RoTs) that capture the moral intuitions behind dialogue system responses. 
\citet{yu-etal-2024-cmoraleval} present CMoralEval, a benchmark dataset designed for the morality evaluation of Chinese large language models (LLMs), consisting of 14,964 explicit moral scenarios and 15,424 moral dilemma scenarios sourced from a Chinese TV program and various media.

\textbf{\textsl{Takeaways from \S\ref{sec:norms_morals}.}} Like social values, research on social norms and morals has high regional coverage, primarily due to data sourced from online media such as Wikipedia and Reddit. However, most studies are conducted in English, overlooking the possibility that LLMs may have different understandings of social norms when prompted in various languages. Multilingual cross-cultural evaluations are needed.

\subsubsection{Social Bias and Stereotype}
\label{sec:bias}

\begin{figure*}
\centering
\resizebox{\textwidth}{!}{
\begin{forest}
for tree={
  grow=east,
  reversed=true,
  rectangle,
  draw,
  align=left,
  anchor=west,
  parent anchor=east,
  child anchor=west,
  font=\scriptsize,
}
[\textbf{Social Bias}\\\textbf{and Stereotype}
    [\textbf{WEAT-based}, tier=1
        [CA-WEAT~\cite{espana-bonet-barron-cedeno-2022-undesired}\\
         WEATHub~\cite{mukherjee-etal-2023-global}\\
         \citet{borah2024regionawarebiasevaluationmetrics}
         ,tier=leaf
         ]
    ]
    [\textbf{Sentence Pairs}, tier=1
        [French CrowS-Pairs~\cite{neveol-etal-2022-french}\\
         Multilingual CrowS-Pairs~\cite{fort-etal-2024-stereotypical}\\
         IndiBias~\cite{sahoo-etal-2024-indibias};
         Indian-BhED~\cite{khandelwal-etal-2024-indianbhed}\\
         CHBias~\cite{zhao-etal-2023-chbias};
         BIBED~\cite{das-etal-2023-toward}\\
         RuBia~\cite{grigoreva-etal-2024-rubia}
         ,tier=leaf
         ]
    ]
    [\textbf{Identity-Stereotype}\\\textbf{Pairs}, tier=1
        [SPICE~\cite{dev-etal-2023-building};
         SeeGULL~\cite{jha-etal-2023-seegull}\\
         SGM~\cite{bhutani2024seegullmultilingualdatasetgeoculturally}
         ,tier=leaf
         ]
    ]
    [\textbf{BBQ Series}, tier=1
        [CBBQ~\cite{huang-xiong-2024-cbbq};
         KoBBQ~\cite{jin-etal-2024-kobbq}\\
         JBBQ~\cite{yanaka2024analyzingsocialbiasesjapanese}\\
         MBBQ~\cite{neplenbroek2024mbbqdatasetcrosslingualcomparison}
         ,tier=leaf
         ]
    ]
    [\textbf{Others}, tier=1
        [\citet{zhu-etal-2024-quite};
         CHAST~\cite{dammu2024theyunculturedunveilingcovert}
         ,tier=leaf
         ]
    ]
]
\end{forest}
}
\caption{Culturally-aware evaluation of social biases and stereotypes}
\label{fig:bias}
\end{figure*}
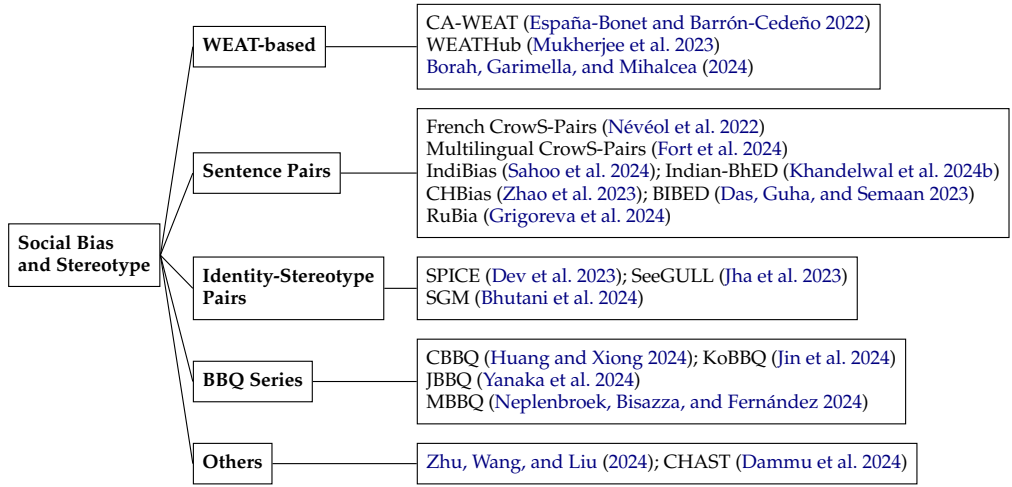

With the growing recognition of the need to detect and mitigate social bias in language models, several bias benchmarks and metrics have been developed.
However, most of them have been built in English, reflecting Western cultures.
As social biases and stereotypes depend on cultural contexts, several studies have emphasized the need for and developed bias benchmarks and metrics that include non-US cultures in their own languages.
Figure~\ref{fig:bias} classifies studies according to the type of stereotype dataset, which is associated with the corresponding bias evaluation methods.

The Word Embedding Association Test (WEAT,~\citet{weat}) has been used to measure the bias of language models at the word embedding level.
By computing the similarity between the word embeddings, it assesses the association between the target and the attribute depicting a particular stereotype.
However, since it originated from the Implicit Association Test (IAT)~\cite{greenwald1998measuring} developed in English in the United States, the stereotypes and the list of words representing the targets and the attributes possess some linguistic and cultural bias.
\citet{espana-bonet-barron-cedeno-2022-undesired} introduce Cultural Aware WEAT (CA-WEAT) lists in 26 languages.
They have native speakers create new lists of words associated with the stereotypes that are considered universally accepted, flowers-pleasant versus insects-unpleasant and musical instruments-pleasant versus weapons-unpleasant.
\citet{mukherjee-etal-2023-global} release WEATHub, a multilingual extension of WEAT.
It features 24 languages, each with native speaker involvement to translate the relevant English WEAT with appending language-specific words and add new human-centered bias dimensions.
\citet{borah2024regionawarebiasevaluationmetrics} propose a data-driven method to extract region-aware gender bias topic pairs for WEAT-based evaluation.
Additionally, they let LLMs generate personae of someone interested in the given topic, and measure the mismatch rate with the associated gender.

The method of measuring bias in language models using sentence pairs that have similar structures but refer to two different social groups has also been widely used.
The bias is measured by analyzing sentence-level probabilities to determine whether the language models tend to favor sentences that align more closely with societal stereotypes.
Research has been actively conducted to create datasets composed of sentence pairs in various languages that reflect stereotypes from diverse cultures.
\citet{neveol-etal-2022-french} present French CrowS-Pairs by adapting the original CrowS-Pairs~\cite{nangia-etal-2020-crows} and newly crowdsourcing stereotyped statements.
\citet{fort-etal-2024-stereotypical} further extend it to Multilingual CrowS-Pairs with seven additional languages.
\citet{sahoo-etal-2024-indibias} construct IndiBias, a Hindi and English dataset for India, by adapting the original CrowS-Pairs and generating identity-stereotype pairs and corresponding sentence pairs by LLM-human partnership.
Indian-BhED~\cite{khandelwal-etal-2024-indianbhed} is another dataset targeting India, which covers stereotypes about caste and religions. It is an English dataset created by the authors based on literature and their own knowledge and validated by experts.
CHBias~\cite{zhao-etal-2023-chbias} is a Chinese dataset of sentences that are retrieved from Weibo based on the bias specifications and annotated by native Chinese speakers. Each sentence in the validation and test sets is paired with a sentence in which the target term is swapped.
BIBED~\cite{das-etal-2023-toward}, a Bengali dataset, includes not only sentence pairs that explicitly mention the target identity terms but also those with names, kinship phrases, and synonymous colloquial lexicons that imply gender, religious, or national identities in the Bengali context.

Another type of stereotype dataset consists of pairs of identities and stereotypes.
The models are typically evaluated by measuring the mean entailment for the pairs of one sentence containing the identity group and another sentence containing the associated stereotype.
SPICE~\cite{dev-etal-2023-building} is an English dataset constructed through an open-ended survey to cover diverse and locally situated stereotypes in India.
SeeGULL~\cite{jha-etal-2023-seegull} is an English benchmark built using LLMs in the loop to cover stereotypes about identity groups spanning 176 countries and state-level identities within the US and India.
\citet{bhutani2024seegullmultilingualdatasetgeoculturally} extend it to SeeGULL Multilingual (SGM) to cover 23 language-country pairs. They evaluated LLMs by asking them to choose a target identity associated with the given stereotype.

As LLMs become prevalent, there has been a surge in the use of the Bias Benchmark for Question Answering (BBQ)~\cite{parrish-etal-2022-bbq}, which can assess bias in LLMs through a question-answering (QA) format.
It comprises ambiguous contexts and disambiguated contexts with discriminatory questions for evaluating QA accuracy and bias scores in each type of context.
CBBQ~\cite{huang-xiong-2024-cbbq} is a Chinese version of BBQ benchmark, which consists of ambiguous contexts, questions, and answer choices written by humans, and disambiguating contexts generated by GPT-4~\cite{openai2024gpt4technicalreport}.
KoBBQ~\cite{jin-etal-2024-kobbq}, for South Korea, is constructed through a culturally sensitive adaptation of the original BBQ, validated by a large-scale survey conducted among South Koreans.
JBBQ~\cite{yanaka2024analyzingsocialbiasesjapanese} is also manually built from English BBQ to target Japanese.
MBBQ~\cite{neplenbroek2024mbbqdatasetcrosslingualcomparison} consists of BBQ samples dealing with the stereotypes that are common in English, Dutch, Spanish, and Turkish, and is used to compare LLMs' behavior across different languages.

Meanwhile, \citet{zhu-etal-2024-quite} focus on revealing ChatGPT's~\cite{chatgpt} nationality bias, that is, bias in discourse about people of a certain nationality.
They use automatic metrics for vocabulary richness, sentiment, and offensiveness, let ChatGPT score itself, and have ChatGPT and experts pairwise compare the offensiveness of the discourses.
To disclose the Covert Harms and Social Threats (CHAST) in LLM-generated conversations in hiring scenarios, \citet{dammu2024theyunculturedunveilingcovert} propose the CHAST metrics based on social science literature and align the evaluation model with expert assessments.
They note that LLMs tend to generate more harmful content when involving the Indian caste compared to the Western-centric race attributes.

\textbf{\textsl{Takeaways from \S\ref{sec:bias}.}}
While embedding-based and probabilistic methods have been widely used to measure social bias and stereotypes in language models, their application to proprietary LLMs often proves challenging.
Furthermore, existing research on LLMs that considers cross-cultural differences of social bias among multiple cultures tends to focus on specific bias categories or a limited set of cultures.
To bridge this gap, there is a growing need for further research on methodologies that enable comprehensive and cross-cultural evaluation of the bias of various language models across diverse cultures.

\subsubsection{Toxicity and Safety}
\label{sec:toxic}

\begin{figure*}
\centering
\resizebox{\textwidth}{!}{
\begin{forest}
for tree={
  grow=east,
  reversed=true,
  rectangle,
  draw,
  align=left,
  anchor=west,
  parent anchor=east,
  child anchor=west,
  font=\scriptsize,
}
[\textbf{Toxicity and Safety}
    [\textbf{Monoculture}, tier=1
        [\citet{arango-monnar-etal-2022-resources};
         AraTrust~\cite{alghamdi2024aratrustevaluationtrustworthinessllms}\\
         KOLD~\cite{jeong-etal-2022-kold};
         KoSBi~\cite{lee-etal-2023-kosbi}\\
         CRU~\cite{ullah-etal-2024-detecting}
         ,tier=leaf
        ]
    ]
    [\textbf{Multi/Cross-}\\\textbf{Culture}, tier=1
        [XtremeSpeech~\cite{maronikolakis-etal-2022-listening}\\
         \citet{korre-etal-2024-challenges}\\
         RTP-LX~\cite{dewynter2024rtplxllmsevaluatetoxicity}; \citet{tonneau-etal-2024-languages}\\ 
         \citet{lee-etal-2023-hate}; CREHate~\cite{lee-etal-2024-exploring-cross}
         ,tier=leaf
        ]
    ]
    [\textbf{Multidimension}, tier=1
        [WalledEval~\cite{gupta2024walledevalcomprehensivesafetyevaluation};
         AART~\cite{radharapu2023aartaiassistedredteamingdiverse}
         ,tier=leaf
        ]
    ]
]
\end{forest}
}
\caption{Culturally-Aware Evaluation in Toxicity and Safety}
\label{fig:toxic}
\end{figure*}
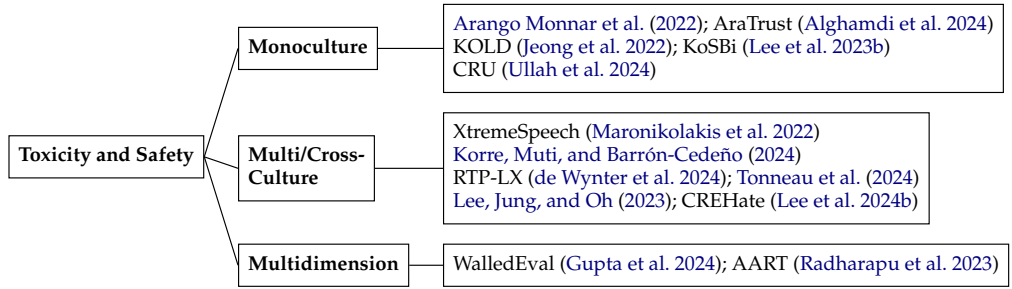

This section covers studies on hate speech, offensive language, toxicity, and safety, highlighting cross-cultural differences in their manifestation and the evaluation of language model safety from diverse cultural perspectives.
The overview for this section is depicted in Figure~\ref{fig:toxic}.

Pointing out that the research on toxicity, offensive language, and hate speech predominantly focused on English, \citet{arango-monnar-etal-2022-resources} build a Spanish dataset by getting annotations for tweets from Chile, and \citet{jeong-etal-2022-kold} construct a Korean offensive language dataset, KOLD, by getting annotations for comments from NAVER news and YouTube.
Also, \citet{lee-etal-2023-kosbi} present a Korean social bias dataset, KoSBi, which consists of the context-sentence pairs generated by HyperCLOVA~\cite{kim-etal-2021-changes} given the target demographic group, with the human-annotated labels of \emph{safe} or \emph{unsafe} (stereotype, prejudice, discrimination, or other).
\citet{alghamdi2024aratrustevaluationtrustworthinessllms} introduce AraTrust, a comprehensive trustworthiness benchmark in Arabic composed of multiple-choice questions on truthfulness, ethics, privacy, illegal activities, mental health, physical health, unfairness, and offensive language, by curating questions from exams, existing datasets, and online websites, or creating them manually.
\citet{ullah-etal-2024-detecting} build CRU, a benchmark in Roman Urdu for cybercrime detection with three types of cybercrimes, including hate speech, cyber terrorism, and cyber harassment. They systematically collect tweets from Twitter and RUHSOLD~\cite{rizwan-etal-2020-hate}, then conduct annotation with experts following the Pakistani legal framework regarding cybercrimes.

Expanding the views on hate speech and toxic language to include multicultural and cross-cultural perspectives, several studies demonstrate cultural insensitivity and biases present in language models and datasets and develop multicultural and inclusive datasets.
\citet{maronikolakis-etal-2022-listening} present XtremeSpeech, a hate speech dataset containing social media contents across Brazil, Germany, India, and Kenya. They specifically recruit local annotators from each country for data collection and annotation.
\citet{dewynter2024rtplxllmsevaluatetoxicity} construct RTP-LX by transcreating English RealToxicPrompts~\cite{gehman-etal-2020-realtoxicityprompts} to 28 languages and manually creating culturally nuanced toxic language, with annotating eight categories of harm.
\citet{korre-etal-2024-challenges} explore the creation of a multilingual parallel hate speech dataset using machine translation. They found that while machine translation adequately preserves the intended meaning of the sentences, it still produces grammatical and syntactical errors, showcasing the challenges of creating a parallel hate speech corpus.
Meanwhile, \citet{lee-etal-2023-hate} highlight the cultural insensitivity of language models by demonstrating that the monolingual hate speech classifiers show lower performance in classifying the translated texts from other cultures.
\citet{lee-etal-2024-exploring-cross} verify the intra-language cultural disparities in hate speech annotation and LLMs' detection performance bias towards Anglosphere countries by constructing CREHate, which comprises online posts with hate speech annotations from five English-speaking countries.
\citet{tonneau-etal-2024-languages} disclose the intra-language geographical bias of English, Arabic, and Spanish hate speech datasets, as inferring the location of each tweet's author reveals that a handful of countries are disproportionately overrepresented in the datasets.

Recent work that comprehensively evaluates the safety of LLMs includes cultural perspectives as one of the various evaluation factors.
\citet{gupta2024walledevalcomprehensivesafetyevaluation} release WalledEval, a comprehensive AI safety evaluation toolkit, with SGXSTest and HIXSTest, which consist of safe and unsafe prompts for testing LLMs' refusal behavior within the cultural context of Singapore and Hindi, respectively.
\citet{radharapu2023aartaiassistedredteamingdiverse} propose an AI-assisted red-teaming method, AART, to create adversarial queries customized for various application contexts with adversarial evaluation dimensions such as locale and language.

\citet{prabhakaran-etal-2024-grasp} propose GRASP, a disagreement analysis framework,
and uncover systematic disagreements across various intersectional subgroups. They suggest that the sociocultural background of human annotators can lead to disagreement in subjective tasks, such as safety and offensiveness annotations.

\textbf{\textsl{Takeaways from \S\ref{sec:toxic}.}} The primary role of language models in the toxicity and safety field used to be moderating communication between online users. However, the advent of LLMs has introduced a new challenge of evaluating the toxicity and safety of contents generated by LLMs. This shift has necessitated broader research on toxicity and safety, encompassing not only communication between users but also between users and AI models.
Additionally, acknowledging intercultural variations within a single language and attempting to analyze people's perceptions from diverse perspectives are noteworthy and deserve further exploration in other tasks as well.

\subsubsection{Emotional and Subjective Topics}
\label{sec:subjective}

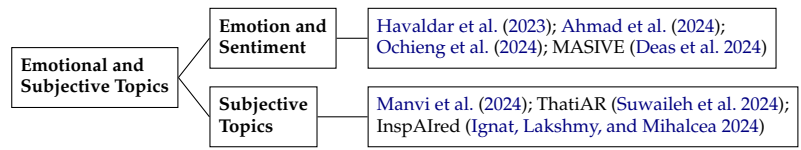
\begin{figure*}
\centering
\begin{forest}
for tree={
  grow=east,
  reversed=true,
  rectangle,
  draw,
  align=left,
  anchor=west,
  parent anchor=east,
  child anchor=west,
  font=\scriptsize,
}
[\textbf{Emotional and}\\\textbf{Subjective Topics}
    [\textbf{Emotion and}\\\textbf{Sentiment}, tier=1
            [\citet{havaldar-etal-2023-multilingual};
             \citet{ahmad-etal-2024-generative};\\
             \citet{ochieng2024metricsevaluatingllmseffectiveness};
             MASIVE~\cite{deas2024masiveopenendedaffectivestate}, tier=leaf]
    ]
    [\textbf{Subjective}\\\textbf{Topics}, tier=1
            [\citet{pmlr-v235-manvi24a};
             ThatiAR~\cite{suwaileh2024thatiarsubjectivitydetectionarabic};\\
             InspAIred~\cite{ignat2024crossculturalinspirationdetectionanalysis}, tier=leaf]
    ]
]
\end{forest}
\caption{Culturally-aware evaluation for emotional and subjective topics}
\label{fig:subjective}
\end{figure*}
This section introduces studies that evaluate cultural biases in language models on emotional and subjective topics, as well as those examining various tasks and datasets affected by individuals' cultural backgrounds. Figure~\ref{fig:subjective} categorizes these researches into emotion detection tasks and other subjective topics.

The evaluation of language models across various cultural contexts has also been explored for emotional and subjective topics.
\citet{havaldar-etal-2023-multilingual} show the bias towards English and American contexts in emotion embeddings of the multilingual model, as well as in the emotion prediction performance of the LLMs.
They also illustrate that the psychological cultural differences of \emph{pride} and \emph{shame} between the US and Japan are not clearly reflected in the language models.
\citet{ahmad-etal-2024-generative} focus on Hausa to compare responses of ChatGPT~\cite{chatgpt} with human native speakers to questions about emotions in the Nigerian cultural context.
\citet{ochieng2024metricsevaluatingllmseffectiveness} qualitatively demonstrate LLMs' struggle to incorporate the complex cultural nuances in sentiment analysis using a code-mixed (English, Swahili, and Sheng) WhatsApp chat dataset.
\citet{pmlr-v235-manvi24a} demonstrate the geographical bias of LLMs on subjective topics, in addition to the objective topics (\S \ref{sec:commonsense_knowledge}). Despite an unbiased model being expected to respond independently of location, the LLMs' predictions for the subjective topics (likability, attractiveness, morality, intelligence, and work ethic) are correlated with the infant survival rate of the location, which is a proxy of socioeconomic conditions.

Some studies extend emotional and subjective tasks that depend on cultural perspectives to diverse cultural contexts.
\citet{deas2024masiveopenendedaffectivestate} expand the emotion set covered by the emotion detection benchmarks to be unbounded, by defining the affective state identification (ASI) task to predict affective states when single words expressing the feeling are masked from the text about emotional experience.
They also release MASIVE, which contains affective states in English and Spanish Reddit posts.
\citet{suwaileh2024thatiarsubjectivitydetectionarabic} present ThatiAR, an Arabic dataset of news sentences with manually annotated labels on subjectivity and LLM-generated rationals and instructions.
They demonstrate how political, historical, and cultural bias and subjectivity of the writers and readers affect detecting subjectivity in the news.
\citet{ignat2024crossculturalinspirationdetectionanalysis} construct a dataset of culturally inspiring content called InspAIred.
The contents are sourced by searching keywords like `inspiration' and `motivation' in subreddits of regions in India and the UK, and labeled by crowdworkers and a fine-tuned model.
The dataset is augmented by GPT-4~\cite{openai2024gpt4technicalreport} to generate inspiring content as a Reddit user from India or the UK.
They analyze inspiring content across cultures, comparing AI-generated and real ones, in terms of stylistic and structural features such as complexity, descriptiveness, and readability, as well as semantic and psycholinguistic features using topic modeling and psycholinguistic markers.

\textbf{\textsl{Takeaways from \S\ref{sec:subjective}.}} Emotional and subjective topics are areas where individual differences can vary significantly, even within the same culture. As a result, reaching a consensus can be challenging because individuals often have diverse opinions on these matters. Therefore, it is essential to consider how individuals' backgrounds and perspectives may affect evaluations and datasets, and efforts should be made to clearly account for the influence of \emph{culture}.

\subsubsection{Linguistics}
\label{sec:linguistics}

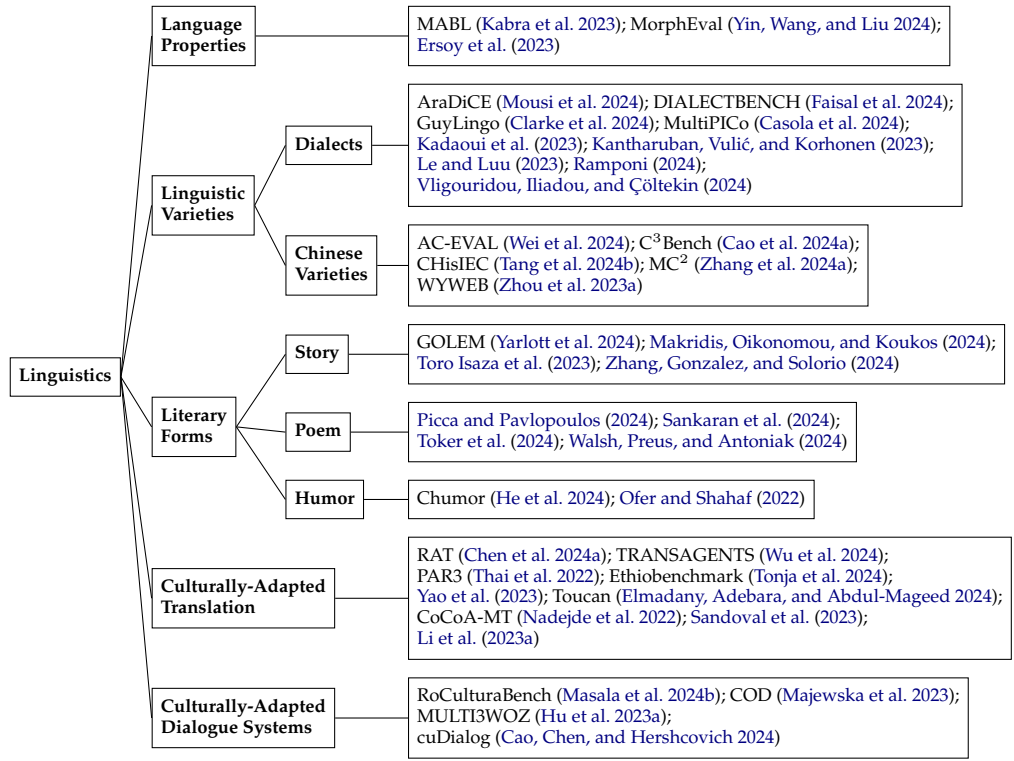
\begin{figure*}
\centering
\resizebox{\textwidth}{!}{
\begin{forest}
for tree={
  grow=east,
  reversed=true,
  rectangle,
  draw,
  align=left,
  anchor=west,
  parent anchor=east,
  child anchor=west,
  font=\scriptsize,
}
[\textbf{Linguistics}
    [\textbf{Language}\\\textbf{Properties}, tier=1
        [MABL~\cite{kabra-etal-2023-multi};
        MorphEval~\cite{yin-etal-2024-chinese};\\
        \citet{ersoy-etal-2023-languages}
        ,tier=leaf
        ]
    ]
    [\textbf{Linguistic}\\\textbf{Varieties}, tier=1
        [\textbf{Dialects}, tier=2
            [AraDiCE~\cite{mousi2024aradice}; DIALECTBENCH~\cite{faisal2024dialectbench}; \\
            GuyLingo~\cite{clarke-etal-2024-guylingo};
            MultiPICo~\cite{casola-etal-2024-multipico}; \\
            \citet{kadaoui2023tarjamat};
            \citet{kantharuban2023quantifying}; \\
            \citet{le2023parallel};
            \citet{ramponi-2024-language};\\
            \citet{vligouridou-etal-2024-treebank} 
            ,tier=leaf
            ]
        ]
        [\textbf{Chinese}\\\textbf{Varieties}, tier=2
            [AC-EVAL~\cite{wei2024ac}; C$^{3}$Bench~\cite{cao2024c3benchcomprehensiveclassicalchinese}; \\
            CHisIEC~\cite{tang-etal-2024-chisiec};
            MC$^{2}$~\cite{zhang-etal-2024-mc2};\\
            WYWEB~\cite{zhou-etal-2023-wyweb}
            ,tier=leaf
            ]
        ]
    ]
    [\textbf{Literary}\\\textbf{Forms}, tier=1
        [\textbf{Story}, tier=2
            [GOLEM~\cite{yarlott-etal-2024-golem}; \citet{makridis2024fairylandai}; \\
            \citet{toro-isaza-etal-2023-fairy};
            \citet{zhang-etal-2024-interpreting} 
            ,tier=leaf
            ]
        ]
        [\textbf{Poem}, tier=2
            [\citet{picca-pavlopoulos-2024-deciphering}; \citet{sankaran2024revisiting}; \\
            \citet{toker-etal-2024-dataset};
            \citet{walsh2024sonnet} 
            ,tier=leaf
            ]
        ]
        [\textbf{Humor}, tier=2
            [Chumor~\cite{he2024chumor};
            \citet{ofer-shahaf-2022-cards}
            ,tier=leaf
            ]
        ]
    ]
    [\textbf{Culturally-Adapted}\\\textbf{Translation}, tier=1
        [RAT~\cite{chen2024benchmarking};
        TRANSAGENTS~\cite{wu2024perhaps};\\
        PAR3~\cite{thai2022exploring};
        Ethiobenchmark~\cite{tonja2024ethiollm};\\
        \citet{yao2023benchmarking};
        Toucan~\cite{elmadany2024toucan};\\
        CoCoA-MT~\cite{nuadejde2022cocoa};
        \citet{sandoval2023rose};\\
        \citet{li2023translate}
        ,tier=leaf
        ]
    ]
    [\textbf{Culturally-Adapted}\\\textbf{Dialogue Systems}, tier=1
        [RoCulturaBench~\cite{masala2024vorbe};
        COD~\cite{majewska2023cross};\\
        MULTI3WOZ~\cite{hu2023multi};\\
        cuDialog~\cite{cao2024bridging}
        ,tier=leaf
        ]
    ]
]
\end{forest}
}
\caption{Culturally-aware evaluation in linguistics}
\label{fig:linguistic}
\end{figure*}

Culture and language are deeply interconnected~\cite{IMAI201670}. Cultural elements, such as formality, are often explicitly reflected in language properties~\cite{heylighen1999formality}. Also, language varieties including dialects provide valuable insights into local cultures, while literary forms like stories and poems offer rich resources for studying culture~\cite{ramponi-2024-language, doi:10.1080/0954025900020204}. Additionally, cultural factors play a critical role in pragmatics, particularly in translation and dialogue systems~\cite{rohmawati2022idioms}. This section will examine how cultural aspects are expressed through language properties, varieties, and literary forms, and how these elements inform applications like translation and dialogue systems. The overall hierarchy of the papers in this section is specified in Figure \ref{fig:linguistic}.

\textbf{Language Properties.}
Formality is a stylistic property of language that typically carries information about the culture of the speaker or the writer~\cite{heylighen1999formality}. \citet{ersoy-etal-2023-languages} analyze the formality of generative
multilingual language models BLOOM~\cite{scao:hal-03850124} and XGLM~\cite{lin2021few} across five languages. They classify 1,200 generations per language as formal, informal, or incohesive and measure the impact of the prompt formality on the generation text.
\citet{kabra-etal-2023-multi} create a figurative language inference dataset, MABL, for seven diverse languages associated with a variety of cultures. They collect figurative language by crowdsourcing native speakers. They also categorize knowledge needed to understand each metaphor by using the commonsense categories defined in \citet{liu-etal-2022-testing}. This work reveals that each language relies on cultural and regional concepts for figurative expressions
\citet{yin-etal-2024-chinese} construct MorphEval, a Chinese Morpheme-informed Evaluation benchmark. It also contains morphemes with cultural implications, which are Chinese yet need some cultural background to understand. They collect data from a dictionary-based resource~\cite{liu2018towards}. From prediction errors, approximately 16\% of errors occur due to cultural implications, hinting lack of cultural understanding in LLMs.

\textbf{Linguistic Variety.}
Linguistic varieties, such as dialects represent linguistic and cultural diversity, as they encapsulate unique elements of local culture. Yet, many of them are in danger of disappearing~\cite{moseley2010atlas}. Following studies emphasize the importance of collecting more low-resource dialectal data to capture the linguistic and cultural intricacies of diverse communities.

Asia Minor Greek dialects are endangered dialects rich in history and culture that face a dire struggle for preservation due to declining speaker base and scarce linguistic resources. Thus, \citet{vligouridou-etal-2024-treebank} present a manually annotated treebank of Pharasiot, one of the Asia Minor Greek dialects, following the Universal Dependencies framework~\cite{de-marneffe-etal-2021-universal}.
\citet{ramponi-2024-language} also introduces endangered language varieties of Italy. They address the challenge of the existing machine-centric assumptions of NLP for Italy's language varieties and suggest responsible and speaker-centric efforts to preserve language varieties of Italy.
Similarly, GuyLingo, a corpus in Creolese has been proposed in \citet{clarke-etal-2024-guylingo}; Creolese is the most widely spoken language in the culturally rich nation of Guyana, but has limited written source, making it a low-resource language in NLP field. 

Addressing the limitations of current NLP models in handling non-standard Vietnamese dialects, \citet{le2023parallel} present a parallel corpus for Central and Northern Vietnamese dialects. The corpus is created manually by Central and Northern dialect annotators.
\citet{kadaoui2023tarjamat} evaluate machine translation performance for various Arabic dialects to English. Arabic sentences are manually collected from the Open Islamic Texts Initiative (OpenITI) dataset~\cite{nigst2021openiti} and various online sources, including news outlets and YouTube videos.
Also, AraDiCE~\cite{mousi2024aradice} evaluates LLMs on their ability to comprehend and generate dialects primarily focusing on the Levantine (LEV) and Egyptian (EGY) dialects. The approach involves using machine translation from English to Modern Standard Arabic (MSA) and MSA to dialects, followed by human post-editing, to create synthetic benchmarks for low-resource dialects.
\citet{zhang-etal-2024-mc2} present MC$^{2}$, a multilingual corpus of minority languages in China, including four underrepresented languages, Tibetan, Uyghur, Kazakh, and Mongolian. They carefully design strategies for the selection of web pages to crawl, ensuring the language purity of the crawled texts. They show that writing systems play a crucial role in developing culturally-aware NLP systems with languages with multiple writing systems, such as Kazakh and Mongolian.

While modern Chinese is studied vigorously in NLP community, there is lack of effort on classical Chinese. Classical Chinese differs from modern Chinese in writing and grammar, thus benchmarks designed in modern Chinese can not be applied well to the studies in the classical Chinese domain.
To address this, C$^{3}$Bench~\cite{cao2024c3benchcomprehensiveclassicalchinese} and WYWEB~\cite{zhou-etal-2023-wyweb} are designed to evaluate the classical Chinese understanding capabilities of LLMs. Both benchmarks include basic NLP tasks such as sentence classification and machine translation.
For historical knowledge, AC-EVAL~\cite{wei2024ac} provides a comprehensive evaluation of LLMs' proficiency in understanding the ancient Chinese language and historical knowledge. The dataset consists of 3k multiple-choice questions, covering historical periods from the Pre-Qin era to the Qing dynasty.
Also, \citet{tang-etal-2024-chisiec} introduce CHisIEC, an ancient Chinese historical information extraction corpus. It is sourced from 13 historical books from the representative Twenty-Four Histories as the raw data, spanning over 1830 years and contains NER and RE tasks.
Furthermore, \citet{liang-etal-2024-text} build a traditional ecological knowledge base from Shanhai Jing, a record of flora and fauna in ancient China, written 2000 years ago. They employ a rule-based knowledge extraction method, which can also be utilized for further ancient language processing.

Expanding to multilingual comprehensive overview of dialects and linguistic varieties, \citet{faisal2024dialectbench} introduce DIALECTBENCH, a large-scale benchmark encompassing 40 language clusters with 281 varieties. They use language resources in papers from the past 10 years of the ACL Anthology. And categorize language clusters and varieties based on the Glottolog language database~\cite{nordhoff2012linked}.
\citet{kantharuban2023quantifying} also conduct a comprehensive evaluation of LLMs across regional dialects, examining 30 dialects across 7 languages for machine translation and 33 dialects across 7 languages for automatic speech recognition.
\citet{casola-etal-2024-multipico} propose MultiPICo, a multilingual corpus of ironic short conversations extracted from Twitter and Reddit. It covers 9 languages and 25 varieties and each conversation is annotated as ironic or not by crowdsourcing workers with different social backgrounds.

\textbf{\textsl{Takeaways from Linguistic Variety.}} Language varieties such as dialects or ancient languages offer valuable insights into local or historical culture. However, many of these language varieties are in danger of extinction~\cite{moseley2010atlas}. Current works primarily focuses on varieties of Chinese~\cite{zhang-etal-2024-mc2, wei2024ac} and Arabic~\cite{mousi2024aradice, kadaoui2023tarjamat}, underscoring the need for further studies on other languages with diverse varieties, such as Spanish, Hindi and English. \\


\textbf{Literary Forms.}
Storybooks, especially fairy tales, are particularly important to children’s mental, emotional, and social development and has been subject to analyzing social bias in its text~\cite{doi:10.1080/0954025900020204}.
\citet{toro-isaza-etal-2023-fairy} conduct a case study that analyze gender bias in fairy tales. They also propose an automatic pipeline that can extract character attributes and story's temporal narrative event chain for each characters. They also present an event annotation scheme to assist bias analysis.
Furthermore, \citet{makridis2024fairylandai} introduces FairyLandAI, a model designed to create personalized fairytales for children. Its architecture mimics the cognitive and creative processes involved in storytelling and character development found in children's literature. FairyLandAI supports personalized storytelling in multiple languages, catering to children's individual language preferences and cultural backgrounds.
Narrative texts, such as fables and folktales, often convey a lesson via a series of events with a clear consequence. \citet{zhang-etal-2024-interpreting} introduce the first dataset specifically designed for interpretive comprehension of themes in narrative texts. They use educational stories from different eras and cultural backgrounds.
Motifs often originate in folklore, which is a recurring cultural “memes” that are grounded in a story. \citet{yarlott-etal-2024-golem} present GOLEM, the first dataset annotated for motific information. The dataset comprises 8k English news articles, opinion pieces, and broadcast transcripts annotated for motific information. The human annotators from three cultural groups, Jewish, Irish and Puerto Rican annotate the type of usage of motifs within a text.

Rhymes and poems are a powerful medium for transmitting cultural norms and societal roles.
\citet{walsh2024sonnet} assess the poetic capabilities of LLMs by evaluating their recognition of poetic forms, which is patterns of sound that exist within specific cultural and linguistic contexts. They create a dataset of over 4.1K poems, tagged and categorized by human annotators, sourced from online platforms and books. The dataset, however, shows biases related to race, class, language, and culture due to uneven distribution across poetic forms. Further research is needed to explore LLMs' poetic abilities in languages beyond English.
Similarly, \citet{sankaran2024revisiting} address gender biases in rhymes and poems by collecting children's rhymes and adolescent poetry, including 20 translated poems from 11 languages. The data were selected to ensure diversity in style, content, and cultural background. Annotators identified gender stereotypes, which were rectified using LLMs and human educators. A survey-based comparison found no significant difference in their effectiveness, highlighting the potential of LLMs in reducing gender bias.
Since poetry was a prominent genre in late antique and medieval Hebrew literature, the corpus is rich in figures of speech like similes and metaphors. However, Hebrew texts are often annotated manually, a time-consuming and labor intensive process. Thus \citet{toker-etal-2024-dataset} present a medieval Hebrew poetry dataset with expert annotations of metaphor, and evaluate several Hebrew language models for automatic metaphor detection.
Iliad1 is one of the most significant pieces of ancient Greek poem. To propel the domain of emotion
analysis in classical literature forward, \citet{picca-pavlopoulos-2024-deciphering} present the first publicly available, emotion-annotated dataset of the Iliad1.

Understanding humor is one of the most difficult cognitive ability of human.
\citet{ofer-shahaf-2022-cards} explore humor in the context of the popular card game “Cards Against
Humanity” where players complete fill-in-the-blank statements using cards that can be offensive or politically incorrect. They introduce 300k online games of Cards Against Humanity, including 785k unique jokes, a large and strongly labeled humor dataset.
Addressing the lack of resources for humor datasets and evaluations in non-English languages, \citet{he2024chumor} introduces Chumor, a Chinese humor understanding dataset sourced from Ruo Zhi Ba, a Chinese Reddit-like platform for sharing intellectually challenging and culturally specific jokes. One of the authors annotated all explanations in the dataset.

\textbf{\textsl{Takeaways from Literary Forms.}} Stories and poems are actively studied for their valuable insights into culture knowledge and biases. However, beyond stories and poems, many other types of literature remain underexplored. For example, in fiction, genres like science fiction, historical fiction, and romance can provide unique cultural perspectives~\cite{doi:10.1177/2158244017723690}. Also, non-fiction works, such as journalism and travel writing, can reveal people's perceptions of their own culture and foreign cultures~\cite{berger2004deconstructing}, showing a promising area for future research.\\

\textbf{Culturally Adapted Translation.}
Cultural adaptation has long been a focus of translation studies~\cite{newmark2003textbook}. Effectively translating culture-specific items, such as idioms, historical references, and culturally unique concepts, is important for achieving effective cross-cultural communication~\cite{rohmawati2022idioms}.

Recent advancements in Machine Translation (MT), particularly multilingual pre-trained models, have improved translation qualities, also for low-resource languages such as Ethiobenchmark proposed by~\citet{tonja2024ethiollm}, a benchmark dataset of diverse downstream NLP tasks covering five Ethiopian languages with English. Similarly, \citet{elmadany2024toucan} introduce Toucan, an Afrocentric MT model supporting 156 African language pairs, which significantly outperforms other models in African language MT, as evaluated using the AfroLingu-MT benchmark. However, a gap remains in effectively translating cultural-specific content due to the inherent cultural differences associated with various languages, not fully captured through MT techniques~\cite{akinade2023varepsilon}. 

One such challenge is translating formal and informal tones appropriately, particularly in languages with honorifics or formality markers. \citet{nuadejde2022cocoa} address this issue with CoCoA-MT, a dataset and benchmark for controlling formality in translations across six languages. By fine-tuning contrastive data, their model successfully controls for formality while maintaining overall translation quality, demonstrating the importance of aligning translations with cultural expectations.

\citet{yao2023benchmarking} also contributes to this effort by enhancing the ability of MT systems to handle culture-specific entities. They introduce a data curation pipeline by creating a parallel corpus enriched with annotations specific to cultural items. Additionally, they suggest a new evaluation metric to assess the \textit{understandability} alongside \textit{accuracy} of culturally adapted translations in a reference-free manner. Similarly, ~\citet{lou2023cceval} introduce CCEval, a Chinese-centric multilingual MT evaluation benchmark designed to assess translation quality across 11 languages, ensuring better alignment with human evaluations through rigorous dataset curation.

Beyond literal translation, \citet{han2023bridging} tackles the challenge of bridging background knowledge gaps through automatic explicitation. Using the WIKIEXPL1 dataset from Wikipedia, they generate contextual explanations that help clarify missing cultural context, improving understanding in multilingual question-answering frameworks.

One of the most challenging areas in culturally adapted MT is literary translation, where the emotional and historical context plays a vital role in conveying meaning~\cite{jones2013faithful, toral2015machine}. \citet{thai2022exploring} introduce the PAR3 dataset, aligning novels with human and machine translations, and find that human translations are significantly preferred by experts. Their post-editing model improves translation quality, showing potential for addressing discourse disruptions and stylistic inconsistencies in literary MT. \citet{chen2024benchmarking} focus on translating classical Chinese poetry, which requires not only accuracy but also fluency and elegance. They propose a Retrieval-Augmented Translation method that enhances translation by integrating external knowledge, addressing the limitations of LLMs handling poetry. Additionally, a novel approach to literary translation is explored by \citet{wu2024perhaps}, who introduce a multi-agent collaboration framework called TRANSAGENTS. This framework mirrors previous publishing processes by using multiple agents to translate complex literary works. To evaluate its effectiveness, they propose two innovative strategies: Monolingual Human Preference (MLP) and Bilingual LLM Preference (BLP), with MLP evaluating based on the preferences of the monolingual readers of the target language and with BLP leveraging LLMs to directly compare the translations with the original texts. Despite lower d-BLEU scores, translations from TRANSAGENTS are preferred by both human evaluators and LLMs, particularly in genres requiring domain-specific knowledge. 

Alongside literary translation, translating culturally rich components such as names and song lyrics has been investigated. \citet{sandoval2023rose} highlight social biases in MT when translating names, particularly those associated with racial and ethnic minorities. They find significant disparities in translation quality for female-associated names from minority groups, emphasizing the need for bias mitigation in MT systems. \citet{li2023translate} tackle song translation, where lyrics must be aligned with melodies. They introduce Lyrics-Melody Translation with an Adaptive Grouping framework, ensuring that translated lyrics fit the original tune, addressing both linguistic and musical challenges between cultures. Additionally, recent studies on K-pop lyric translation further highlight the complexity of translating music while preserving both meaning and melody. For example, \citet{kim2024kpopmt} introduce a novel dataset focused on K-pop lyric translation, highlighting the need for dedicated datasets to better address the singability and cultural nuances of lyric translation. Moreover, \citet{kim2023k} tackle the challenge of translating K-pop fan terminology through the KpopMT dataset, which focuses on in-group language systems used by K-pop fandoms. This dataset shows the difficulty of translating fan-specific terms and styles, with evaluations revealing low performance from current translation systems, including GPT models. Together, these studies emphasize the need for culturally sensitive and genre-specific translation techniques.

\textbf{\textsl{Takeaways from Culturally Adapted Translation.}} Despite substantial progress in culturally adapted MT, much of the current work continues to align culture predominantly with language and nationality. Future research could delve into more nuanced levels of cultural adaptation within MT, such as tailoring translations to generational language preferences, more diverse regional dialects, or specific group terminologies. \\

\textbf{Culturally Adapted Dialogue Systems.}
Task-oriented dialogue (ToD) systems are crucial for multilingual interactions, but creating culturally adapted datasets is challenging. Early datasets were often on a small scale, lacked naturalness, and failed to capture cultural nuances due to translation-based approaches~\cite{ding2021globalwoz, hung2022multi2woz}. To overcome these issues, recent efforts, as follows, focus on generating culturally relevant dialogue data and improving language-specific model performance. 

\citet{majewska2023cross} introduce the Cross-lingual Outline-based Dialogue (COD) dataset, which utilizes a novel outline-based annotation process to create dialogues across diverse languages, covering Arabic, Indonesian, Russian, and Kiswahili while improving cultural specificity.  \citet{hu2023multi} contribute with MULTI3WOZ, a large-scale multilingual ToD dataset designed to avoid translation artifacts and ensure cultural adaptation across languages. 

To capture implicit cultural cues in dialogue systems, \citet{cao2024bridging} propose cuDialog, a benchmark that leverages cultural dimensions from the Hofstede Culture Survey. Covering 13 cultures and 5 genres, this benchmark emphasizes the importance of understanding cultural differences, such as communication styles and shared metaphors, in dialogue systems. Expanding the scope of human-like interaction,~\citet{wang2023humanoid} introduce Humanoid Agents, a system that simulates human-like behavior in dialogue agents by incorporating elements of System 1 thinking~\cite{arvai2013thinking}, such as basic needs, emotions, and relationship closeness. This allows agents to adjust conversations based on emotional states and social relationships, offering a more intuitive, adaptive framework that complements linguistic and cultural adaptation in dialogue systems.

\citet{masala2024vorbe} introduce RoCulturaBench, a dataset manually curated by a team of Romanian academics from the humanities field, addressing various significant aspects of the culture, ranging from artistic and scientific contributions to cuisine and sports. They significantly improve task performance. 

Each person's sociocultural background can affect their pragmatic
assumptions in communication~\cite{schramm1954communication}. \citet{shaikh-etal-2023-modeling} introduces the CULTURAL CODES dataset, which operationalizes cross-cultural pragmatic inference. It is based on a collaborative two-player word reference game called Codenames Duet, and includes 794 games with 7k turns, distributed across 153 unique players. They show accounting for background characteristics can improve model performance, indicating that integrating sociocultural priors can align models toward more socially relevant behavior in conversations.

\textbf{\textsl{Takeaways from Culturally Adapted Dialogue Systems.}} As demonstrated by datasets like cuDialog~\cite{cao2024bridging}, RoCulturaBench~\cite{masala2024vorbe}, and CULTURAL CODES~\cite{shaikh-etal-2023-modeling}, models that incorporate cultural dimensions and sociocultural priors show improved performance and alignment with real-world conversational contexts. However, these approaches still fall short in capturing dynamic cultural adaptation within ongoing interactions. Future work could explore adaptive dialogue systems that tailor responses in real-time, adjusting to subtle cues like shifts in tone, topic sensitivity, or cultural context, ultimately creating more contextually responsive and socially aware interactions.

\section{Vision Models and Culture}
\label{vis}

Recently, there has been increasing recognition of the significance of cultural inclusion in large models (LMs), which has inspired work on studying cultural understanding in vision language models. Vision language models have been used for a long time for tasks like image captions, VQAs, image understanding, etc. Still, increasing interest and the need to understand the model outputs have led to research directions of testing these tasks for cultural inclusiveness. However, most vision language models (VLMs) have since been predominantly trained on data from Western languages and cultures, most notable being MS-COCO \citep{lin2014microsoft}, Flickr 30K \citep{young2014image} and LAION \citep{schuhmann2022laion}, which limits their use case in non-western and low-resource languages. Additionally, cultural nuances in the images significantly affect the interpretation of the images (along with the text), making such a study very important. 

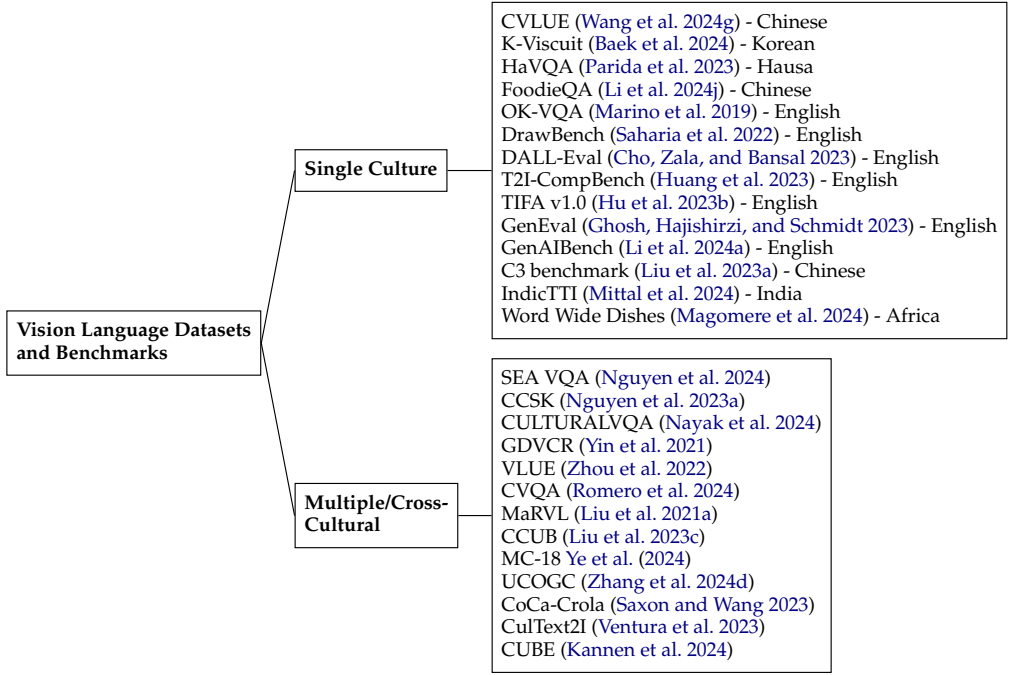
\begin{figure*}
    \centering
    \resizebox{\textwidth}{!}{
        \begin{forest}
            for tree={
                grow=east,
                reversed=true,
                rectangle,
                draw,
                align=left,
                anchor=west,
                parent anchor=east,
                child anchor=west,
                font=\scriptsize,
            }
            [\textbf{Vision Language Datasets}\\\textbf{and Benchmarks}
                [\textbf{Single Culture}, tier=1
                    [CVLUE \citep{Wang2024CVLUEAN} - Chinese \\
                    K-Viscuit \citep{Baek2024EvaluatingVA} - Korean \\
                    HaVQA \cite{parida2023havqa} - Hausa \\
                    FoodieQA \citep{li2024foodieqa} - Chinese \\
                    OK-VQA \cite{marino2019ok} - English \\
                    DrawBench \cite{saharia2022photorealistic} - English\\
                    DALL-Eval \cite{cho2023dall} - English\\
                    T2I-CompBench \citep{huang2023t2i} - English\\
                    TIFA v1.0 \citep{hu2023tifa} - English \\
                    GenEval \citep{ghosh2024geneval} - English\\
                    GenAIBench \citep{li2024genai} - English\\
                    C3 benchmark \citep{liu2023cultural} - Chinese\\
                    IndicTTI \citep{mittal2024navigating} - India \\
                    Word Wide Dishes \citep{magomere2024you} - Africa
                    , tier=leaf
                    ]
                ]
                [\textbf{Multiple/Cross-}\\\textbf{Cultural}, tier=1
                    [SEA VQA  \citep{nguyen-etal-2024-culturax}\\
                    CCSK \citep{nguyen2023extracting} \\
                    CULTURALVQA \citep{nayak2024benchmarking} \\
                    GDVCR \citep{yin2021broaden} \\
                    VLUE \citep{zhou2022vlue} \\
                    CVQA \citep{romero2024cvqa} \\
                    MaRVL \citep{liu2021visually} \\
                    CCUB \citep{liu2023towards}\\
                    MC-18 \citet{ye2024altdiffusion} \\
                    UCOGC \citep{zhang2024partiality} \\
                    CoCa-Crola \cite{saxon2023multilingual} \\
                    CulText2I \citep{ventura2023navigating}\\ 
                    CUBE \citep{kannen2024beyond}, tier=leaf
                    ]
                ]
            ]
        \end{forest}
    }
    \caption{Cultural Datasets and Benchmarks in Image-based Multimodal Tasks}
    \label{fig:vlm-benchmarks}
\end{figure*}

To address these challenges and to develop VLMs that can effectively comprehend the cultural contexts of different countries, two directions are commonly approached in the literature: a) establishing a comprehensive test benchmark across culture-specific tasks \citep{liu2021visually,romero2024cvqa}; and b) proposing new and detailed culture specific datasets which can be tested or probed for cultural details. Both these directions are vital, as we need better quality culturally aware datasets and tasks or benchmarks to assess models' capabilities to accurately interpret and respond to culturally specific inputs. Notably, prior research has tried to create VLM test benchmarks tailored to particular countries. We divide the current literature into two parts: language output (includes tasks such as image captioning, visual question answering, etc.) in \S\ref{langout} and image output (image generation) in Section \S\ref{visout} and study the nuances that literature has covered, to make these tasks culturally aware. We highlight the representative single-culture and multi-culture benchmarks in Figure \ref{fig:vlm-benchmarks}.

\subsection{Language Output Tasks}
\label{langout}


Language output tasks are the ones that have a language decoder as the output module and a mixed vision-language encoder to process either input. To have a culturally aware model for language output tasks, the vision and language encoder should be able to identify cultural concepts within the input text and image, and the decoder should generate culturally relevant outputs. Since humans from different cultures often perceive and interpret images differently, models should be designed to reflect these varying perspectives \citep{Jahoda1974PictorialDP}. \citet{Ye2023ComputerVD} argue that people from different cultural backgrounds observe vastly different concepts even in the same images, and multilingual datasets have more semantic content than monolingual datasets, accommodating this diversity. 
To further investigate this diversity; the research community has developed benchmarks to test the cultural capability of these models. In the following sections, we look into these benchmarks in the context of two language output tasks: a) image captioning and b) Visual Question Answering (VQA).

\subsubsection{Visual Question Answering and Related Tasks}

VQA \citep{antol2015vqa} is a task that requires model knowledge to answer textual questions based on a given context image. This task is essential for assessing a model's reasoning ability across visual and textual domains. Most existing VQA benchmarks are limited to the English language \citep{marino2019ok}. \citet{Ananthram2024SeeIF} show that VLMs have Western biases across subjective and objective visual tasks with culturally diverse images and annotations; they argue that while multilingual prompting can somewhat mitigate the bias, a more diverse pretraining mix is a more suitable and effective solution for mitigating the Western bias. There have been recent attempts to expand this task to multilingual VQA \citep{tang2024mtvqa}. These works have shown the lack of nuanced cultural understanding, making this an open research question. Benchmarks can be specific to a single culture or across many cultures. The former can dive deep into finer and unique elements of a single culture (e.g., food eaten at a specific festival in a country); the latter is helpful when we want to evaluate universal elements that are understood differently in different cultures (e.g., how clothing exists universally but looks different across cultures). In the following sections, we discuss these specific and multicultural benchmarks in more detail.

\textbf{Dataset Creation Pipelines.} Dataset creation for cultural assessment of models can be done by creating data from the web (scraped from the web with noisy web annotations), using the human-in-loop method (e.g., human filter data collected from the web and manually annotating them) or giving control to humans to create data (e.g., taking pictures from the surroundings) 

\citet{Baek2024EvaluatingVA} propose a semi-automated pipeline for constructing cultural VLM benchmarks, where they demonstrate the usability of the pipeline by constructing a dataset tailored to Korean culture: K-Viscuit. This pipeline uses human-VLM collaboration, where VLMs generate questions based on guidelines, human-annotated examples, and image-wise relevant knowledge, which native speakers then review for quality and cultural relevance. \citet{nguyen2023extracting} create a VQA dataset for Vietnamese culture with 33,000+ pairs of question-answer over three languages: Vietnamese, English, and Japanese, on approximately 5,000 images; the QA-pairs are first generated in Vietnamese and then translated to Japanese and English manually to study cross-cultural perspectives.

Some works, such as Multicultural Reasoning over Vision and Language (MaRVL) \citep{liu2021visually}, start with culturally relevant concepts and objects sourced from native speakers. Then the native speakers are asked to find relevant images to the concepts. On top of the concepts and images obtained through the process,  statements are elicited from native speaker annotators about pairs of images. Along similar lines as MaRVL, \citet{Wang2024CVLUEAN} start by defining object categories related to culture (Chinese) and then collect images related to each object category; they start out with categories used in the MaRVL paper but remove the ones that are not relevant to Chinese culture, and add a few relevant ones. \citet{zhou2022vlue} create a new benchmark consisting of 5 tasks for evaluating the generalization capabilities of vision language models and use MaRVL as one of the Out-Of-Distribution test sets.
\citet{romero2024cvqa} proposes a culturally diverse multilingual Visual Question-answering benchmark that covers 28 countries on four continents, covering 26 languages with 11 scripts, and involves native speakers as well as cultural experts in the data collection process; the native annotators were asked to source images from popular open-use licensing sources such as Flickr, GapMinder, Unsplash, Pixabay as well as personal photos.

\textbf{Training Methodologies and Models.}
Addressing cultural diversity in VQA has become a critical challenge as models often struggle with context-dependent interpretations, especially when cultural knowledge is required. To study context-dependent interpretations, \citet{bongini2020visual} created a VQA dataset based on cultural heritage images from the Artpedia \citep{stefanini2019artpedia} dataset. Their methodology involves a module that detects whether a question requires cultural context, followed by gathering relevant external knowledge. This approach highlights the need for models to incorporate cultural awareness, often missing in conventional VQA tasks. Building on this, other researchers have focused on reducing biases and improving cultural equity in VQA. \citet{yin2023givl} introduced new pre-training objectives that explicitly model differences in visual concepts across regions. By addressing biases against underrepresented groups, they aim to ensure more equitable performance across diverse geographical areas. Another significant development comes from \citet{li2023cultural}, who propose an annotation-free method for adapting visual cultural concepts. Their work constructs a concept mapping set and leverages high-resource cultures to help models understand low-resource ones, making it easier for models to generalize across cultural contexts without requiring extensive manual annotation. They also propose a multimodal data augmentation technique, CultureMixup, which mixes cultural concepts in images to enhance the model’s ability to reason visually across languages and cultures. Finally, \citet{nguyen2024multi} demonstrate that the inclusion of translated multilingual data in training improves the performance of models on geographically diverse tasks such as GeoDE, further emphasizing the importance of cultural and linguistic diversity in building robust VQA models.

\textbf{Mono-Cultural Benchmarks and Multi-cultural Benchmarks.}
Benchmarks can be specific to a single culture or across many cultures. The former can dive deep into finer and unique elements of a single culture (e.g., food eaten at a specific festival in a country); the latter is helpful when we want to evaluate universal elements that are understood differently in different cultures (e.g., how clothing exists universally but looks different across cultures). In the following sections, we discuss these specific and multicultural benchmarks.

Culture-specific benchmark involves creating tasks ranging from image-text retrieval to visual question answering, visual grounding, and visual dialogue, as the data collection methodology varies a little across the tasks.  Works have been done on creating culture-specific benchmarks, such as CVLUE \citep{Wang2024CVLUEAN} for Chinese, K-Viscuit \citep{Baek2024EvaluatingVA} for Korean, \citet{parida2023havqa} for the Hausa language, etc. While creating culture-specific benchmarks, images are either manually curated (e.g., asking annotators to find relevant images on the web) as in CVLUE, K-Viscuit, or sources from already existing datasets such as Visual Genome as in \citet{parida2023havqa}. FoodieQA \citep{li2024foodieqa} is another recent work with a manually curated, fine-grained image-text dataset capturing the intricate features of food cultures across various regions in China. These benchmarks highlight the presence of sub-cultures within a culture that other generalized benchmarks might miss.

Most VQA benchmarks explicitly testing VLMs' cultural awareness are multicultural. These benchmarks highlight the lack of multicultural perspectives amongst the current VLMs.  Some benchmarks, such as CULTURALVQA~\cite{nayak2024benchmarking}, have questions that probe for the understanding of various facets of culture, such as clothing, food, drinks, rituals, and traditions across various countries; benchmarking various VLMs on CULTURALVQA, reveals disparity in their
level of cultural understanding across regions,
with strong cultural understanding capabilities
for North America while significantly lower
performance for Africa. \citet{yin2021broaden} construct a Geo-Diverse Visual Commonsense Reasoning dataset (GD-VCR) to test VLMs’ ability to understand cultural and geo-location-specific commonsense; the benchmark is based on TV series and movies across countries from four regions: Western, East Asian, South Asian, and African countries.

\textbf{Evaluation Frameworks and Metrics.}
While most work has been on creating benchmarks that measure cultural awareness, there has been little on creating frameworks for measuring cultural alignment. \citet{kannen2024beyond} introduce a framework to evaluate the cultural competence of VLMs along dimensions such as cultural awareness and cultural diversity, along with an approach to construct and build a large dataset of cultural artifacts to enable evaluation along these dimensions. \citet{Baek2024EvaluatingVA} propose a human-VLM collaboration pipeline, where VLMs generate questions based on guidelines, human-annotated examples, and image-wise relevant knowledge, which are then reviewed by native speakers for quality and cultural relevance. 

In a nutshell, these works highlight the growing recognition that cultural awareness, bias reduction, and multilingual data are essential for advancing VQA systems that can reason effectively across diverse contexts.


\subsubsection{Image Captioning}
Culturally aware image captioning includes recognizing the cultural context of the image (cultural relevance of objects, recognizing culture-specific objects, etc.) and describing the image based on the cultural context.
\citet{burda2024culturally} compares the performance of various vision-language models (GPT-4V, Gemini Pro Vision, LLaVA, and OpenFlamingo) on identifying culturally specific information in images and creating accurate and culturally sensitive image captions. They define a new evaluation metric, Cultural Awareness Score (CAS), to measure the degree of cultural awareness in image captions and provide a dataset of 1.5k, labeled with ground truth for images containing cultural background and context.
\citet{cao2024exploring} probe GPT-4V using the MaRVL \citep{Liu2021VisuallyGR} benchmark, aiming to investigate its capabilities by using variations of image captioning viz. caption classification, pairwise captioning, and culture tag selection, and they note that GPT-4V can identify more cultural concepts than humans but has lower performance than humans when generating captions in low resource languages. \citet{Ye2023ComputerVD} find that multilingual descriptions have on average 29.9\%
more objects, 24.5\% more relations, and 46.0\% more at-
tributes than a set of monolingual captions and make a case for having multilingual captions for better cultural inclusion. \citet{Yun2024CICAF} propose a Culturally-aware Image Captioning (CIC) that generates captions and describes cultural elements extracted from cultural visual elements in culture-specific images. \citet{thapliyal-etal-2022-crossmodal} start with a set of 36 languages (which have a high web coverage) for captioning, then sample images from a geo-localized Open Images dataset~\cite{kuznetsova2020open} using an algorithm that maximizes the percentage of selected images taken in an area where the assigned language is spoken.

Some works extend board games and tests that are used to assess cultural awareness among humans to an image captioning tasks for LLMs. For example, \citet{kunda2020creative} suggest games such as Dixit Board game and its variants \citep{bekesas2018cosmocult}, which involve generating creative captions, could be played between VLM agents to access cultural understanding of each agent.

\textbf{Captioning-pecific Cultural Elements.} \citet{Ma2023Food500CA} refer to existing literature on food datasets and creates a new food dataset that spans across various geographical regions and presents a case for in-domain generalization in VLMs rather than out-of-domain generalization and tailoring the VLMs to specific elements. Multiple works also look at the offensiveness of memes \citep{liu2022figmemes} and changes in offensiveness and annotation of memes based on culture \citep{sap-etal-2022-annotators, pramanick-etal-2021-detecting}.


\textbf{\textsl{Takeaways from \S\ref{langout}.}}
Multiple benchmarks for VQA, visual reasoning, and captioning have been created, each varying in scope, diversity, and cultures they cover. Some cover a broad spectrum of cultural elements, while others focus on specific cultural elements, like food, in-depth. There is also variation in approaches to geo-diversity. Some studies ensure that geo-diverse annotators label similar images, whereas others incorporate both geo-diverse annotators and images. Moreover, researchers also have different assumptions about cultural diversity. Some link geographic diversity to cultural diversity, while others use linguistic diversity as a proxy. Many datasets enhance existing image collections with region-specific annotations by local annotators, while others gather culturally specific images directly from the web, providing a rich source of contextually relevant visuals. The research community could benefit from consistent methods for studying cultural diversity using benchmarks and from having common standards for measuring cultural understanding in visual tasks. 

To expand benchmarks to new cultures, some studies use culturally adapted translation (e.g. machine translation of texts), recognizing that identical objects may carry different cultural meanings. However, using local annotators (who understand the language of the culture being studied) can reduce biases introduced by translation, providing more authentic cultural insights.

\subsection{Image Output Tasks}
\label{visout}
Image Output tasks are the ones that have an image component as the output. They broadly fall into these categories: a) Text-to-image generation (T2I); b) Text-based image manipulation; and c) Image in-painting via textual prompts. The models used for these image output tasks do not have an explicit ``decoder'' in the traditional sense (like in LLMs). The process of image generation is usually handled by non-transformer decoders, such as diffusion models \citep{ramesh2022hierarchical, saharia2022photorealistic, rombach2022high, zhang2017stackgan}, GANs\citep{xu2018attngan}, or variational autoencoders\citep{razavi2019generating}. Initial years of research focused on image generation quality and prompt image alignment. More recently, it has become evident that cultural diversity in generated images remains a significant gap, prompting recent efforts to develop culture-specific approaches, metrics, and benchmarks to ensure more inclusive and contextually aware outputs.

\textbf{Mono-cultural and Multicultural Benchmarks.} 
Generating culturally specific images is a challenging task that requires not only that a model produce coherent images but also incorporate culture-specific themes, styles, and contexts. This issue is amplified by the fact that T2I models are limited by the scarcity of languages they are trained on, leading to bias in the generation of cultural elements. 
Various works have discussed these biases specific to single cultures. \citet{liu2023cultural} discuss this gap in Chinese in the context of the generation of relic images, \citep{magomere2024you} for African food culture from 5 countries. \citet{jha2024visage} take it beyond a singular culture and study these biases and stereotypes of people across various countries, e.g., Omani, Ukrainian, Swiss, Canadian, Mongolian, Indian, Australian etc. On similar lines, \citet{zhang2024partiality} investigated cultural representativeness, but unlike other works that select representatives by geography, they work with what they call ``cultural clusters'' (e.g., Latin-American, Latin-European, Middle Eastern, Nordic, etc.) and choose 3 countries with the largest populations to represent these cultural clusters; they found homogenization of some data in T2I models, especially in disadvantaged cultures (e.g., from Africa). \citet{bansal2022well} studied them from an ethical perspective and observed changes in image generations conditional on ethical interventions. They study image generations on three social axes – gender, skin color, and culture and found that models can generate images of diverse groups with prompts containing ethical interventions (e.g., by using keywords like ‘irrespective of gender’ for gender bias and ‘culture’ for cultural bias). \citet{ventura2023navigating} take this even further and study these cultural embeddings across three tiers: cultural dimensions, domains, and concepts; they also propose the CulText2I dataset consisting of images generated by six distinct TTI models for evaluating these axes. As cultures are region-specific, work has been done to fine-tune these models with datasets curated to represent culture-specific concepts, e.g., artwork, landmarks, and artistic styles of a culture. E.g. \citet{amadeus2024pampas} fine-tuned DreamBooth \citep{ruiz2023dreambooth}, a T2I model, to evaluate the model's understanding of regionalism, culture, and historical value of the state of RS, Brazil. \citep{deng2024research} fine-tuned the Stable Diffusion model using the Low-Rank Adaptor (LoRA) to generate historical Chinese Artifacts. However, as much as it is desired, it is not always possible to create a model for each culture.

\textbf{Culture-pecific T2I models.} There is a need for either a) a more robust model trained with multilingualism and aligned for cultural concepts or b) better architectural approaches to make models culturally inclusive. \citet{ye2024altdiffusion} attempted to address the first gap by training a multilingual T2I model, trained on 18 languages, and showed that their model outperformed Stable Diffusion in generating culture-specific concepts. \citet{liu2023towards} and \citet{zhang2021m6} attempt to close the second gap. Where \citet{liu2023towards} proposed a new approach to making a model culturally inclusive by pre-training the T2I synthesis model and adding semantic context using their Cross-Cultural Understanding Benchmark (CCUB) Dataset, M6-UFC \citep{zhang2021m6} extend the transformer-based architecture to generate culturally diverse images conditioned on the context provided by text prompts in regional languages (in this case Chinese). Each research either chooses one of the T2I models or multiple of them but thee is no standard list of models for comparison. Popular ones include both open source and closed source models e.g. DALE·E, Stable Diffusion, Imagen. \citep{basu2023inspecting}, while investigating for geographical representativeness of generated images on 27 countries in two popular T2I models (DALL·E and Stable Diffusion), observed that DALL·E-2 was more representative of the cultural artifacts when using country-specific prompts, as compared to Stable Diffusion showing that closed source models may have slightly better cultural alignment than open source T2I models. During their human evaluation, they found that when the input prompt did not include any specific country name, users from 25 out of 27
countries felt that the generated images were less representative of the country-specific artifacts.




\textbf{Evaluation Metrics.} Unlike LLMs, in T2I tasks, it is hard to develop a standard evaluation task that is objective. Evaluating image generation models involves a variety of metrics that can assess quality, coherence, diversity, and alignment with textual prompts. For example, while GPT-3 was introduced with impressive zero-shot performance across many classification tasks, DALL·E-2, OpenAI's T2I model \citep{ramesh2022hierarchical} was shown to have good ``human opinion scores''. As T2I models have become increasingly better in image quality, many metrics have been proposed to evaluate these models.  Most of these metrics are qualitative (looking at the images and evaluating if they are correct representations of culture), though there is an increasing amount of work to set up qualitative metrics. Fréchet Inception Distance (FID) \citep{yu2021frechet} and Inception Score (IS) are the most popular and are commonly used to measure the visual quality and diversity of generated images against real-world distributions. CLIP Score, based on CLIP model \citep{radford2021learning}, is also used to evaluate the coherence between generated images and text inputs. However, these metrics do not cover culture-specific nuances. \citet{struppek2023exploiting} propose to measure bias in these T2I models by showing that image generations are skewed by simply inserting single non-Latin characters in a textual description. They rely on 3 metrics to measure cultural biases, 2 for studying generated images and 1 for prompts used. \citet{kannen2024beyond} showed that measuring cultural awareness and cultural diversity is important for a framework to evaluate the cultural competence of T2I models. 

\textbf{Evaluation Benchmarks.} More recently, there have been attempts at developing T2I benchmarks considering cultural nuances. \citet{zhang2024partiality} who discuss cultures as ``cultural clusters'', built Unique Cultural Objects from Global Clusters (UCOGC) dataset as an evaluation benchmark for the diversity of T2I models. As this benchmark covers both material and nonmaterial cultural subjects in both comprehensiveness and diversity, it's a good benchmark for evaluating the quality of generated culture representativeness in T2I models. Other recent benchmarks include T2I-CompBench \citep{huang2023t2i}, TIFA v1.0 \citep{hu2023tifa}, and GenEval \citep{ghosh2024geneval}, and GenAIBench \citep{li2024genai} which leverage diverse prompts and metrics to evaluate aspects such as image-text coherence, perceptual quality, attribute binding, faithfulness, semantic competence, and compositionally. We also have Drawbench \cite{saharia2022photorealistic} and DALL-Eval \cite{cho2023dall}, which aim for comprehensiveness in benchmarks. Where DrawBench proposes the evaluation of various categories (colors, numbers of objects, spatial relations, text in the scene, and unusual interactions between objects); DALL-Eval proposes the evaluation of visual reasoning (object counting, VQA etc.) as well as social bias (gender, colour etc.). More recently, \citet{saxon2023multilingual} question these benchmarks because though they all aim for different goals, it is challenging to determine if these benchmarks accurately represent the practical tasks expected of the model within real-world contexts. The proposed CoCa-Crola benchmark uses 3 distinct metrics, Distinctiveness, Self-Consistency, and Correctness, as a technique for benchmarking the degree to which any generative text-to-image system provides multilingual parity to its training language in terms of tangible nouns. \citet{liu2023cultural} propose a C3 benchmark to study cultural relevance and image quality and propose evaluation on 6 metrics: cultural appropriateness, object presence, object localization, semantic consistency, visual aesthetics, and cohesion. Unlike \citet{saxon2023multilingual}, who focus on generating simple concepts through translation, \citet{mittal2024navigating} focus on prompts describing multiple elements in the generated image. They investigate bias in T2I models in 30 Indic languages on Stable Diffusion, Alt Diffusion, Midjourney, and Dalle3 and evaluate on 4 proposed metrics: Cyclic Language-Grounded Correctness (CLGC), Language-Grounded Correctness (LGC), Image-Grounded Correctness (IGC) and Self-Consistency Across Languages (SCAL).
 

 
\textbf{\textsl{Takeaways from \S\ref{visout}.}} Although there has been increasing work on evaluating large multimodal models for cultural awareness, there is a greater need for models trained with balanced multilingual and culture-specific data to ensure solid multilingual and cultural capabilities. Additionally, there is no standardized list of evaluation methods, with each study selecting the methods independently. Metrics for image output tasks, such as measuring cultural awareness in VQA and T2I models, are also not standardized and remain a significant future direction to explore. Overall, developing consistent metrics to test cultural awareness in text-to-image models remains a significant future direction that could be explored.

\subsection{Art Forms Related Tasks}
\label{sec:art}

Art forms and paintings evoke different emotions across different cultures and have been considered used by the community to study the expression of emotions across cultures \citep{mohammad-kiritchenko-2018-wikiart, 9577962}. There have been multiple studies on art and generating art using Vision models recently. One of the main goals when studying and examining art forms is to match the objects in an image to their symbolic meaning. \citet{zhang2023m3exam} create a dataset for a dataset for art
understanding deeply rooted in traditional Chinese culture; they address three tasks: identifying salient visual elements, matching elements with their
symbolic meanings, and explanations for the conveyed message. \citet{hamilton2021mosaic} create a web application named MosAIc, that allows users to find pairs of semantically
related artworks that span different cultures, media, and millennia; they use Conditional Image Retrieval (CIR), which combines visual similarity search with user-supplied filters or ``conditions''. To study the similarity between arts across cultures and evaluate cultural-transfer
performance, \citet{mohamed2022artelingo} creates a dataset of 80k artworks, with many artworks being annotated by multiple people in three languages. \citet{Fan2023CICHMKGAL} create a multimodal knowledge graph linking visual entities and concepts associated with the entities. \citet{zhang2024cultiverse} addresses the challenge of translating the nuanced symbolism in art, which involves interpreting complex cultural contexts, aligning cross-cultural symbols, and validating cultural acceptance. \citet{ozaki2024towards} create a dataset of artworks and explanations in multiple languages with nuances and country-specific phrases.

\subsection{Miscellaneous Tasks}
The following works introduce new tasks, including variants of image captioning, image classification, or somewhere in between, to better access the cultural understanding in Vision language models. \citet{buettner2024incorporating} attempts to improve object recognition models to be more robust to objects from geographically diverse regions. M5 \citep{schneider2024m}, for example, collects data for 12 languages to pair it with photos from the regions that speak those languages. The authors then create a benchmark for tasks such as visually grounded reasoning, visual question-answering
(VQA), visual natural Language
inference (VNLI), visio-linguistic Outlier
detection (VLOD), and captioning. They introduced new novel benchmarks, such as M5-VGR and M5-VLOD, including a new Visio-Linguistic Outlier Detection task. The images for M5 are sources from the Dollar Street dataset \citep{NEURIPS2022_5474d9d4}, comprising around 38K photos taken in 63 different regions or countries. These photos depict the lives of families, including their homes, neighborhoods, or everyday objects, in a culturally diverse way. There is no explicit paring between languages/cultures and images.  \citet{pappas2016multilingual}
conduct a crowdsourcing experiment to annotate the sentiment score of visual concepts from 11 languages associated with 16,000 multilingual visual concepts. The MVSO dataset \citep{jou2015visual} is used as the source of visual concepts, and the photo-sharing service Flickr is used as the source of images. \citet{zhang2024can} create a dataset that spans 30 countries almost 2 centuries; their goal is to test if VLMs can identify cultural markers required to determine the time and place a photo was taken. On similar lines, \citet{hsiao2021culture} provide a data-driven
approach to identify specific cultural factors affecting the clothes people wear where they use news articles and vintage photos spanning a century to create a model that detects influence relationships between happenings in the world and people’s choice of clothing. \citet{li2022multi}  construct a multimodal knowledge graph for classical Chinese
poetry (PKG), in which the visual information
of words in the poetry are incorporated for the task of poverty image retrieval. \citet{zhang2024aligning} propose a preference-based reinforcement learning method that fine-tunes the vision models to distill the knowledge from both LLMs reasoning and the aesthetic models to better align the vision models with human aesthetic standards, which vary with culture. \citet{khanuja2024image} introduces a new task
of translating images to make them culturally
relevant by changing concepts in an image that varies with culture.\\

\textbf{\textsl{Takeaways from \S\ref{vis}.}} Apart from takeaways mentioned in specific subsections, Some papers model cultural change in images across time, an important aspect missed when using images directly from the internet as a source. Most of the vision papers do assume that a language implies a culture, but they use the assumption that a geographical region corresponds to a culture. Automatic data scraping methods that rely on getting culture-specific images based on the language of the caption (from sources like Wikipedia) can lead to language culture bias, where multiple cultures sharing the same language may be merged into a single, undifferentiated culture.
\section{Other Modalities and Culture}
\label{oth}

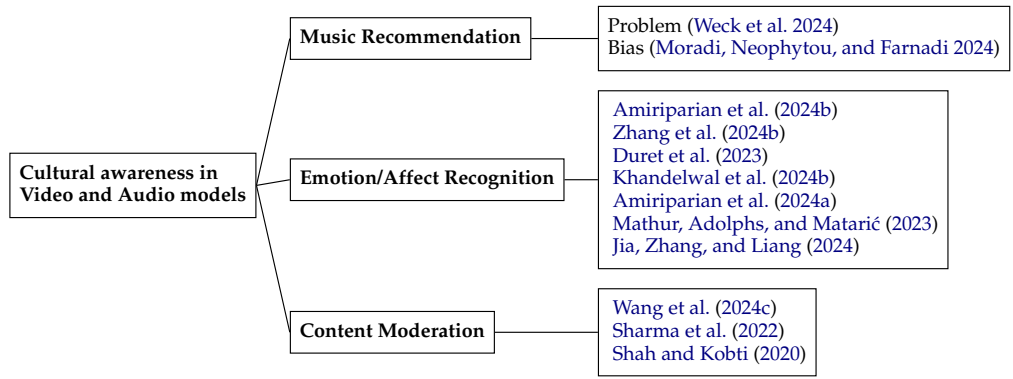
\begin{figure*}
\centering
\resizebox{\textwidth}{!}{
\begin{forest}
for tree={
  grow=east,
  reversed=true,
  rectangle,
  draw,
  align=left,
  anchor=west,
  parent anchor=east,
  child anchor=west,
  font=\scriptsize,
}
[\textbf{Cultural awareness in}\\\textbf{Video and Audio models}
    [\textbf{Music Recommendation}, tier=1
        [Problem~\cite{weck2024muchomusic}\\
        Bias~\cite{moradi2024advancing}
         ,tier=leaf
         ]
    ]
    [\textbf{Emotion/Affect Recognition}, tier=1
        [~\citet{amiriparian2024exhubert}\\
        ~\citet{zhang2024speechagents}\\
        ~\citet{duret2023enhancing}\\
        ~\citet{khandelwal-etal-2024-indianbhed}\\
         ~\citet{amiriparian2024muse}\\
         ~\citet{mathur2023towards}\\
         ~\citet{jia2024bridging}
         ,tier=leaf
         ]
    ]
    [\textbf{Content Moderation}, tier=1
        [~\citet{wang2024multihateclip} \\
         ~\citet{sharma2022detecting}\\
         ~\citet{Shah2020MultimodalFN}
         ,tier=leaf
         ]
    ]
]
\end{forest}
}
\caption{Areas explored for cultural adaptation in video and audio modalities, with representative examples. Music recommendation involves only the audio modality, emotion/affect recognition has been explored in both audio and video modalities, and content moderation has been explored for videos and memes.}
\label{fig:oth}
\end{figure*}

In this section, we include papers that look at cultural adaptation in other modalities, such as videos, audio, etc., tasks that do not fall under the text-only or vision-language [text+images] tasks but have text (semantic content) as one of the components. We highlight major areas and representative papers in figure \ref{fig:oth}.

\subsection{Audio and Speech}
\label{aud}

While understanding the cultural context in music and speech may not always involve text (semantic content) as one of the component major components, music recommendations require understanding the culture-specific preferences of the user as well as the cultural context in the query \cite{weck2024muchomusic}. \citet{moradi2024advancing} study cultural biases in music recommendation systems and provide a method to improve fairness in music recommendation systems. \citet{li2024survey} argue that understanding the cultural context should be one of the goals that should be prioritized while building Foundation Models for music. 

One of the applications where understanding data streams such as speech becomes important along with semantic content is emotion recognition, as the expression of emotions varies across cultures. \citet{belani2022automatic} study the relation between emotional expression and code-switching for Spanish. \citet{amiriparian2024exhubert} gather a comprehensive multilingual, multicultural speech emotion corpus with 37 datasets, 150,907 samples, and a total duration of 119.5 hours. \citet{tran2023personalized} demonstrate personalized and adapted speech encoders for continuous emotion recognition.  \citet{Sapinski2015EmotionRF} look at emotion detection based on audio characteristics and the semantic content in tandem.  \citet{zhang2024speechagents} propose a multi-modal LLM-based multi-agent system designed for simulating human communication along with rich emotions expressed through speech and semantic content (text). As the research in textless speech-to-speech translation continues to grow, it is important to ensure that expressions and emotions are translated (and mapped) correctly across languages. \citet{duret2023enhancing} propose a method to enhance expressivity transfer in textless speech-to-speech translation. \citet{wunarso2017towards} create a dataset for speech-emotion detection for Indonesian. 

There have been a few works that look at the presence of subculture within a border culture and collect data to understand cultural nuances. \citet{javed2024indicvoices} collect a dataset (INDICVOICES) of natural and spontaneous speech covering 16237 speakers covering 145 Indian districts and 22 languages to capture the cultural, linguistic, and demographic diversity of India. SEACrowd\cite{lovenia2024seacrowd} carry similar efforts and collect data for 1000 Southeast Asia (SEA) languages spanning 3 modalities, with one of the goals being reducing cultural misrepresentation and flattening.

\textbf{\textsl{Takeaways from \S\ref{aud}.}}  While most relevant works in the audio domain focus on emotion recognition and music recommendations, 
there has been a lack of works that simultaneously model audio and text (semantic content) to understand the cultural context part from emotions. This capability could be useful for applications such as voice assistants. Works such as IndicVoices~\cite{javed2024indicvoices} and SEACrowd~\cite{lovenia2024seacrowd} are some initial efforts in the direction of collecting culturally diverse speech-text data. One of the limitations of datasets such as SEACrowd and IndicVoices is that they collect speech data in a controlled setup, typically by asking questions and recording responses, which may not accurately capture the nuances of everyday conversations.

\subsection {Video}
\label{vid}

In the case of video modality, most work has been task-specific cultural adaptation, focused mainly on emotion detection and content moderation. Cultural factors also affect personality \cite{walker2011universals} and how people interact in certain situations, \citet{khan2020vyaktitv} create a multimodal dataset of peer-to-peer Hindi conversations to study the variance of personality with factors such as income and cultural orientation. \citet{funk2024multilingual} studies how culture affects the non-verbal features (such as facial expressions and tone) of the speakers during conversations. \citet{amiriparian2024muse} create a dataset for the cultural humor detection challenge, which focuses on cross-lingual and cross-cultural multimodal humor detection, as humor detection depends not only on words but also on gestures and facial expressions. SEWA DB \cite{kossaifi2019sewa} is a dataset of conversations between people coming from different cultures during various social situations; the dataset contains videos annotated with facial landmarks, facial action units (FAU), various vocalizations, mirroring, and continuously valued valence, arousal, liking, agreement, and prototypic examples of (dis)liking. The AVEC challenge through the years has looked at (and created datasets) for cross-cultural emotion detection \citep{ringeval2019avec}. Detecting emotional cues (affect recognition) is an important part of Human-Computer Interaction systems, \citet{mathur2023towards} study intercultural affect recognition models using videos of real-world dyadic interactions from six cultures. \citet{zhao-etal-2022-m3ed} create a multi-modal, multi-scene, multi-label emotional dialogue dataset from 56 TV series capturing Chinese culture. \citet{migon2019detecting} presents a video analysis application to detect personality, emotion, and cultural aspects of pedestrians in video sequences, along with a visualizer of features (the features include elements of well-known frameworks such as Hofstede Cultural Dimensions). The works studying the effect of culture on emotional (body) gestures and the speech uttered during the gestures have been reviewed in \citet{noroozi2018survey}. \citet{jia2024bridging} propose a multimodal strategy for emotion recognition based on facial expressions, voice tones, and transcripts from video clips. \citet{liu2024funnynet} propose a multimodal approach based on transcripts, video-frames, and audio for detecting funny moments in video clips of tv-series. \citet{bruno2019knowledge} presents a case for embedding cultural knowledge into personal robots, as home activity recognition can be improved using cultural knowledge is used \citep{menicatti2017modelling}. \citet{Rehm2009} provides guidelines for creating video Recordings of multimodal interactions across cultures.

With the rise in online video-sharing platforms, multimodal hate speech detection has become an integral part of content moderation \citep{hee2024recent}. \citet{wang2024multihateclip} create a multilingual
dataset of videos annotated for hatefulness, offensiveness, and normalcy and argue that the dataset provides a cross-cultural perspective on gender-based hate speech. The rise in memes as a source of information sharing \citet{shifman2013memes} has also fueled interest in automatically detecting harmful and biased memes \citep{sharma2022detecting}. As memes marked as normal by one culture can be offensive to others, understanding the cultural context in the memes becomes necessary \citep{hegde2021images}. \citet{Shah2020MultimodalFN} propose a methodology that uses situational and normative knowledge to detect fake news using text and images. \citet{lyu2023gpt} use GPT-4V for hate-speech detection in multimedia using cultural insights; they also look at multimodal sentiment analysis in cultural context using GPT-4V.

\textbf{\textsl{\textsl{Takeaways from \S\ref{vid}.}}}
The analysis of cultural nuances in day-to-day interactions between people and videos of day-to-day activities is often missing, while the major focus is memes, personality, and hate speech. Videos can be an important source of cultural information. Although there have been works looking at information extraction from videos \citep{an2023vkie}, extracting culture-specific information from videos can be an important next step. There has been a rise in video generation models \citep{ho2022imagen}, and how people perceive a video also depends on cultural background \citep{Scott2015ModellingHF}, so the video generation models should consider cultural context as one of the input features for the video generation model.

\section{Language and Region Coverage}
\label{coverage}

\begin{figure}[]
  \centering
  \subfigure[Distribution of the language]{
    \includegraphics[width=\linewidth]{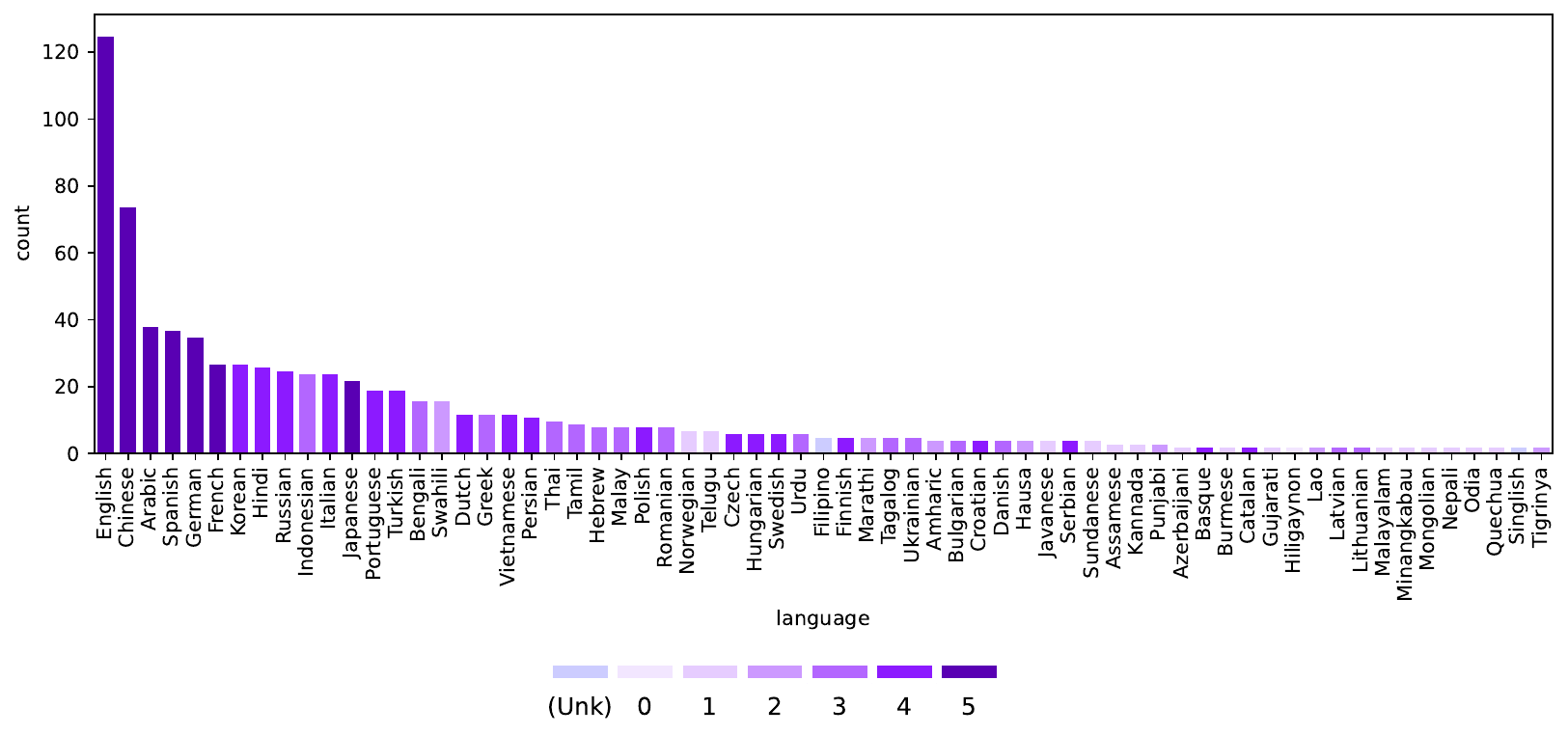}
    \label{fig:lang-dist}
  }
  \hspace{1cm}
  \subfigure[Distribution of the countries]{
    \includegraphics[width=0.8\linewidth]{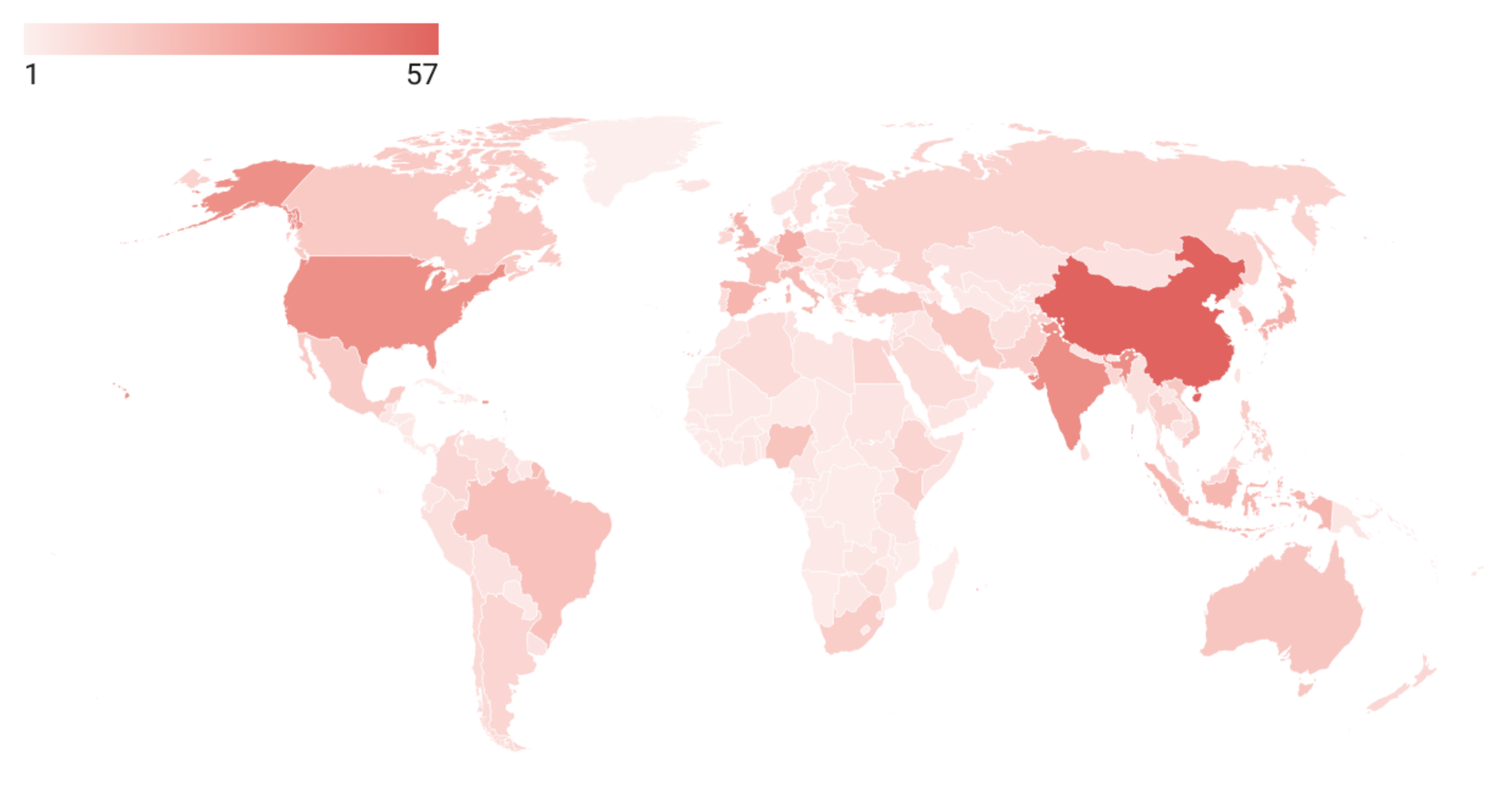}
    \label{fig:country-dist}
  }
  \caption{Distribution of the languages and the countries covered in model evaluations. (a)~The colors represent the language resource classes from \citet{joshi-etal-2020-state}. The plot includes only languages that appear two or more times.}
\end{figure}


We manually annotated the languages and regions covered by benchmarks and evaluations presented in \S\ref{eval} and benchmarks in \S\ref{vis} and \S\ref{oth}.
For languages with multiple names (e.g., endonyms) or variations (e.g., dialects), we standardize them based on the conventions in \citet{joshi-etal-2020-state}. For regions, we visualize the distribution at the country level, including only countries explicitly mentioned by the authors in their manuscripts. We chose country-level analysis since most papers emphasize national cultures. Sub-regional cultures were aggregated under their respective countries, and broader regions were excluded from our analysis.

Figure~\ref{fig:lang-dist} presents the frequency distribution of languages used in the evaluations, showing only languages with a frequency of two or more.
The colors indicate the language resource classes from \citet{joshi-etal-2020-state}, with darker colors (with higher numbers) representing higher-resource languages.
As discussed in previous surveys \cite{liu2024culturally, adilazuarda2024towards}, most studies collect data in English.
Chinese, Spanish, and German are other high-resource languages observed frequently in cultural studies.
Also, research in Korean, Indonesian, Bengali, and Swahili has been relatively active compared to other languages of their resource levels.
Notably, for most languages classified as level 2 or below, there are at most seven studies, with the only exception of Swahili, underscoring the gap in research.

Figure~\ref{fig:country-dist} visualizes the target countries of the research on cultural language models.\footnote{We generate the choropleth map with Datawrapper at \url{https://www.datawrapper.de/maps/choropleth-map}.} Most of the studies focus on WEIRD (Western, Educated, Industrialized, Rich, and Democratic) countries \cite{henrich2010weirdest}, along with regions such as East Asia, Indonesia, and India. In contrast, countries in Africa, Central and South America, Eastern Europe, and Central Asia are significantly underrepresented. Even for studies including underrepresented nations, the diversity of data sources tends to be limited, often relying on global surveys like the World Values Survey (WVS) \cite{wvs2022} or publicly available online platforms such as Wikipedia. Additionally, the volume of data points across different regions frequently varies, further contributing to an uneven representation.

The scope of our analysis of cultural coverage is limited by the inability to account for macro-level regions such as continents and broad cultural groups with boundaries that do not align with national borders.
In particular, while Arabic is one of the languages studied extensively, many of those studies broadly lump the cultural sphere together under the label of the `Arab world.'
Recognizing the challenges of defining the exact boundaries of cultures, researchers should nonetheless strive for the most accurate representation of the cultures they examine. Additionally, our analysis lacks finer-grained sub-regional or historical district breakdowns, which represents another notable gap in current cultural NLP research.



\section{Implications to AI Ethics, Social Sciences, and HCI}
\label{ethical_implications}

\subsection{Ethics of Cultural Alignment}

The incorporation of cultural information into general-purpose foundational models is an important research direction. Whether the outputs of these models are intentional or accidental, they co-create our sense of meaning and identity and have an impact on shaping our collective knowledge~\citep{lu-etal-2022-ai-identity}. Lack of appropriate cultural representation can lead to several harms, including disparate access due to performance gaps, imposition of hegemonic classifications, violation of cultural values, misinformation, or stereotypes about cultures, misrepresentation of cultural experiences, and outright erasure of cultures~\citep{prabhakaran2022culturalincongruencies, kayEpistemicInjusticeGenerative2024}.

Communication across different cultures differs substantially, some are low context while others are highly contextual in the way they communicate, both in the real world~\citep{VerbalCommunicationStylesandCulture} and online~\citep{wurtz-cross-cultural-comm}. Imposition of one culture's communication style on another can lead to erasure and flattening of cultures. A mismatch or misalignment in these styles can lead to problems pertaining to intercultural communication like misunderstanding, distrust, and conflict. Therefore, it is crucial to be careful in approaches towards cultural alignment. Further, since using generative models for writing can impact the opinions of the users themselves~\citep{jackesch-etal-2023-cowriting-opinion}, there are also broader questions about the systemic impact on users and society at large~\citep{burtonHowLargeLanguage2024a} that need to be taken into account during this process.

To tackle the problem of lack of cultural knowledge, several papers have tried to adapt existing models and incorporate features for other cultures across different tasks, hoping to improve performance on tasks requiring cultural knowledge. These include tasks like Affect Detection~\citep{neiberg_intra-_2011}, Offensive Language Detection~\citep{zhou_cultural_2023}, Humor Detection~\citep{xie_multimodal_2023}, Recipe Adaptation~\citep{cao_cultural_2024}, among others.
Across these efforts, there is an underlying assumption that since cultural knowledge is required for these tasks, cultural alignment will improve cross-cultural performance on them. However, while there are improvements on the performance on the datasets, it is unclear whether the improvement is due to the actual incorporation of cultural knowledge or due to surface level features in the datasets that the models are picking up. When evaluated with domain experts, models often fail to appropriately use cultural information. For instance, in experiments with comedians for co-creating humor, LLMs fail to produce non-bland or generic outputs, especially when text is about cultures other than the dominant ones embedded into the model~\citep{mirowski-etal-2024-llm-humour} or when used for mental health support, LLMs fail to adapt based on the cultural background of the users and provide misaligned recommendations~\citep{song2024typing}.

There are also issues concerning what cultural information is available to encode. Since data for inclusion is primarily scraped from the internet, which is a biased sample of what cultural information exists, it only captures some aspects of knowledge~\citep{bender-etal-2021-stochiastic-parrots}. The long tail of cultural information, which pertains to everyday tasks, is unspecified or not recorded and hence does not make it to the datasets. There are efforts to address this and benchmark the performance of models on everyday knowledge ~\citep{myung2024blendbenchmarkllmseveryday}, but the area is largely ignored, making it a core limitation for both model designers and practitioners.
Further, since large language models are increasingly being used as writing assistants and sources of synthetic data, which has an impact on the diversity of the content that is generated~\citep{padmakumar2024doeswritinglanguagemodels} and the values embedded therein~\citep{wright2024revealingfinegrainedvaluesopinions, pmlr-v202-santurkar23a}. The data that is being used to train these models will constantly reinforce one set of values and more biased models. This will lead to poorer representation of diverse cultural representation in model outputs, resulting into a potential cultural model collapse~\citep{shumailovAIModelsCollapse2024a}. 

Beyond harms associated with non-inclusion or simplistic inclusion of cultures, there are also harms associated with explicit inclusion of a culture. ~\citet{kirkBenefitsRisksBounds2024} outline this by creating a taxonomy discussing the benefits and harms of personalizing language models. They show that while there are clear use cases of aligning language models such as the increased autonomy, empathy, and usefulness, one should be considerate of the often overlooked harms that such alignment can bring. The study shows how each benefit that personalization brings has a potential harm resulting from it, recommending that model designers and practitioners have to take these trade-offs into account when creating or deploying these models. For instance, they increased usefulness or empathy in models can lead to dependency on these models and contribute to their anthropomorphism. Similarly, at a societal level, adaptation to each culture can contribute to increased polarization and labor displacement.

Further, it is unclear what the correct approach is when the culture that needs to be included has fundamentally opposing values to the ones where models are usually created. When NLP research suggests alignment, it is typically associated with cultures which are not at odds with the value systems of western nations. Further, there is also cultural information that is unsafe. For instance, medical advice from certain cultures challenges western notions of medicine and advises against relying on it, instead promoting local forms of medicine, which can at times be harmful. Another related example is alignment to fringe or extremist communities. With adapted generative models, the potential for harm that they can cause would also be much higher when used for nefarious purposes by malicious agents~\citep{kaffee-etal-2023-thorny}. Alignment in such scenarios can lead to unsafe behavior, thus bringing forward this trade-off between two desirable characteristics of model behavior.

Finally, there are also questions about whether building general-purpose systems suitable for all audiences is the right way forward~\cite {gabrielArtificialIntelligenceValues2020}. In their work, \citet{zhixuan2024preferencesaialignment} highlight some assumptions made by current AI alignment efforts, namely that human preferences (which is the dominant method of encoding values) are an adequate representation of human values and AI systems should be aligned with preferences of one or more humans to ensure that they behave safely and in alignment with our values. They challenge these assumptions, critiquing the normativity of expected utility theory as the dominant method for alignment of AI assistants. They argue that these systems should instead be domain-specific and aligned based on standards negotiated upon by the corresponding users and stakeholders, allowing for meeting diverse needs and co-existing in the presence of pluralistic values. In a cultural context, such a system would benefit multicultural societies where certain normative standards have been established.


\subsection{Accelerating Social Science Research}

There has been a lot of optimism and uptick in adoption from the social sciences towards AI systems aiding in research. They have been used in several fields, including psychology, sociology, political science, history, and many others, for different tasks. 
While the appropriateness and motivation for using them as an accurate representation of society is an ongoing discussion~\citep{grossmannAITransformationSocial2023, 10.1145/3613904.3642703}, they certainly aid in performing several subtasks relevant for social science research~\citep{ziemsCanLargeLanguage2024, bailCanGenerativeAI2024}. Across these fields, they serve a variety of different purposes. 
For some, they aid in the analysis of large volumes of content~\citep{tornbergHowUseLargeLanguage2024}, extracting dispositions from social media~\citep{petersLargeLanguageModels2024} to make inferences about users, while for others it is simulation of responses of human samples for surveys~\citep{Argyle-etal-2023-human-samples, pmlr-v235-manvi24a} or serving as agents for agent-based modeling methods for predicting hypothetical behavior~\citep{NBERw31122, grossmannAITransformationSocial2023}. 
They have also been used for interventions to existing ecosystems~\citep{yangSocialSkillTraining2024, argyleLeveragingAIDemocratic2023}, trying to address existing issues like misinformation or polarity. Since most of these tasks are rely on models appropriately reflecting people from different cultures faithfully, careful cultural alignment is a crucial part in this process. Machine learning models are trained to generalize and learn from abstractions in data. This can lead to flattening of identity~\citep{wang2024largelanguagemodelsreplace} or poor performance on and misrepresentation of non-majority cultures.
Such an effect can cast doubt on inferences made on top of results extracted from these systems. 
Thus, cross-cultural alignment with robust human evaluation is imperative for reliable inferences made in the social science.

For achieving this alignment and embedding cultural information and knowledge, however, practitioners often use sources and literature from the social sciences (\S\ref{sec:source}). Simultaneously, generative models are proposed as a means to replace human participants in surveys~\citep{Argyle-etal-2023-human-samples, pmlr-v235-manvi24a}. Such a loop can lead to vicious circle, where cultures are misrepresented and cultural change is not incorporated. So, social scientists need to be aware of sources of data used for training these models, and the biases that may be embedded in them corresponding to the populations they are studying. Further, generative model designers need to be careful while culturally aligning these models to not over-rely on social science survey data, when not directly extracted from human participants from different cultures but is rather, synthetically generated.

%


\subsection{Human Computer Interaction and Cultural Alignment}
\label{sec:hci}
Another important aspect of cultural alignment is how people interact with the culturally aligned LLMs and the corresponding interaction patterns. Understanding these interaction patterns includes studying how cultural alignment for models affects their use in applications such as creating generative art, generating culturally relevant stories \citep{toro-isaza-etal-2023-fairy}, professional communication, cross-cultural communication, etc. \citet{weidinger2023sociotechnical} propose a three-layered approach to evaluating this effect in AI systems: the first layer is their capability, the second is how the system affects human interaction with the system and the last layer being understanding the impact of
the system on the broader context in which
it is embedded, such as society, the economy,
and the natural environment. Using this framework to understand the embedding of cultural information in generative models from an HCI perspective, we find that most research has focused on building and evaluating the capability (cultural understanding and awareness). On the other hand, ensuring cultural inclusion during human interaction with the system and studying broader systemic impact has received very little attention. For instance, generative models are tested for the understanding of culture-specific references in a conversation and the ability of a model to produce (culturally) relevant responses, testing their capability. However, the risk of people being deceived, misled, or enraged by that output because of them being misaligned with cultures depends on factors such as the context in which an AI system is used, who uses it, and the features of an
application. Such evaluation is rarely performed before deployment of current generative systems, and is imperative for safe deployment in cultures distant from the ones that models are biased towards. Further, since the usage of these models can impact the users themselves, these factors have to be studied in the context of population-level effects. For instance, political values and opinions from biased models affecting the opinions of users~\citep{jackesch-etal-2023-cowriting-opinion, fisher2024biasedaiinfluencepolitical}
it can lead to a shift in norms and values that a culture identifies itself with \citet{wagner2021measuring}.

An HCI-based perspective on cultural inclusivity in generative models would include adapting the LLMs to the needs and expectations of culture and the intended applications (e.g., high risk vs. low risk). As the models become more general purpose, there needs to be a distinction between the tasks and applications that would require culturally agnostic capabilities vs. those that would require culture-specific capabilities \citep{cetinic2022myth}. The HCI component should drive the data-collection for cultural alignment, as some cultures might be over-represented and while others might be misrepresented due to variance of technology access and expectations across different cultures. Participatory frameworks for co-designing these models and the data used for training them, involving stakeholders from the corresponding cultures, is one effective approach of addressing current gaps in culturally misaligned models~\citep{10.1145/3551624.3555290}.
One use case where such an approach is crucial is creation of generative art using these models. Since aesthetic standards, expression, and the emotions that different elements of art invoke are different in different cultures (Section~\ref{sec:art}), it is imperative to have faithful representation of cultural artifacts.  

\section{Pointers to Future Research}
\label{fut}

\textbf{\textsl{Expand Research on Low-Resource Cultures and Languages.}}
Research on low- and mid-resource cultures and languages has progressed but remains limited compared to high-resource counterparts, as discussed in section \ref{sec:linguistics} and \ref{coverage}. Thus, more efforts are needed to collect data and evaluate the language models on low-resource cultures and languages. For example, creating benchmarks for dialects---which capture unique aspects of local culture---is important, especially as many of these dialects are at risk of disappearing~\cite{moseley2010atlas}. In regions with low technology access, e.g., Sudan, involving native annotators becomes essential due to the limited available web data. However, these methods are not scalable, underscoring the need for more research into scalable data collection for low-resource languages and cultures. Additionally, it is necessary to develop alignment methodologies that can perform effectively with relatively small datasets.\\

\textbf{\textsl{Vary Data Collection Strategies Across Cultures.}}
While collecting data for low-resource languages and cultures, considerable emphasis should be given to technology access of the cultures. Technology access determines how the data collection strategy should be varied. For example, although cultures like Estonian and Finnish are low-resource cultures, data collection strategies for that culture would involve scraping region-specific web data, recruiting annotators, and running crowdsourced experiments to understand cultural preferences due to high penetration of technology (mobiles, internet, etc.) in those cultures\footnote{Internet penetration statistics: \url{https://worldpopulationreview.com/country-rankings/internet-penetration-by-country}}. On the other hand, for cultures with low technology penetration (e.g., Sudan), the data collection would involve on-ground annotators talking to native people to collect data. The population of people following the culture would affect data collection, as some languages and cultural practices are spoken and followed by people in a restricted domain \citep{liu2022not}. While the data collection strategies would vary across cultures, care must be taken to standardize the data (for example by having humans in the loop) to ensure equity of model performance across cultures, as both methods would lead to differences in the quality of data.\\

\textbf{\textsl{Approach Defining Cultural Boundaries with Caution.}}
In language and vision research, culture is often represented through language or geographical regions, typically at the country level. However, countries do not always align with cultural boundaries~\cite{https://doi.org/10.1525/aa.2004.106.3.443}. For instance, Indonesia---one of the most ethnographically diverse nations globally---contains a wide array of local cultures not captured by a single national identity. Recent efforts aim to incorporate these local cultures into cultural benchmarks~\cite{putri2024llmgenerateculturallyrelevant, koto2024indoculture, koto-etal-2023-large}, though such attempts remain limited to certain regions. Using language as a cultural proxy also presents challenges, as languages like English, Spanish, or Arabic can span multiple cultural contexts~\cite{lee-etal-2023-hate}. Therefore, it is crucial to carefully define cultural boundaries when conducting cross-cultural research or developing culture-specific benchmarks. One effective approach to address this challenge is to engage in interdisciplinary collaboration with sociolinguistic researchers, who can provide deeper insights into the nuances of cultural and linguistic diversity.\\

\textbf{\textsl{Ensure Inclusive Cultural Representations.}}
Even within the same region or cultural group, social values and norms can vary significantly based on demographics such as age, gender, and race~\cite{weber2017examining}. Therefore, when constructing cultural datasets or benchmarks, it is essential to involve annotators with diverse demographic backgrounds, even within a single cultural group. Moreover, as cultural values and norms can vary between individuals, using annotators from a specific demographic group might not be fully representative of the culture. For instance, when gathering responses to commonsense questions like `What is a common school cafeteria food in your country?', relying on a small, homogeneous group of annotators can lead to incomplete or biased representations. A diverse and sizable pool of annotators is essential to capture a full range of perspectives. Additionally, evaluating the level of agreement among annotators can help determine if the \emph{gold} answer truly reflects the culture context~\cite{havaldar-etal-2024-building}.\\

\textbf{\textsl{Develop LLMs that can Adapt and Evolve with Cultural Change over Time.}}
Another important factor to consider is the dynamic nature of culture. As \citet{naylor1996culture} noted, no culture is static; people continually adapt to changes in their physical and sociocultural environment. Especially in today's globalized world, interactions between different cultural groups can quickly lead to the emergence and transformation of new cultural identities~\cite{holton2000globalization}. While some studies have examined historical cultures~\cite{wei2024ac, tang-etal-2024-chisiec}, there remains a notable gap in research on how to adapt LLMs as cultures evolve. Addressing this challenge requires moving beyond static LLMs that only align with current cultural norms. Instead, LLMs should act as repositories for cultural preservation and adaptable systems that can respond to ongoing cultural transformations.\\

\textbf{\textsl{Alternative Image Data Collection to Mitigate Biases in Web Images.}}
Most vision benchmarks rely on images sourced from the web, as discussed in section \ref{sec:source}. However, web images are susceptible to various biases like availability bias (different subjects, light conditions, locations, camera settings, and other features may be more likely to be uploaded on the web than others), apprehension bias (people may pose and look differently when they know that they are being photographed), and negative set bias~\cite{goldman2024statistical}. For example, certain subjects and locations are more likely to be uploaded online, and people may pose differently when photographed. These biases could result in omitting everyday objects and cultural concepts on the web. To mitigate these biases, we could actively photograph culturally relevant concepts and objects with guidance from local residents and anthropology experts. Additionally, using frames from videos that document themes such as `a typical day in the life of a person with a specific identity' or content from regional TV-shows can help capture more realistic and broader cultural images.\\



\textbf{\textsl{Expand Cultural Evaluation Methods to Diverse Interaction Settings.}}
As discussed in section \ref{eval}, culturally aware LLMs are mostly evaluated using multiple-choice questions (MCQ). However, this approach has limitations, as it cannot fully capture the complexities of real-life human-AI interactions. MCQs primarily evaluate a predefined set of cultural knowledge and focus on explicit cultural norms. However, in real-world scenarios, human-AI communication involves natural dialogue, where LLMs need to interpret implicit cultural cues and generate culturally sensitive responses. One promising research direction is evaluating the long-form generation of LLMs, which recent studies have started to explore, as shown in Figure \ref{fig:eval-type}. However, most evaluations depend on human judgment or LLM-as-a-judge~\cite{NEURIPS2023_91f18a12} methods, underscoring a gap in culturally specific and robust automatic evaluation techniques. Therefore, further research is needed to develop reliable evaluation methods for assessing LLMs in natural, conversational settings, including long-form generation.\\

\textbf{\textsl{Balance Development of Culturally-specific LLMs and Comprehensive Universal LLMs.}}
Currently, various techniques are employed to culturally align LLMs, as discussed in sections \ref{sec:lang-align} and \ref{vis}. Most works have focused on training and developing culture-specific LLMs, particularly for non-Western local cultures. However, there has been comparatively less emphasis on creating cross-cultural models capable of reasoning across diverse cultural contexts. Given the diverse cultural backgrounds of users, it is essential for LLMs to possess a comprehensive cultural knowledge encompassing all high- and low-resource cultures. Therefore, it is essential to seek a balance between developing culture-specific LLMs tailored to local needs and creating comprehensive cross-cultural LLMs that can serve a global audience.\\

\textbf{\textsl{Develop Culturally Aware LLMs from User Perspectives.}}
Section \ref{sec:hci} discusses current applications of culturally aware LLMs, such as generating culturally relevant art, storytelling, and facilitating cross-cultural interactions. However, there remains a gap in understanding how users interact with culturally-aware LLMs from the user's perspective. This gap could be addressed through observational studies of user behavior in real-world scenarios. For instance, by observing when users are offended by an LLM's lack of cultural knowledge, we could gather insights for building safer, more culturally sensitive models. Additionally, studying interactions between multiple LLM agents and humans could reveal new applications, such as LLMs facilitating communication between individuals from diverse cultural backgrounds who speak different languages. Thus, observing real-world use cases from the user's perspective is important for developing practical, culturally aware LLMs.\\

\section{Conclusion}
This survey presented a comprehensive review of papers studying cultural inclusion in text-based and multimodal models. We surveyed recent research efforts toward cultural awareness in LLMs and have consolidated the efforts under various themes. We have defined cultural awareness in LLMs by leveraging definitions of culture in psychology and anthropology. We then discussed methodologies adopted for creating cross-cultural datasets, strategies for cultural inclusion in downstream tasks, and methodologies that have been used for benchmarking cultural awareness in LLMs. We also discussed several important topics, such as the role of HCI in cultural inclusion, the role of cultural alignment in accelerating social science research, and ethical issues related to cultural inclusion. We hope this survey will serve as a useful reference for future research on cultural alignment in AI systems.

\bibliography{main}

\end{document}